\documentclass[10pt,twocolumn,letterpaper]{article}

\usepackage[pagenumbers]{cvpr} 

\usepackage[accsupp]{axessibility} 

\definecolor{cvprblue}{rgb}{0.21,0.49,0.74}
\usepackage[pagebackref,breaklinks,colorlinks,allcolors=cvprblue]{hyperref}
\usepackage{pifont}
\usepackage{xcolor}
\usepackage{wrapfig}
\usepackage{dblfloatfix}
\usepackage{makecell}
\usepackage{graphicx}
\usepackage{subcaption}
\usepackage{tabularx}
\usepackage{caption}
\usepackage{capt-of}
\usepackage{multirow}
\usepackage[table]{xcolor}
\usepackage{longtable}
\usepackage{array}
\newcolumntype{L}[1]{>{\raggedright\arraybackslash}p{#1}}
\newcolumntype{C}[1]{>{\centering\arraybackslash}p{#1}}

\usepackage{booktabs}
\newcommand{\cmark}{{\color{green}\ding{51}}}
\newcommand{\xmark}{{\color{red}\ding{55}}}
\usepackage[table]{xcolor}
\usepackage{tikz}
\usepackage{graphicx}
\usepackage{array} 
\usepackage{float}

\usetikzlibrary{arrows.meta,positioning}
\usepackage{tcolorbox}
\tcbuselibrary{breakable,skins}

\newtcolorbox{promptbox}[1][]{%
  enhanced,
  breakable,
  colback=white,
  colframe=black,
  boxrule=0.5pt,
  left=6pt,right=6pt,top=6pt,bottom=6pt,
  width=\textwidth,
  #1}
\makeatletter
\newcommand\blfootnote[1]{%
  \begingroup
  \renewcommand\thefootnote{}%
  \long\def\@makefntext##1{\parindent 0pt\noindent ##1}%
  \footnotetext{#1}%
  \addtocounter{footnote}{-1}%
  \endgroup
}
\makeatother

\title{SEA: Evaluating Sketch Abstraction Efficiency via Element-level Commonsense Visual Question Answering}

\author{
\parbox{\textwidth}{\centering
Jiho Park \quad Sieun Choi \quad Jaeyoon Seo \quad Minho Sohn \quad Yeana Kim \quad Jihie Kim\textsuperscript{*}\\
Dongguk University, Republic of Korea\\
{\tt\small \{jiho8345, sieunchoi, pianoprince, alsghqlgodrl, yeana.kim\}@dgu.ac.kr, jihie.kim@dgu.edu}
}}

\begin{document}
\maketitle
\blfootnote{\textsuperscript{*} Corresponding author.\\
\textbf{Project page:} \url{https://zihos.github.io/SEA/} \\ \textbf{Author contributions:} JP led the project. JP, SC developed the methodology, designed experiments, and wrote the paper. JP, SC, JS, MS, YK collected the dataset and conducted experiments. JK supervised the project.}


\begin{abstract}
A sketch is a distilled form of visual abstraction that conveys core concepts through simplified yet purposeful strokes while omitting extraneous detail. Despite its expressive power, quantifying the efficiency of semantic abstraction in sketches remains challenging. Existing evaluation methods that rely on reference images, low-level visual features, or recognition accuracy do not capture abstraction, the defining property of sketches. To address these limitations, we introduce \textbf{SEA} ($\textbf{S}$ketch $\textbf{E}$valuation metric for $\textbf{A}$bstraction efficiency), a reference-free metric that assesses how economically a sketch represents class-defining visual elements while preserving semantic recognizability. These elements are derived per class from commonsense knowledge about features typically depicted in sketches. SEA leverages a visual question answering model to determine the presence of each element and returns a quantitative score that reflects semantic retention under visual economy. To support this metric, we present \textbf{CommonSketch}, the first semantically annotated sketch dataset, comprising 23,100 human-drawn sketches across 300 classes, each paired with a caption and element-level annotations. Experiments show that SEA aligns closely with human judgments and reliably discriminates levels of abstraction efficiency, while CommonSketch serves as a benchmark providing systematic evaluation of element-level sketch understanding across various vision-language models.
\end{abstract}
\section{Introduction}
\label{sec:intro}

\begin{figure}
\centering
\includegraphics[width=0.99\linewidth]{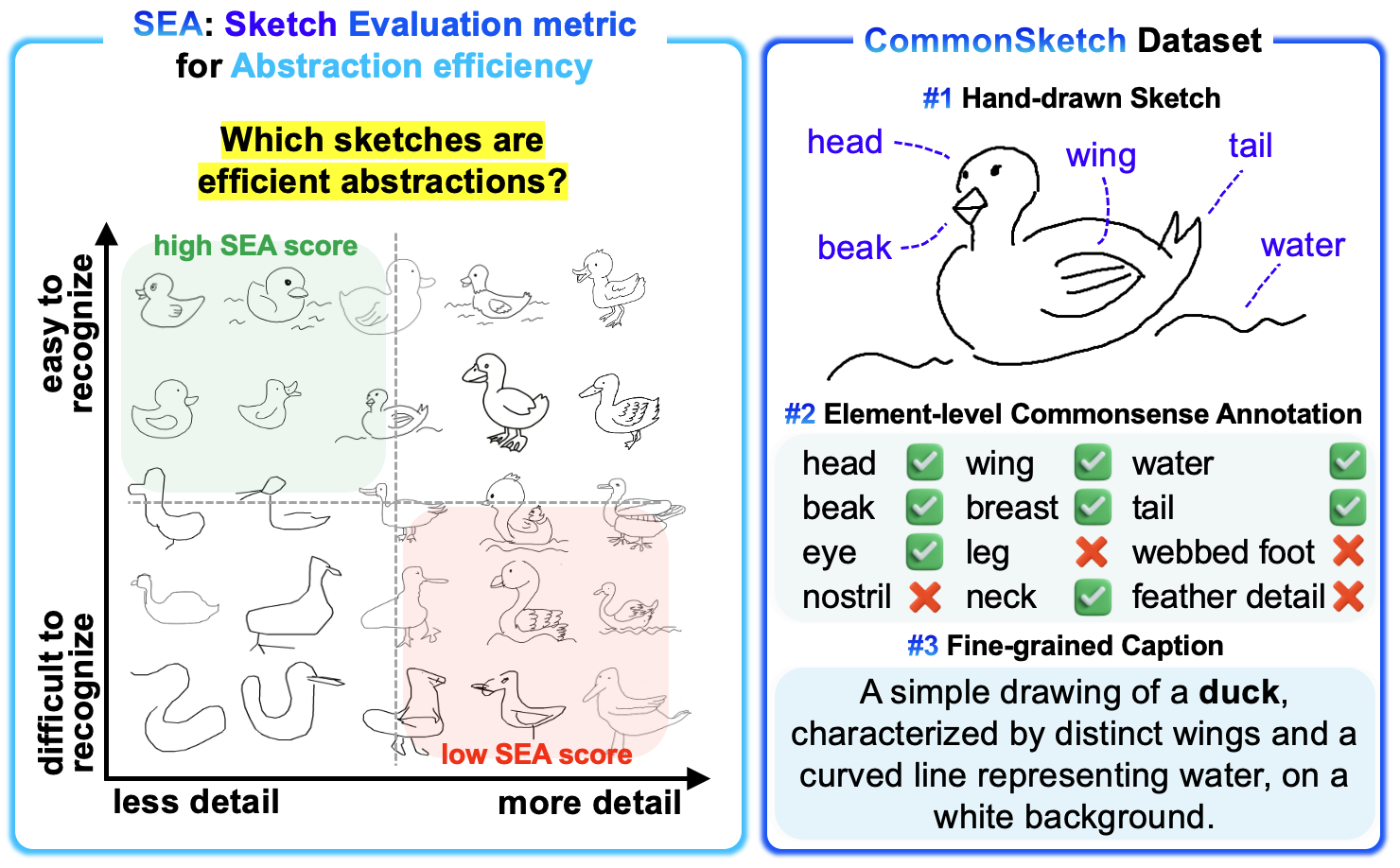}
\caption{\textbf{Overview of SEA and CommonSketch.}
 Left: SEA quantifies abstraction efficiency by balancing recognizability and detail. High scores (top-left) favor simple yet identifiable sketches, while low scores (bottom-right) denote ambiguity or over-detail. Right: CommonSketch includes element-level annotations and captions, enabling element-aware evaluation of sketch abstraction.
}
\label{fig:teaser}
\end{figure}

Sketches are among the most compact yet expressive forms of visual communication~\cite{viola2020visual}, conveying semantic intent through only a few strokes. This makes sketch understanding a useful setting for studying vision--language models (VLMs), since sketches must be interpreted from sparse, abstract, and selectively preserved visual cues. However, despite recent progress in sketch generation~\cite{vinker2022clipasso,hu2024scale,koley2024s,navard2024knobgen,arar2025swiftsketch} and recognition~\cite{ha2018neural,li2020sketch,yang2021sketchaa}, most prior work still formulates sketch understanding as category prediction or sketch--photo matching~\cite{sangkloy2022sketch,sain2024freeview,sain2025sketchdowntheflops} rather than reasoning about which visual elements are retained under abstraction. Such label-centric approaches overlook a defining property of sketches: they are deliberate abstractions, in which only a small subset of visually diagnostic elements is retained to convey meaning.

A key limitation of existing sketch datasets is that they rarely support reasoning within a sketch at the level of its constituent visual elements. Conventional datasets based on sketch--label~\cite{ha2018neural,eitz2012hdhso} or sketch--photo pairs~\cite{sangkloy2016sketchy,mukherjee2024seva} emphasize categorical correctness or appearance alignment, but they do not explicitly represent the process of abstraction---which elements must be drawn for a concept to remain recognizable, and which details can be omitted. As a result, current benchmarks provide limited support for analyzing how sketches preserve meaning under visual simplification.
To address this gap, we introduce \textbf{CommonSketch}, a dataset that associates each object class with a set of commonsense visual representatives, i.e., semantically diagnostic elements that people typically depict in sketches (e.g., wings for bird, handle for mug, or spokes and wheels for bicycle). CommonSketch goes beyond class-level recognition by providing human-drawn sketches, captions, and element-level annotations that enable verification of whether these class-specific representatives are present in each drawing. This shifts sketch understanding from label prediction to element-level reasoning about visual semantics and abstraction.

Beyond dataset limitations, current evaluation metrics fail to capture the unique nature of sketch abstraction. Most existing works rely on general-purpose image metrics, such as Top-$K$ accuracy, FID~\cite{heusel2017gans}, SSIM~\cite{wang2004image}, LPIPS~\cite{zhang2018unreasonable}, and DreamSim~\cite{fu2023dreamsim}, which are fundamentally designed to measure categorical recognizability or pixel-level appearance similarity. Because these generic metrics overlook the inherent characteristics of sketches, they cannot assess how efficiently a concept is conveyed with minimal visual representations, thereby failing to measure abstraction efficiency.
We therefore propose \textbf{SEA} (Sketch Evaluation metric for Abstraction efficiency), a metric designed to evaluate how effectively a sketch balances abstraction and recognizability. Given the commonsense visual representatives defined in CommonSketch, SEA measures which elements are visually expressed or omitted in a sketch and relates this abstraction pattern to the sketch's recognizability. High SEA scores are assigned to sketches that preserve recognizability while using relatively few visual elements. In this way, SEA provides a direct measure of abstraction quality beyond similarity-based evaluation.

As illustrated in Fig.~\ref{fig:teaser}, our framework addresses two complementary goals: identifying sketches that achieve efficient abstraction while remaining recognizable, and constructing a dataset that supports this analysis through human-drawn sketches, captions, and element-level annotations.
Our contributions are the following:
\begin{itemize}
    \item We introduce \textbf{CommonSketch}, a dataset that defines class-wise commonsense visual representatives and provides element-level annotations for verifying their presence in sketches, enabling a higher-level understanding of visual abstraction.
    \item We propose \textbf{SEA} (\textbf{S}ketch \textbf{E}valuation metric for \textbf{A}bstraction efficiency), a metric that quantifies how effectively a sketch conveys its concept through those representatives under minimal visual complexity.
    \item We empirically show that CommonSketch and SEA together enable a new perspective on sketch understanding by explicitly linking recognizability with abstraction efficiency.
\end{itemize}

\section{Related Works}
\label{sec:relatedwork}

\noindent\textbf{Sketch datasets.}
A variety of sketch datasets have been introduced for sketch understanding and generation. However, most existing datasets provide either sketch images with class labels or sketch--photo pairs. QuickDraw~\cite{ha2018neural} is one of the largest datasets for sketch classification, but its limited fidelity and lack of fine-grained annotations reduce its utility for generative modeling. TU-Berlin~\cite{eitz2012hdhso} contains more complex sketches, but still lacks instance-level descriptions. Sketchy~\cite{sangkloy2016sketchy} provides sketch--photo pairs, yet its visually complex sketches and limited semantic annotations restrict its suitability for controllable sketch generation. Recent datasets such as SEVA~\cite{mukherjee2024seva}, which uses CLIPasso~\cite{vinker2022clipasso} to derive stroke-based sketches from images, still rely heavily on sketch--photo pairs and do not provide the fine-grained instance-level captions needed for controllable generation. In contrast, our proposed dataset, CommonSketch, provides sketches with detailed semantic annotations grounded in commonsense knowledge. It supports a broad range of sketch understanding and generation tasks through element-level labels and structured captions. A comparison with existing sketch datasets is provided in Tab.~\ref{tab:sketch_datasets}.

\begin{table}[t]
  \centering
  \caption{\textbf{Comparison of prior sketch datasets and ours.} This table summarizes key attributes of several sketch datasets: TU-Berlin, Sketchy, QuickDraw, SEVA, and ours.}
  \label{tab:sketch_datasets}
  \resizebox{\columnwidth}{!}{
  \begin{tabular}{lccccccc}
    \toprule
    \textbf{Dataset} & \textbf{\# Classes} & 
    \makecell{\textbf{\# Sketches/}\\\textbf{Class}} & 
    \makecell{\textbf{Total \#}\\\textbf{Sketches}} & 
    \makecell{\textbf{Common}\\\textbf{Sense}} & \textbf{Caption} & \textbf{QA} \\
    \midrule
    TU-Berlin~\cite{eitz2012hdhso} & 250 & 80 & 20K & \xmark & \xmark & \xmark \\
    Sketchy~\cite{sangkloy2016sketchy}   & 125 & avg.\ 600 & 75K & \xmark & \xmark  & \xmark \\
    QuickDraw~\cite{ha2018neural}     & 345 & avg.\ 144K & $\sim$50M & \xmark & \xmark & \xmark \\
    SEVA~\cite{mukherjee2024seva}  & 128 & avg.\ 703 & 90K & \xmark & \xmark  & \xmark \\
    \midrule
    \textbf{CommonSketch (ours)}& 300 & avg. 77 & 23K & \cmark & \cmark & \cmark \\
    \bottomrule
  \end{tabular}}
\end{table}
\captionsetup[sub]{font=small, skip=4pt} 

\begin{figure*}[t]
  \centering
  \begin{minipage}{0.49\textwidth}
    \centering
    \subcaptionbox{%
      \textbf{Data construction pipeline.} We collect \emph{human-drawn} sketches, verify labels via captioning, extract sketch commonsense, and have annotators mark element presence per sketch.
      \label{fig:overview-a}}{%
      \includegraphics[width=\linewidth,keepaspectratio]{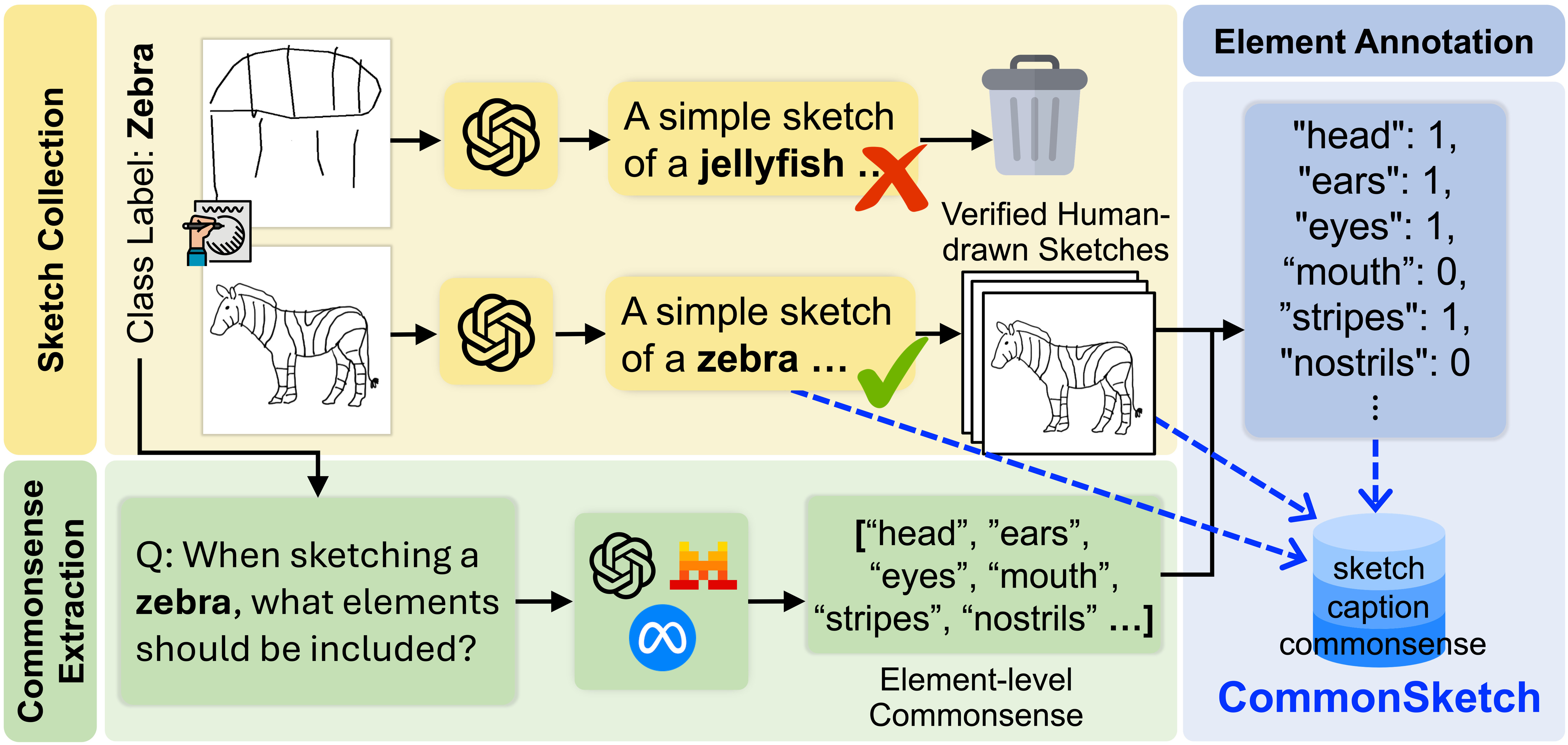}
    }\\[2pt]
    \begingroup
      \captionsetup{skip=0pt}
      \subcaptionbox{%
        \textbf{Average elements by category.} Mean commonsense elements per
        category. Animals highest (12.3); sports equipment lowest (6.4).%
        \label{fig:overview-b}}{%
        \includegraphics[width=\linewidth,keepaspectratio]{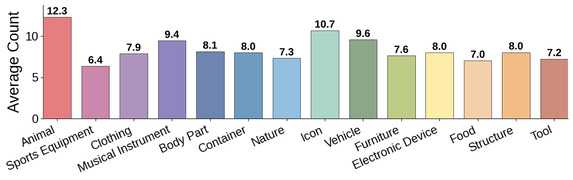}
      }
    \endgroup
  \end{minipage}\hfill
  \begin{minipage}{0.49\textwidth}
    \centering
    \begingroup
      \captionsetup{skip=5pt}
      \subcaptionbox{%
        \textbf{Classes per category \& element-level commonsense example.} Distribution of 300 classes across 14 categories, and a sample commonsense element set for \emph{zebra}.
        \label{fig:overview-c}}{%
        \includegraphics[width=\linewidth,keepaspectratio]{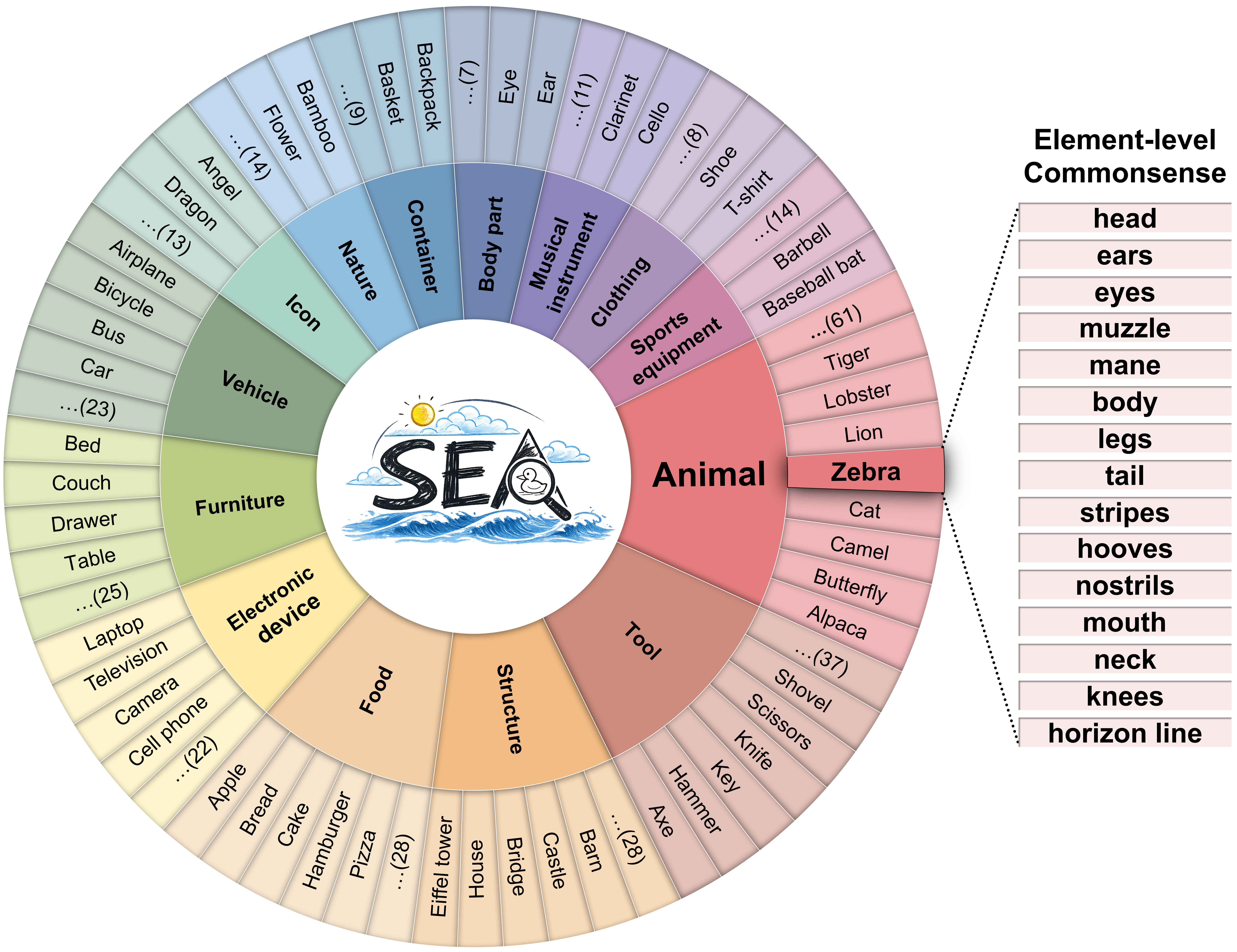}
      }
    \endgroup
  \end{minipage}
  \captionsetup{skip=2pt}
  \caption{\textbf{CommonSketch Overview.}
23{,}100 human-drawn sketches with paired captions and element-level commonsense across 300 classes in 14 categories; (a) construction/annotation pipeline, (b) category-wise element statistics, (c) class distribution with an example.}
  \label{fig:zebra}
\end{figure*}

\noindent\textbf{Image evaluation metrics.}
Sketch generation is often evaluated using classification-based measures or generic image-generation metrics. However, these metrics were largely developed for photorealistic images, whereas sketches are sparse, abstract, and line-based. This gap limits their suitability for sketch evaluation. CLIPasso~\cite{vinker2022clipasso} and Kampelm\"uhler et al.~\cite{kampelmuhler2020synthesizing} measure recognizability using classifier accuracy. Generic metrics include FID~\cite{heusel2017gans} for distributional similarity, and SSIM~\cite{wang2004image}, LPIPS~\cite{zhang2018unreasonable}, and DreamSim~\cite{fu2023dreamsim} for reference-based structural or perceptual similarity. Hu et al.~\cite{hu2024scale} also use FID to compare generated and real sketches. CLIPScore~\cite{hessel2021clipscore}, measures text-image alignment, but does not capture sketch-specific properties such as abstraction or semantic expressivity. As a result, these metrics do not directly assess the structural quality and abstraction characteristics of sketches.

\noindent\textbf{Sketch evaluation metrics.}
Recent sketch generation and sketch-conditioned synthesis methods are typically evaluated using recognizability, fidelity, or generic image-generation metrics rather than metrics tailored to abstraction quality~\cite{vinker2022clipasso,arar2025swiftsketch,koley2024s,navard2024knobgen}. SketchRef~\cite{lin2024sketchref} provides a more direct measure of structural consistency by using mean Object Keypoint Similarity (mOKS) and CLIP-based cosine similarity to evaluate feature preservation and recognizability, respectively. However, it cannot be applied when no reference image is available, as in text-to-sketch generation. Geometry-Aware Classification Layer (GACL)~\cite{yang2022finding} computes annotation-free quality scores based on classification geometry. Because it depends on a classification layer, the evaluation remains classification-oriented. Higher scores also tend to favor detailed and visually complex renderings, making GACL unsuitable for assessing visual abstraction. In addition, its scores depend on the choice of category-supervised backbone and representation, so absolute values are not reliably comparable across datasets or domains. To address these limitations, we propose SEA, a reference-free metric that uses commonsense knowledge to evaluate whether a sketch conveys the semantic elements of its class through appropriate abstraction.

\section{CommonSketch}
\label{sec:commonsketch}
\subsection{Dataset Overview}
We introduce CommonSketch, a novel dataset comprising 23,100 instance-level sketches across 300 object classes and 14 categories (Fig.~\ref{fig:zebra}). For every class, we define a set of commonsense elements consisting of the externally visible parts typically drawn when depicting the object, as illustrated on the right side of Fig.~\ref{fig:overview-c} (e.g., head, ears, eyes, mouth, stripes, and nostrils for a zebra). Each sketch is paired with a natural language caption and element-level commonsense annotations specifying which visual components are present. The overall construction pipeline is visualized in Fig.~\ref{fig:overview-a}, while Fig.~\ref{fig:overview-b} and Fig.~\ref{fig:overview-c} summarize the category-wise distributions and the complete category-class taxonomy, respectively.

\subsection{Dataset Construction}
\noindent\textbf{Sketch collection.} 
Sketches were collected from 12 volunteers through an open call under a standardized drawing protocol. Participants were non-art majors and were not pre-screened for drawing ability. Given a class label, each participant was asked to draw a single object within 60--80 seconds using a tablet and pen. Sketches were drawn with the default drawing application on each participant's device and saved as 512{\(\times\)}512 PNG files. Each sketch consisted of black lines (\#000000) on a white background (\#FFFFFF), with no post-processing applied.

\noindent\textbf{Caption generation.} 
To construct high-quality captions paired with each sketch, collected sketches were processed through GPT-4o~\cite{hurst2024gpt} to generate descriptive captions. Additionally, we leveraged caption generation as a validation mechanism: sketches whose generated captions did not contain the target class label were discarded and redrawn. This dual-purpose approach ensured both the creation of a comprehensive caption dataset and the maintenance of label consistency, thereby improving overall sketch quality.

\noindent\textbf{Commonsense extraction.} 
Using validated sketches as input,  we extracted element-level visual commonsense knowledge for each class through GPT-4o prompting. The complete prompt used for extraction is provided in the supplementary material. In this paper, we primarily utilized the commonsense elements generated by GPT-4o, while also replicating the extraction process using other open-source Large Language Models (LLMs) to evaluate reproducibility and consistency. The resulting candidate elements were thoroughly reviewed by human annotators. During this review, only elements that are externally observable and visually representable in sketches were retained, whereas internal or non-visible components, such as the heart or brain were excluded. Through this refinement process, we constructed a balanced and representative set of visual commonsense elements as shown in Fig.~\ref{fig:overview-a}.

\noindent\textbf{Commonsense element annotation.}
Human annotators established ground truth through binary annotation. They labeled each commonsense element as either present or absent for every individual sketch. This process required annotators to assess whether these extracted commonsense elements were recognizably depicted within each sketch. Through this systematic annotation process, we tracked the frequency of each visual element across sketches and adjusted the dataset to maintain a balanced distribution of element occurrences. Based on this process, we constructed a visual question-answering benchmark for evaluating the recognition of visual elements in sketches. The results are shown in Tab.~\ref{tab:vlm_element_eval}.

\subsection{Category and class composition.}
To ensure a balanced representation across diverse semantic domains, we curated 300 classes from the TU-Berlin~\cite{eitz2012hdhso} and QuickDraw~\cite{ha2018neural} datasets. These classes are organized into 14 distinct categories (Fig.~\ref{fig:overview-b}): food, animal, clothing, tool, sports equipment, vehicle, musical instrument, body part, container, furniture, electronic device, nature, structure, and icon. 
The categorization scheme was adapted from the THINGS database~\cite{hebart2019things}, with the original taxonomy expanded by adding ``structure'' and ``icon'' categories and broadening ``plant'' to ``nature.'' Further analysis of commonsense element distributions, including lift-based characterization of shared sketch primitives and elements characteristic of each category, is provided in the supplementary material.

\noindent CommonSketch comprises instance-level sketches, descriptive captions, and an element-level commonsense database, facilitating a granular analysis of visual element presence. Full details on the construction pipeline, including GPT-4o prompts and the complete category taxonomy, are available in the supplementary material.

\begin{figure*}[t]
  \centering
  \includegraphics[width=0.95\linewidth]{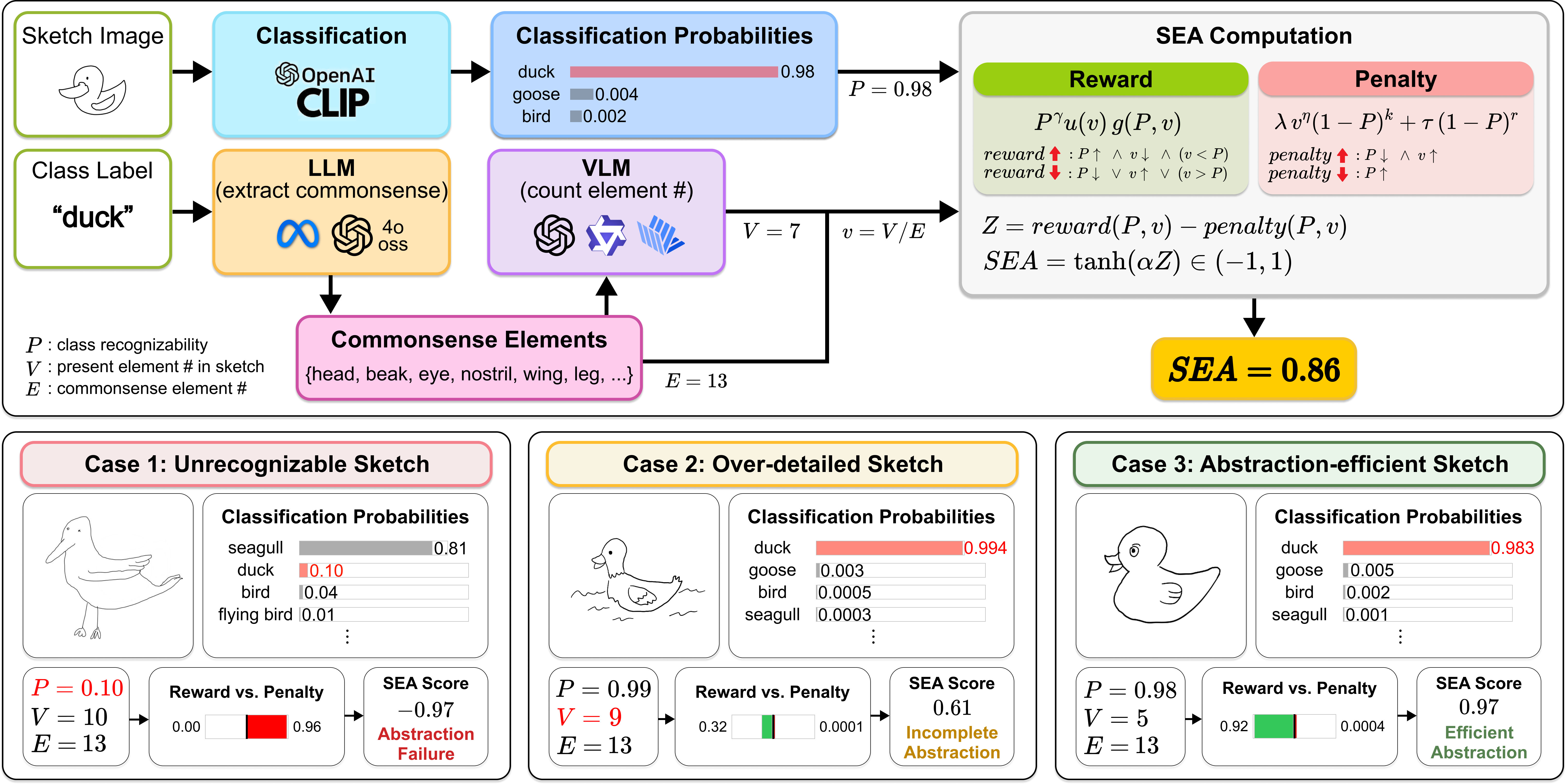}
    \caption{\textbf{Computation pipeline and case-based interpretation of the SEA metric.} Given a sketch and its class label, SEA combines class recognizability $P$ from a classifier with the commonsense element space $E$ extracted by an LLM and the number of visually grounded elements $V$ identified by a VLM. It then computes a reward--penalty balance and maps it to a bounded score $SEA \in (-1,1)$, where higher scores indicate sketches that preserve recognizability with minimal yet sufficient visual detail. Illustrative cases show abstraction failure due to low recognizability (left), incomplete abstraction caused by excessive detail (middle), and abstraction-efficient sketching that achieves high recognizability with fewer expressed elements (right).}
  \label{fig:sea-pipeline}
\end{figure*}

\section{SEA: Sketch Evaluation metric for Abstraction efficiency}
\label{sec:sea}
To quantitatively evaluate the efficiency of sketch abstraction, we propose the Sketch Evaluation metric for Abstraction efficiency (SEA), which integrates the three signals illustrated in in Fig.~\ref{fig:sea-pipeline}: the prediction probability $P$, the size of the commonsense element space $E$, and the number of actually represented elements $V$ in the sketch. SEA is designed to be smooth and sensitive to the trade-off between recognizability and visual abstraction.

\noindent\textbf{Notation.}
For each class, let $\mathcal{E}=\{e_1,\ldots,e_E\}$ denote the set of drawable commonsense elements obtained from the LLM, and let $\mathcal{V}\subseteq \mathcal{E}$ be the subset detected as present in the sketch by the VLM-based visual question answering (VQA) module as shown in Fig.~\ref{fig:sea-pipeline}. We define
\[
E = |\mathcal{E}|,\qquad 
V = |\mathcal{V}|,
\qquad 
v = V/E \in [0,1],
\]
where $v$ is the normalized visual ratio, representing the fraction of visual elements expressed in the sketch.
The prediction probability $P\in(0,1)$ is obtained from the zero-shot classifier as shown in Fig.~\ref{fig:sea-pipeline}, representing the confidence assigned to the correct class.  
A small constant $\delta>0$ is added inside logarithms and ratios for numerical stability. Unless otherwise noted, we set $\delta = 10^{-6}$ in all experiments.

\noindent\textbf{Overall structure.}
The SEA score is obtained by mapping a latent efficiency signal $Z$ through a hyperbolic tangent function:
\[
SEA = \tanh(\alpha Z),
\]
where $\alpha>0$ controls the sensitivity around the decision boundary. The signal $Z$ is defined as the difference between a reward term and a penalty term:

\begin{equation*}
Z=\mathrm{reward}(P,v)
-\mathrm{penalty}(P,v).
\end{equation*}
A positive $Z$ indicates that the sketch is efficiently abstracted with high prediction probability with minimal visual detail, whereas a negative $Z$ reflects either over-drawing or insufficient prediction probability.

\noindent\textbf{Reward term.}
The reward term encourages sketches that maintain recognizability while depicting a minimal set of visual elements:
\[
\mathrm{reward}(P,v)
= P^{\gamma}\,u(v)\,g(P,v).
\]
The first factor
\[
u(v)=\log\!\frac{1+\delta}{v+\delta}
\]
is an \emph{economy of expression} term: $u(v)$ increases as normalized visual ratio $v$ decreases, rewarding more abstract sketches.  
The second factor
\[
g(P,v) = \tanh\!\Big(\frac{\beta}{2}\log\frac{P+\delta}{v+\delta}\Big)
\]
acts as a \emph{centered gate} that enforces consistency between recognizability and visual ratio.
This gate establishes a self-consistency line where $g(P,v)=0$ at $v=P$. When $v < P$, indicating that the sketch achieves high recognizability with minimal elements, $g(P,v)$ becomes positive and amplifies the reward. In contrast, if $v > P$, the sketch is deemed overly detailed relative to its recognition probability, resulting in a negative $g(P,v)$ that suppresses the reward signal. The parameter $\beta$ controls how sharply this transition occurs, while $\gamma$ determines how strongly reward is suppressed when $P$ is small.

\noindent\textbf{Penalty term.}
The penalty term discourages unnecessary visual complexity and the failure of recognizability:
\[
\mathrm{penalty}(P,v) = \lambda\,v^{\eta}(1-P)^{k} + \tau\,(1-P)^{r}.
\]
The first component, $\lambda v^{\eta}(1-P)^k$, penalizes cases where a sketch exhibits a high visual ratio $v$ yet suffers from low recognition probability $P$. Specifically, $\lambda$ determines the overall magnitude of the visual complexity cost, while $\eta$ controls the growth of this cost relative to $v$, and $k$ modulates its sensitivity to recognition failure. The second component, $\tau(1-P)^r$, serves as a baseline penalty for sketches that are simply unidentifiable, independent of their visual ratio. The parameters $\tau$ and $r$ define the intensity and decay rate of this baseline penalty, respectively.

\noindent\textbf{Interpretation.}
In summary, SEA increases when a sketch achieves high recognizability with minimal visual elements, precisely the behavior illustrated in Fig.~\ref{fig:sea-pipeline}. Conversely, SEA decreases for sketches that are either overly detailed relative to their recognizability or fail to be identified despite their simplicity. Because the reward and penalty terms explicitly decompose the contributions from the recognition probability $P$ and the normalized visual ratio $v$, inspecting these terms allows us to diagnose \emph{why} abstraction fails in a given case: whether a low score stems primarily from insufficient recognizability (low $P$) or excessive visual complexity (high $v$). The final output $SEA \in (-1, 1)$ thus provides a continuous, differentiable, and interpretable measure of abstraction efficiency, suitable for both analysis and optimization.

\section{Experiments}
\label{sec:experiments}
\begin{figure*}
\centering
\includegraphics[width=0.99\linewidth]{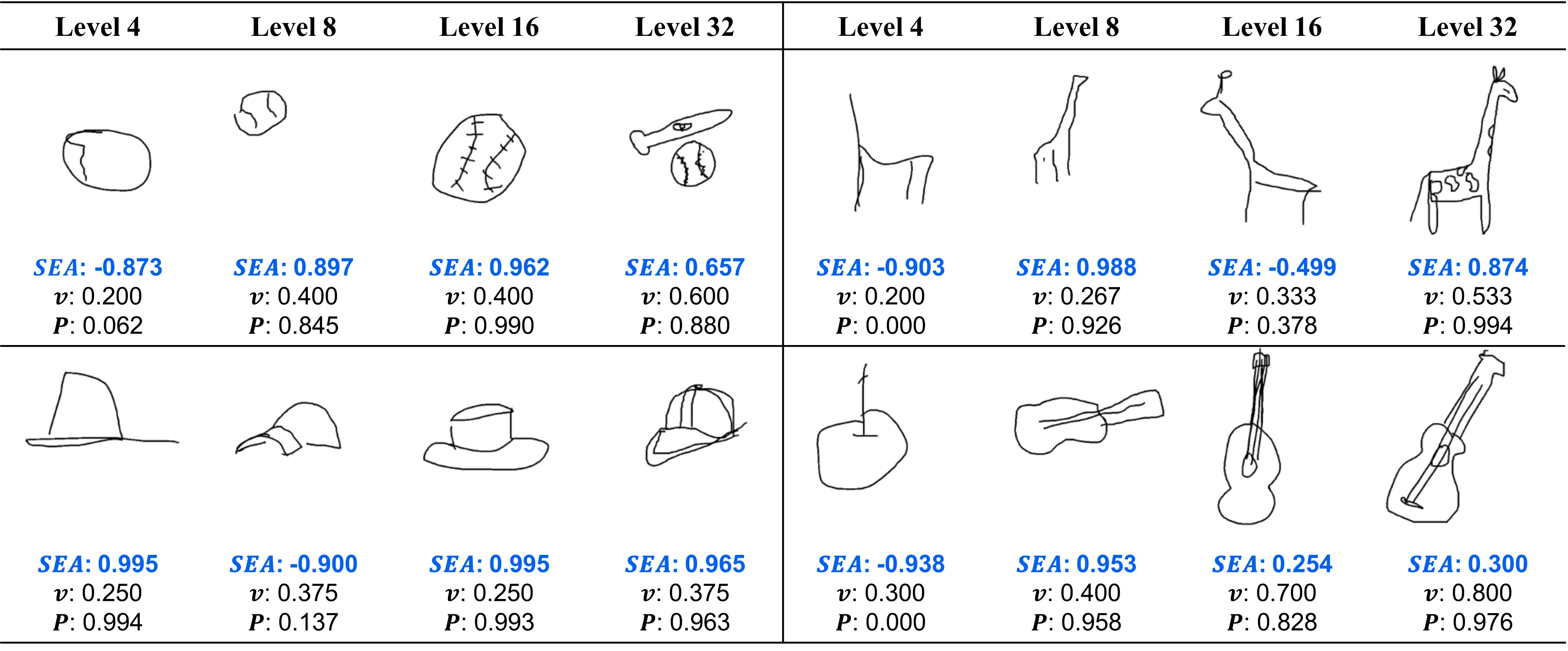}
\caption{\textbf{Qualitative comparison of SEA scores across abstraction levels (4, 8, 16, 32) on four classes shared by \textit{SEVA} and \textit{CommonSketch}—\textit{Baseball} (top left), \textit{Hat} (bottom left), \textit{Giraffe} (top right), and \textit{Guitar} (bottom right).} Each example reports the SEA score with visual ratio $v$, and prediction probability $P$, showing how abstraction can improve efficiency when recognizability is preserved.}
\label{fig:seva-qual}
\end{figure*}
\noindent\textbf{Implementation details.}
\label{sec:impl}
We benchmark CommonSketch against widely used sketch datasets: TU-Berlin~\cite{eitz2012hdhso}, QuickDraw~\cite{ha2018neural}, and SEVA~\cite{mukherjee2024seva}. 
To compute the SEA metric, we utilize various models to extract its core three signals: several LLMs (GPT-4o~\cite{hurst2024gpt}, GPT-OSS 20B~\cite{agarwal2025gpt}, Qwen-2.5 32B~\cite{qwen2.5}, Llama 3 8B~\cite{grattafiori2024llama}, and Mistral 7B~\cite{DBLP:journals/corr/abs-2310-06825}) to extract the commonsense elements ($E$); various VLMs (GPT-4o, Qwen2.5-VL 7B~\cite{bai2025qwen2}, mPLUG-Owl3 7B~\cite{ye2024mplug}, InternVL3 8B~\cite{zhu2025internvl3}, Molmo 7B~\cite{deitke2024molmo}, PaliGemma2 3B~\cite{steiner2024paligemma}, SmolVLM 500M~\cite{marafioti2025smolvlm}, LLaVA 1.5 7B~\cite{liu2023visual}, and BLIP~\cite{li2022blip}) to identify the visual elements represented in the sketch ($V$); and zero-shot classifiers (CLIP-ViT-L/14~\cite{radford2021learning}, OpenCLIP~\cite{ilharco2021openclip}, and CoCa~\cite{yu2022coca}) to obtain the prediction probabilities ($P$). 
To ensure consistency and reproducibility, all hyperparameters for SEA are fixed across all experiments: $\alpha=2.2$, $\beta=8.0$, $\lambda=1.0$, $\eta=0.8$, $k=2.3$, $\tau=0.4$, $r=1.7$, $\gamma=1.7$, and $\delta=10^{-7}$. Further details on the hyperparameter settings and selection are available in the supplementary material.

\newcommand{\mstd}[2]{#1{\tiny$\pm$#2}}

\begin{table}[t]
\caption{\textbf{Quantitative SEA results on SEVA across abstraction levels.} Mean $\pm$ standard deviation of SEA, reward, penalty, visual ratio $v$, and prediction probability $P$.}
\label{tbl:seva-quan}
\centering
\scriptsize
\setlength{\tabcolsep}{6pt}
\begin{tabular}{l|cccc}
\toprule
Metric & \textbf{Level 4} & \textbf{Level 8} & \textbf{Level 16} & \textbf{Level 32} \\
\midrule
SEA & \mstd{-0.56}{0.32} & \mstd{-0.23}{0.30} & \mstd{0.29}{0.24} & \mstd{0.43}{0.29} \\
\midrule
reward($P$,$v$)     & \mstd{0.16}{0.20} & \mstd{0.30}{0.20} & \mstd{0.52}{0.20} & \mstd{0.57}{0.26} \\
penalty($P$,$v$)    & \mstd{0.53}{0.11} & \mstd{0.41}{0.09} & \mstd{0.24}{0.09} & \mstd{0.18}{0.11} \\
visual ratio ($v$)  & \mstd{0.22}{0.06} & \mstd{0.31}{0.08} & \mstd{0.40}{0.08} & \mstd{0.45}{0.10} \\
prediction ($P$)    & \mstd{0.17}{0.17} & \mstd{0.37}{0.15} & \mstd{0.64}{0.11} & \mstd{0.75}{0.11} \\
\bottomrule
\end{tabular}
\end{table}

\subsection{Evaluation of the SEA Metric}
\begin{figure}
\centering
\includegraphics[width=0.95\linewidth]{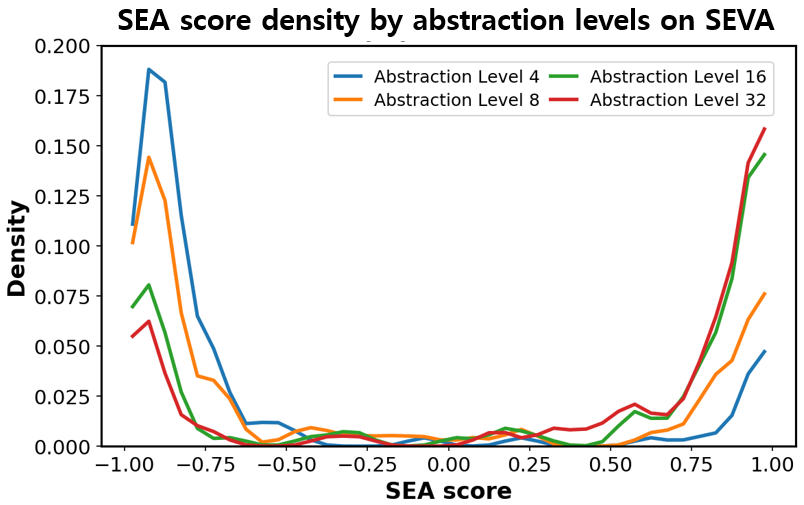}
\caption{\textbf{Distribution of SEA scores across abstraction levels in SEVA.} Sketches at lower abstraction levels (4, 8) are concentrated near low SEA scores, whereas those at higher abstraction levels (16, 32) exhibit a noticeable shift toward higher SEA scores.}
\label{fig:seva-density}
\end{figure}

\noindent\textbf{Validation of SEA on the SEVA dataset.}
We validate SEA on the SEVA dataset, which contains sketches drawn under four time-constrained abstraction levels of 4, 8, 16, and 32 seconds. We evaluated SEA on six classes shared by CommonSketch and SEVA. Tab.~\ref{tbl:seva-quan} shows that SEA scores generally increase as the abstraction level increases. Fig.~\ref{fig:seva-density} further confirms this trend at the distribution level, where lower abstraction levels 4 and 8 are concentrated in the low-SEA region, whereas higher levels 16 and 32 shift toward higher SEA scores. The qualitative examples in Fig.~\ref{fig:seva-qual} illustrate why this trend emerges. SEA assigns higher scores when recognizability $P$ is preserved while the normalized visual ratio $v$ remains relatively low, indicating that the metric favors sketches that maintain recognizability with fewer visual elements rather than simply rewarding additional detail. These results indicate that SEA serves as an effective measure of abstraction efficiency.

\begin{table}[t]
    \centering
    \scriptsize
    \caption{\textbf{Comparison of open-source VLMs and a proprietary VLM on element-level commonsense VQA.}
    We report Precision, Recall, F1, and Accuracy. Best values are shown in bold, and second-best values are highlighted in gray.}
    \label{tab:vlm_element_eval}
    \begin{tabular}{l|cccc}
        \toprule
        \textbf{Model} & \textbf{Precision  $\uparrow$} & \textbf{Recall  $\uparrow$} & \textbf{F1 Score $\uparrow$} & \textbf{Accuracy  $\uparrow$} \\
        \midrule
        \multicolumn{5}{c}{\textit{Open-source VLMs}} \\
        \midrule
        LLaVA~\cite{liu2023visual}            & 0.749                  & 0.819                  & 0.782                         & 0.706                         \\
        BLIP~\cite{li2022blip}                & 0.731                  & 0.760                  & 0.746                         & 0.666                         \\
        Molmo~\cite{deitke2024molmo}          & 0.798                  & \textbf{0.949}         & \cellcolor{gray!20}0.867      & \cellcolor{gray!20}0.812      \\
        Qwen2.5-VL~\cite{bai2025qwen2}        & \cellcolor{gray!20}0.898 & 0.782                & 0.836                         & 0.802                         \\
        mPLUG-Owl3~\cite{ye2024mplug}         & 0.883                  & 0.686                  & 0.772                         & 0.739                         \\
        InternVL~\cite{zhu2025internvl3}      & 0.804                  & \cellcolor{gray!20}0.890 & 0.845                       & 0.789                         \\
        PaliGemma2~\cite{steiner2024paligemma} & 0.673                 & 0.809                  & 0.735                         & 0.624                         \\
        SmolVLM~\cite{marafioti2025smolvlm}   & 0.808                  & 0.346                  & 0.485                         & 0.526                         \\
        \midrule
        \multicolumn{5}{c}{\textit{Proprietary VLM}} \\
        \midrule
        GPT-4o~\cite{hurst2024gpt}            & \textbf{0.935}         & 0.832                  & \textbf{0.881}                & \textbf{0.855}                \\
        \bottomrule
    \end{tabular}
\end{table}
\begin{table}[t]
\centering
\caption{\textbf{Quantitative results of commonsense element extraction quality.} Open-source models are compared to GPT-4o.}
\scriptsize
\label{tbl:commonsense_llm}
\setlength{\tabcolsep}{6pt}
\begin{tabular}{l|ccc}
\toprule
\textbf{Model} & \textbf{Soft F1 $\uparrow$} & \textbf{CLIP Score $\uparrow$} & \textbf{BERT Score $\uparrow$} \\
\midrule
GPT-OSS 20B~\cite{agarwal2025gpt}                  & \textbf{0.850} & \textbf{0.941} & \textbf{0.813} \\
Qwen2.5 32B~\cite{qwen2.5}                         & 0.833 & 0.930 & 0.797 \\
Llama 3 8B~\cite{grattafiori2024llama}             & 0.828 & 0.935 & 0.802 \\
Mistral 7B~\cite{DBLP:journals/corr/abs-2310-06825}& 0.834 & 0.933 & 0.804 \\
\bottomrule
\end{tabular}
\end{table}

\noindent\textbf{Component analysis for the open-source SEA pipeline.}
While our primary SEA pipeline leverages GPT-4o, we establish a high-fidelity open-source alternative to ensure accessibility. For element extraction ($E$), GPT-OSS 20B is adopted because it closely aligns with GPT-4o's semantic quality across the Soft F1~\cite{corbeil2021assessing}, CLIP Score~\cite{radford2021learning}, and BERT Score~\cite{zhang2019bertscore} metrics. Regarding visual element identification ($V$), we extensively evaluated various VLMs on CommonSketch, leveraging its human-annotated, element-wise ground truth for reliable benchmarking. For models that produce structured outputs, we provide the full list of elements and obtain JSON-formatted predictions. For BLIP and LLaVA, whose outputs are not consistently structured, we instead query each element with a binary \texttt{yes}/\texttt{no} prompt. 
As reported in Tab.~\ref{tab:vlm_element_eval}, GPT-4o achieves the best overall performance. While Molmo attains the highest recall and a strong F1 score, its anomalously high recall (0.949) indicates a severe false-positive bias. This over-prediction undermines the reliability of quantifying visual abstraction, making Qwen2.5-VL a more suitable choice for the SEA metric as it exhibits the highest alignment with GPT-4o's prediction patterns. We therefore adopt Qwen2.5-VL as the open-source VLM for commonsense element VQA and use it in the human assessment in Sec.~\ref{sec:userstudy}. Further analysis of model behavior is provided in the supplementary material.

\begin{figure}[t]
\centering
\includegraphics[width=0.99\linewidth]{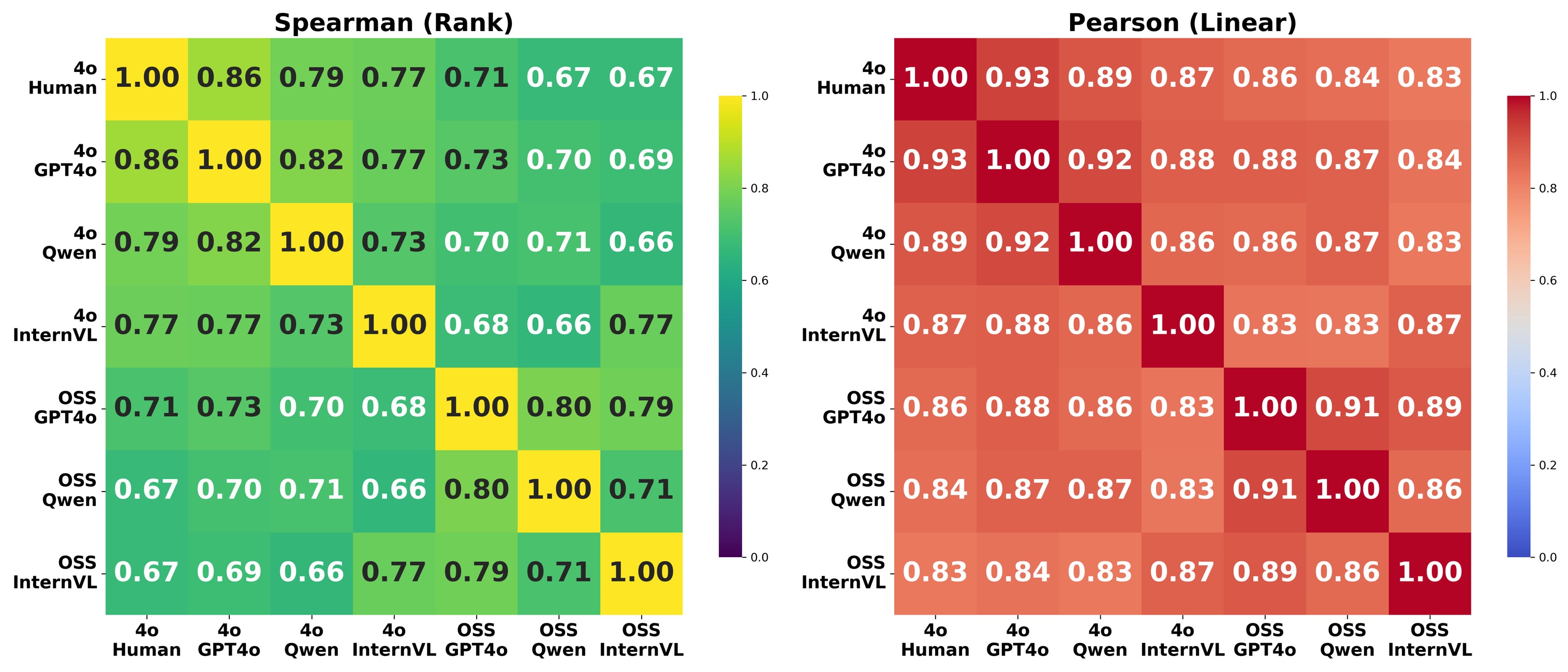}
\caption{\textbf{Consistency and Human Alignment of SEA Scores.} The heatmaps show Spearman (left) and Pearson (right) correlations across different model configurations.}
\label{fig:heatmap_rebuttal}
\end{figure}

\noindent\textbf{Model Robustness and Human Alignment.}
Fig.~\ref{fig:heatmap_rebuttal} further illustrates the consistency of SEA scores across different pre-trained models and their alignment with human judgment. The axes of the heatmaps display model names in two rows: the upper row specifies the LLM used for commonsense extraction, and the lower row specifies the VLM used for element-wise VQA. As observed in the heatmaps of Fig.~\ref{fig:heatmap_rebuttal}, both metrics demonstrate high overall consistency. Specifically, the Spearman correlation is particularly strong between pairs sharing the same LLM backbone, exhibiting a slight decrease when the LLM varies. The Pearson correlation generally maintains high correlation and consistency. Independent of the user study in Sec.~\ref{sec:userstudy}, we conducted an additional evaluation with 27 participants to verify that this inter-model consistency translates to human perception. We selected 8 classes from CommonSketch, presenting participants with 10 ranking questions per class. Each question displayed three sketch images arranged in increasing order of their SEA scores. Across 160 ranking questions evenly split between open and closed models participants showed high agreement rates with the suggested ordering, achieving 88.0\% for the open models and 87.8\% for the closed models, as reported in Tab.~\ref{tab:human_agreement_models}. These results suggest that our proposed SEA metric is robust across different pre-trained model configurations.

\begin{table}[t]
\centering
\scriptsize
\caption{\textbf{Human agreement with the rank-ordering of SEA scores.} Here, \textit{SEA} denotes the closed-source pipeline, while \textit{OpenSEA} denotes the open-source pipeline. We report participant agreement with the score ordering from each pipeline ($N=27$).}
\label{tab:human_agreement_models}
\begin{tabular}{lccc}
\toprule
\textbf{Metric} & \textbf{LLM} & \textbf{VLM} & \textbf{Agreement (\%)} \\
\midrule
SEA      & GPT-4o   & GPT-4o      & 87.8 \\
OpenSEA  & GPT-OSS  & Qwen2.5-VL  & 88.0 \\
\bottomrule
\end{tabular}
\end{table}

\begin{figure}[t]
\centering
\includegraphics[width=0.99\linewidth]{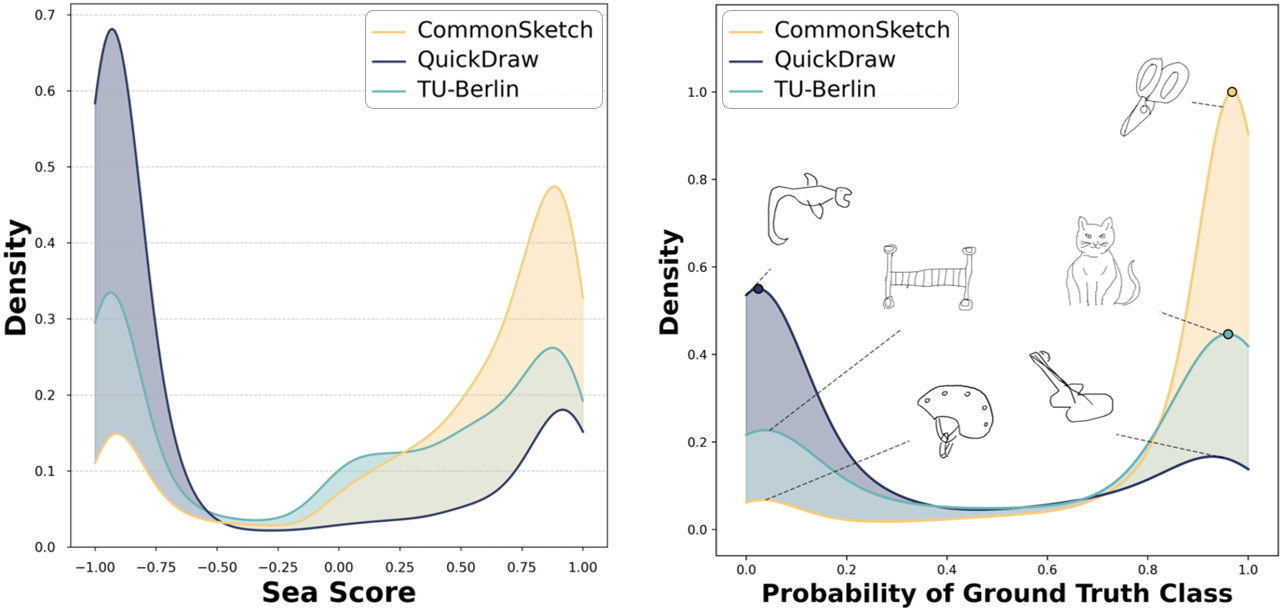}
\caption{\textbf{Comparison of sketch quality across datasets.} We compare CommonSketch with QuickDraw and TU-Berlin using the distributions of SEA scores (left) and predicted probabilities for the ground-truth class (right). In the probability distribution, CommonSketch has $\mu$=0.86, $\sigma$=0.24, and mode=0.97, while QuickDraw has $\mu$=0.29, $\sigma$=0.37, and mode=0.02, and TU-Berlin has $\mu$=0.62, $\sigma$=0.41, and mode=0.96.}
\label{fig:dataset_qual}
\end{figure}

\subsection{Validation of the CommonSketch Dataset}
Fig.~\ref{fig:dataset_qual} compares the distributions of SEA scores and recognition probabilities across CommonSketch, QuickDraw, and TU-Berlin. CommonSketch shows the highest density in the high-SEA region and the strongest concentration of ground-truth label prediction probabilities in $[0.8, 1.0]$, indicating the highest recognizability among the three datasets. TU-Berlin follows, whereas QuickDraw is distributed more heavily in the low-SEA region. Because prediction probability is a key component of SEA, the strong recognizability of CommonSketch naturally leads to higher SEA scores. In contrast, many QuickDraw sketches are incomplete due to its game-based collection protocol, which terminates drawing once the model guesses the label or the time limit is reached. As a result, QuickDraw tends to receive lower recognition probabilities and lower SEA scores. These results demonstrate that CommonSketch combines visually well-formed sketches, reliable captions, and high-quality commonsense element annotations, making it a valuable benchmark for diverse sketch-related tasks, including the element-level VQA evaluation with VLMs in Tab.~\ref{tab:vlm_element_eval}.

\subsection{User Study}\label{sec:userstudy}
\begin{figure}
\centering
\includegraphics[width=0.99\linewidth]{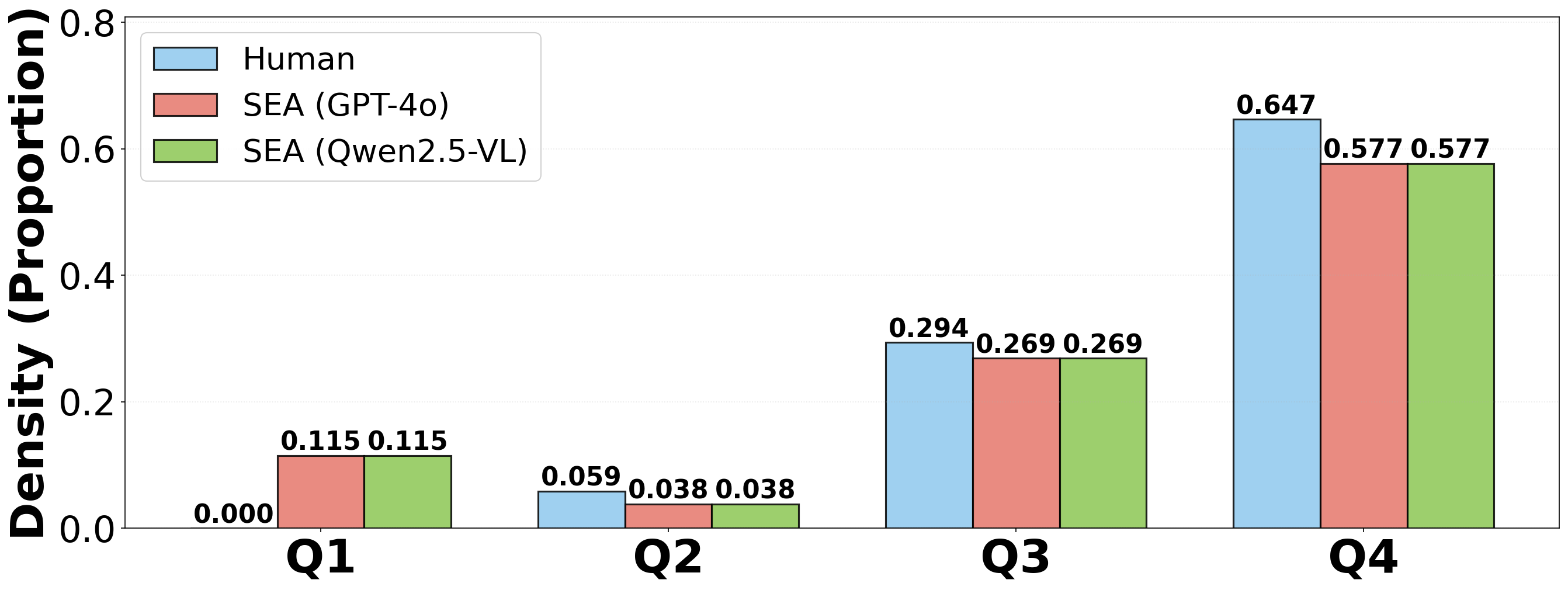}
\caption{\textbf{Visualization of abstraction score distributions on CommonSketch VQA}, comparing human judgments with SEA (GPT-4o, Qwen2.5-VL); bars show the proportion of samples in four equal-width score bins (Q1–Q4).}
\label{jaeyoon4bins}
\end{figure}
To examine the alignment between SEA and human perception, we collected ratings from 37 participants across 88 images and compared the averaged human score distribution against our metric. As shown in Fig.~\ref{jaeyoon4bins}, both the closed-source (GPT-4o~\cite{hurst2024gpt}) and open-source (Qwen2.5-VL~\cite{bai2025qwen2}) SEA configurations exhibit distributions that closely match human judgments across the four score bins, confirming that our metric accurately reflects human consensus on abstraction quality. Additional details of the survey protocol are provided in the supplementary material. 
Fig.~\ref{seva2class} presents qualitative results on unseen object classes outside the 300 CommonSketch classes, evaluating sketches across varying levels of abstraction. For challenging cases like the mosquito, both SEA and human ratings remain consistently negative across all abstraction levels, indicating poor recognizability and unsuccessful abstraction. For the tank class, sketches that lack sufficient detail receive negative scores from both, whereas the recognizable sketch at Level 32 achieves strongly positive scores from both SEA and human evaluators. These results suggest that SEA remains applicable to object classes outside the 300 CommonSketch classes and produces scores consistent with human judgments when the sketch image and class label are available.

\begin{figure}
\centering
\includegraphics[width=0.95\linewidth]{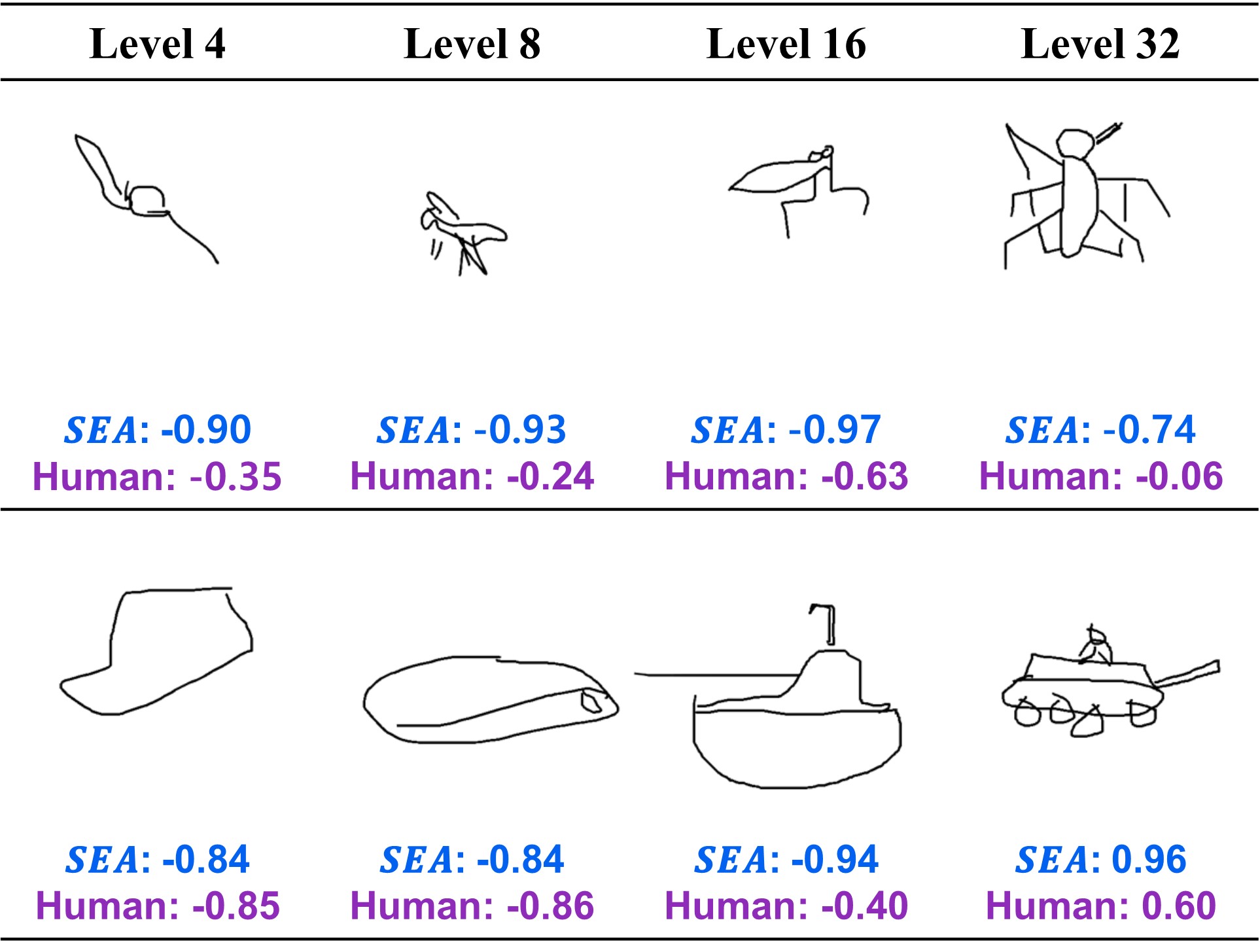}
\caption{\textbf{Comparison of SEA and human scores for SEVA-only sketches.} Top row shows mosquito sketches and bottom row tank sketches.}
\label{seva2class}
\end{figure}
\section{Conclusion}
\label{sec:conclusion}
We present CommonSketch, the first element-level-annotated sketch dataset, and SEA, the novel reference-free metric that quantifies abstraction efficiency as a balance between recognizability and drawing simplicity. Together, they establish the first element-aware benchmark and metric for sketch abstraction. Our experiments show that SEA aligns with human judgments, that CommonSketch reveals systematic limitations in element-level reasoning of vision-language models, and that the two provide a practical foundation for training and evaluating models of sketch understanding and visual abstraction. At the same time, SEA remains dependent on its underlying VQA and classification models, and CommonSketch currently focuses on single-object sketches limited linguistic and cultural coverage; relaxing this model dependence and broadening the dataset’s diversity and scope are important directions for future work.

\section*{Acknowledgment}
This research was supported by the MSIT(Ministry of Science and ICT), Korea, under the ITRC(Information Technology Research Center) support program(IITP-2026-RS-2020-II201789), and the Artificial Intelligence Convergence Innovation Human Resources Development(IITP-2026-RS-2023-00254592) supervised by the IITP(Institute for Information \& Communications Technology Planning \& Evaluation).
This study was approved by the Institutional Review Board (IRB) of Dongguk University (Approval No. DUIRB2025-05-08).

{
    \small
    \bibliographystyle{ieeenat_fullname}
    \bibliography{main}
}
\maketitlesupplementary

\noindent In this supplementary, we provide:
\begin{itemize}
    \item dataset/annotation details and prompts;
    \item extended SEA analyses and qualitative results;
    \item extra comparisons on classifiers, commonsense extraction, and annotators;
    \item user study details.
\end{itemize}

\section{Dataset Analysis Details}
\label{sec:appx-dataset}

\subsection{Per-category class list}
CommonSketch spans 300 object classes grouped into 14 high-level categories. Each category aggregates semantically related concepts to support both category-level and class-level analyses in SEA. \cref{tab:category-classes} reports the complete breakdown, listing the class count and full class names for every category. This category taxonomy was used consistently in dataset construction, commonsense element extraction, and all evaluation protocols.

\begin{table}[h]
\begin{center}
\scriptsize
\setlength{\tabcolsep}{3pt}
\caption{\textbf{Classes by category.}
For each category, we provide the class count and the full set of class labels used in the dataset.}
\label{tab:category-classes}

\begin{tabularx}{\linewidth}{l r X} 
\toprule
Category & \#Classes & Classes \\
\midrule
animal & 61 &
alpaca, ant, bat, bee, bird, boar, butterfly, camel, cat, caterpillar,
chameleon, cow, crab, crocodile, deer, dog, dolphin, dragonfly, duck, elephant,
feather, fish, flamingo, frog, giraffe, goose, hedgehog, hippopotamus, horse,
jellyfish, kangaroo, koala, lion, lobster, mole, monkey, moose, mouse, octopus,
owl, panda, parrot, peacock, penguin, rabbit, rooster, scorpion, seahorse,
shark, sheep, sloth, snail, snake, spider, squid, squirrel, swan, tiger, turtle,
whale, zebra \\
body part & 7 &
ear, eye, foot, hand, mouth, nose, tooth \\
clothing & 8 &
belt, bowtie, crown, flip flops, hat, shoe, sock, t shirt \\
container & 9 &
backpack, basket, bucket, envelope, mailbox, present, purse, suitcase,
wine bottle \\
electronic device & 22 &
alarm clock, calculator, camera, cell phone, charger, computer, fan,
headphones, ipod, keyboard, laptop, megaphone, microphone, microwave, oven,
radio, robot, satellite, telephone, television, toaster, walkie talkie \\
food & 28 &
apple, asparagus, banana, bread, broccoli, cake, carrot, cookie, cupcake, donut,
garlic, grapes, hamburger, hot dog, ice cream cone, lollipop, mushroom,
noodle, onion, peanut, pear, pineapple, pizza, pretzel, pumpkin, sandwich,
strawberry, watermelon \\
furniture & 25 &
bathtub, bed, book, calendar, candle, ceiling fan, chandelier, couch, crayon,
door, drawer, fireplace, floor lamp, hourglass, lantern, light bulb, map,
marker, paintbrush, paper clip, pencil, stairs, table, toilet, vase \\
icon & 13 &
angel, diamond, dragon, jack o lantern, mermaid, mona lisa, patrick star,
santa claus, skull, snowman, sponge bob, stop sign, teddy bear \\
musical instrument & 11 &
bell, cello, clarinet, drums, guitar, harp, piano, saxophone, trombone, trumpet,
violin \\
nature & 14 &
bamboo, beach, bush, cactus, cloud, clover, dandelion, flower, leaf, moon,
palm tree, rainbow, sun, tree \\
sports equipment & 14 &
barbell, baseball, baseball bat, basketball, dumbbell, golf club, helmet,
parachute, roller skate, skateboard, snorkel, soccer ball, table tennis,
tennis racquet \\
structure & 28 &
arch of triumph, barn, bench, big ben, bridge, campfire, castle, church,
eiffel tower, fence, ferris wheel, fire hydrant, fountain, hospital, house,
igloo, leaning tower of pisa, lighthouse, moai stone, pyramids of giza,
roller coaster, skyscraper, sphinx, statue of liberty, stonehenge,
streetlight, traffic light, windmill \\
tool & 37 &
axe, bandage, binoculars, boomerang, bottlecap, broom, cannon, comb, compass,
drill, fork, frying pan, grenade, hammer, key, knife, ladder, lighter, matches,
mug, pipe, rake, rifle, saw, scissors, screwdriver, shovel, spoon, stethoscope,
sword, syringe, teapot, tent, toothbrush, toothpaste, umbrella, wine glass \\
vehicle & 23 &
airplane, ambulance, bicycle, blimp, bulldozer, bus, canoe, car, cruise ship,
flying saucer, helicopter, hot air balloon, motorcycle, pickup truck, rocket,
sailboat, space shuttle, submarine, tractor, train, truck, van, wheel \\
\bottomrule
\end{tabularx}
\end{center}
\end{table}
\subsection{Per-class element list}
\label{sec:per-class-elements}

~\cref{tab:per-class-elements-merged} provides the full per-class commonsense element lists extracted by GPT-4o~\cite{hurst2024gpt} and GPT-OSS~\cite{agarwal2025gpt}, which form CommonSketch’s class-wise commonsense database and are used by SEA for element-level presence checking and abstraction analysis. We also replicate the extraction with open-source MLLMs (GPT-OSS 20B, Qwen-2.5 32B~\cite{qwen2.5}, Mistral 7B~\cite{jiang2023mistral7b}, and Llama 3 8B~\cite{grattafiori2024llama}) to assess reproducibility, but omit the full Qwen, Mistral, and Llama lists here to avoid an overly long table; all model-specific lists will be released with the CommonSketch dataset. We report elements in their original surface forms (including casing and hyphen/underscore variants) without normalization to preserve raw model outputs.\\

\noindent\begin{minipage}{\columnwidth}
\captionsetup{
  type=table,
  font=small,
  width=\columnwidth,
  singlelinecheck=false
}
\scriptsize
\setlength{\tabcolsep}{3pt}
\renewcommand{\arraystretch}{0.95}

\captionof{table}{\textbf{Per-class element list (sample).}
Sample rows from \cref{tab:per-class-elements-merged} are shown to illustrate our per-class commonsense annotations.
Elements are listed as extracted: black denotes overlap between GPT-4o and GPT-OSS, \textcolor{red}{red} indicates GPT-4o-only elements, and \textcolor{blue}{blue} indicates GPT-OSS-only elements.
The 4o/OSS column reports the number of elements extracted by each model.}

\label{tab:per-class-elements-sample}

\begin{tabularx}{\columnwidth}{@{}l l r >{\raggedright\arraybackslash}X@{}}
\toprule
Category & Class & 4o/OSS & Elements \\
\midrule

\multirow[t]{3}{*}{animal}
& alpaca & 13/11 &
body, ears, eyes, head, legs, tail, \textcolor{red}{feet}, \textcolor{red}{fleece},
\textcolor{red}{hooves}, \textcolor{red}{muzzle}, \textcolor{red}{neck},
\textcolor{red}{nostrils}, \textcolor{red}{smile}, \textcolor{blue}{fur\_lines},
\textcolor{blue}{motion\_lines}, \textcolor{blue}{mouth},
\textcolor{blue}{nose}, \textcolor{blue}{whisker\_lines} \\

& ant & 11/9 &
abdomen, antennae, head, legs, mandibles, thorax, \textcolor{red}{compound eyes},
\textcolor{red}{jointed legs}, \textcolor{red}{mouth}, \textcolor{red}{petiole},
\textcolor{red}{stinger}, \textcolor{blue}{body}, \textcolor{blue}{eyes},
\textcolor{blue}{segment\_lines} \\

& bat & 14/12 &
body, ears, eyes, head, mouth, nose, tail, wings, \textcolor{red}{feet},
\textcolor{red}{fingers}, \textcolor{red}{fur}, \textcolor{red}{legs},
\textcolor{red}{teeth}, \textcolor{red}{wing membrane},
\textcolor{blue}{claws}, \textcolor{blue}{fur\_lines},
\textcolor{blue}{motion\_lines}, \textcolor{blue}{wing\_vein\_lines} \\

\bottomrule
\end{tabularx}
\end{minipage}

\clearpage
\onecolumn

\subsection{Additional Sketch Examples}

\begin{figure}[H]
\centering
\includegraphics[width=0.99\linewidth]{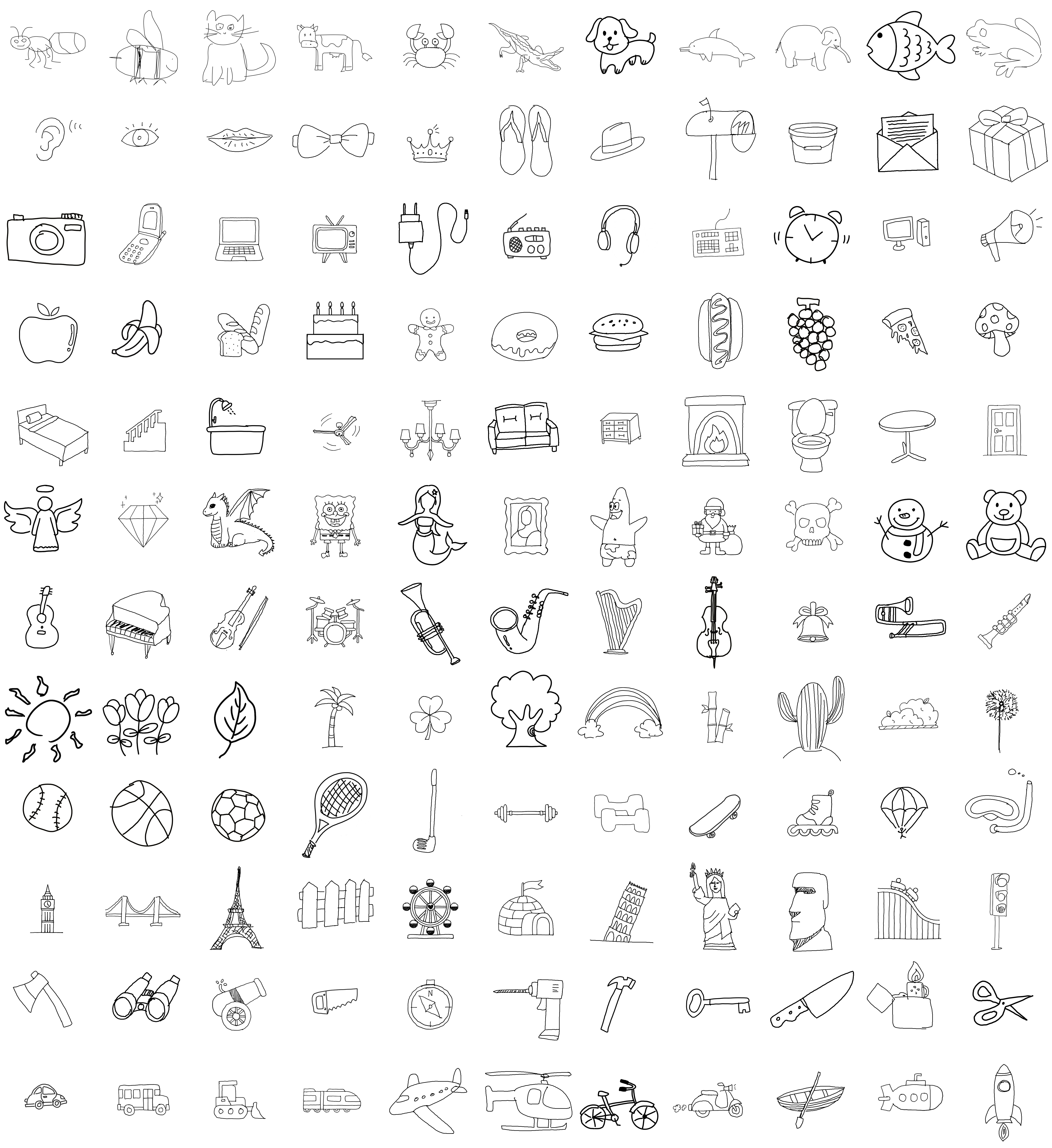}
\caption{\textbf{Additional sketch examples from CommonSketch.} The first three rows are dedicated to the \texttt{animal} category, showing a broad slice of its classes. The fourth row groups representative sketches from \texttt{body\_part}, \texttt{clothing}, and \texttt{container}. Each remaining row corresponds to one category (\texttt{electronic\_device}, \texttt{food}, \texttt{furniture}, \texttt{icon}, \texttt{musical\_instrument}, \texttt{nature}, \texttt{sports\_equipment}, \texttt{structure}, \texttt{tool}, \texttt{vehicle}) in order.}
\label{fig:appx-category-sketch-mosaic}
\end{figure}

\clearpage
\twocolumn

\subsection{Element Frequency Analysis}
\label{sec:element-frequency}
We provide statistics on how commonsense elements are distributed across classes and categories. For each element, we treat its presence as a binary attribute at the class level and compute both its global frequency across all classes and its category-wise frequency within each category. To quantify how strongly an element is associated with a category relative to its overall prevalence, we use a lift score, following its standard use in association rule mining for interpretability and comparability~\cite{agrawal1993mining,jiawei2006data}:
\begin{equation*}
\mathrm{lift}(e,c)=\frac{P(e\mid c)}{P(e)}
=\frac{n(e,c)/n(c)}{n(e)/N},
\end{equation*}
where $n(e,c)$ is the number of classes in category $c$ that contain element $e$, $n(c)$ is the number of classes in category $c$, $n(e)$ is the number of classes containing $e$ across the full dataset, and $N$ is the total number of classes. Using this formulation, we rank elements within each category by lift to identify elements that are relatively overrepresented in that category compared with the dataset-wide base rate. \cref{tab:global-top-elements} reports, for each semantic category, the three elements with the highest lift. To avoid unstable estimates from extremely rare elements, we include only elements with $n_{\text{in\_cat}} \ge 3$. Complementarily, \cref{tab:global-lift-rank} reports the most frequent elements in the full dataset, ranked by the number of classes in which each element appears. This table summarizes the global prevalence of recurring visual components independently of category-level enrichment.
\begin{table}[t]
\centering
\small
\setlength{\tabcolsep}{4pt}
\caption{\textbf{Category-wise frequent elements.}
For each category, we list the three elements with the highest lift.
Columns report the per-category class count $n_{\text{in\_cat}}$, within-category coverage $p_{\text{cat}}$, and lift.
Only elements with $n_{\text{in\_cat}} \ge 3$ are included.}
\label{tab:global-top-elements}
\begin{tabular}{l r l r r r}
\toprule
Category & Rank & Element & $n_{\text{in\_cat}}$ & $p_{\text{cat}}$ & Lift \\
\midrule
animal             & 1 & abdomen      &  4 & 0.07 & 4.92 \\
animal             & 2 & antennae     &  6 & 0.10 & 4.92 \\
animal             & 3 & beak         & 11 & 0.18 & 4.92 \\
container          & 1 & handle       &  5 & 0.56 & 4.76 \\
container          & 2 & body         &  6 & 0.67 & 1.74 \\
electronic device & 1 & display      &  4 & 0.18 & 13.64 \\
electronic device & 2 & microphone   &  4 & 0.18 & 13.64 \\
electronic device & 3 & screen       &  5 & 0.23 & 13.64 \\
food               & 1 & slice        &  5 & 0.18 & 10.71 \\
food               & 2 & bite mark    &  3 & 0.11 & 10.71 \\
food               & 3 & seeds        &  4 & 0.14 &  8.57 \\
furniture          & 1 & tip          &  4 & 0.16 &  4.36 \\
furniture          & 2 & frame        &  3 & 0.12 &  3.60 \\
furniture          & 3 & base         &  5 & 0.20 &  2.40 \\
icon               & 1 & arms         &  4 & 0.31 &  6.59 \\
icon               & 2 & mouth        &  4 & 0.31 &  2.10 \\
icon               & 3 & legs         &  5 & 0.38 &  1.96 \\
musical instrument& 1 & bridge       &  3 & 0.27 & 27.27 \\
musical instrument& 2 & bell         &  4 & 0.36 & 21.82 \\
musical instrument& 3 & keys         &  3 & 0.27 & 20.45 \\
nature             & 1 & leaves       &  3 & 0.21 & 16.07 \\
nature             & 2 & leaf         &  3 & 0.21 &  8.04 \\
nature             & 3 & base         &  3 & 0.21 &  7.14 \\
sports\_equipment  & 1 & grip         &  3 & 0.21 &  8.04 \\
structure          & 1 & tower        &  4 & 0.14 & 10.71 \\
structure          & 2 & flag         &  3 & 0.11 &  6.43 \\
structure          & 3 & windows      &  3 & 0.11 &  5.36 \\
tool               & 1 & angle        &  3 & 0.08 &  8.11 \\
tool               & 2 & edge         &  3 & 0.08 &  8.11 \\
tool               & 3 & rivet        &  3 & 0.08 &  8.11 \\
vehicle            & 1 & exhaust pipe &  7 & 0.30 & 13.04 \\
vehicle            & 2 & bumper       &  4 & 0.17 & 13.04 \\
vehicle            & 3 & cabin        &  4 & 0.17 & 13.04 \\
\bottomrule
\end{tabular}
\end{table}

\begin{table}[t]
\centering
\small
\setlength{\tabcolsep}{4pt}
\caption{\textbf{Global element frequency ranking.} We report the 20 elements that appear in the largest number of classes across all 300 classes.}

\label{tab:global-lift-rank}

\begin{minipage}{0.48\linewidth}
\centering
\begin{tabular}{lrr}
\toprule
Element & $n_{\text{classes\_global}}$ & Rank \\
\midrule
body     & 115 &  1 \\
head     &  67 &  2 \\
legs     &  59 &  3 \\
eyes     &  57 &  4 \\
tail     &  52 &  5 \\
mouth    &  44 &  6 \\
handle   &  35 &  7 \\
ears     &  29 &  8 \\
base     &  25 &  9 \\
nose     &  24 & 10 \\
\bottomrule
\end{tabular}
\end{minipage}
\hfill
\begin{minipage}{0.48\linewidth}
\centering
\begin{tabular}{lrr}
\toprule
Element & $n_{\text{classes\_global}}$ & Rank \\
\midrule
neck     &  24 & 11 \\
feet     &  20 & 12 \\
stem     &  15 & 13 \\
water    &  14 & 14 \\
wings    &  14 & 15 \\
arms     &  14 & 16 \\
teeth    &  13 & 17 \\
nostrils & 12 & 18 \\
claws    & 12 & 19 \\
tip      & 11 & 20 \\
\bottomrule
\end{tabular}
\end{minipage}

\end{table}

\subsection{Cross-Dataset Sketch Quality Comparison}
\label{sec:cross-dataset-quality}

To contextualize the sketch quality of CommonSketch relative to widely used benchmarks, we compare it with QuickDraw~\cite{ha2018neural} and TU-Berlin~\cite{eitz2012hdhso} using a shared evaluation protocol. We select 14 classes, one from each CommonSketch category, restricting the comparison to classes present in all three datasets. For each sketch, we compute the \emph{Probability of the Ground Truth Class} using three classification models, CLIP~\cite{radford2021learning}, OpenCLIP~\cite{ilharco2021openclip}, and CoCa~\cite{yu2022coca}, and average their outputs to obtain a recognizability score. \cref{fig:all_kde_plots} shows kernel density estimates (KDEs) of these scores for each class, allowing comparison at the distribution level rather than through a single summary statistic. Higher and more concentrated densities indicate more consistently recognizable sketches, whereas broader or lower-probability distributions suggest greater ambiguity. We also show representative sketches near the \emph{mode} of each dataset's KDE to provide a visual reference for the typical quality level. Across the 14 shared classes, CommonSketch exhibits competitive or stronger modes overall, supporting the quality of our data collection pipeline and its suitability for SEA-based abstraction and element-level analysis.

\subsection{Element Annotation Details}
\label{sec:appx-vqa}

We provide further details on the refinement of commonsense elements and the strict criteria applied during the annotation process, supplementing the methodologies described in CommonSketch.

\paragraph{Disambiguation of Symbolic and Anatomical Features.}
During the refinement of raw elements generated by LLMs, a key challenge was distinguishing between realistic physical attributes and stylized or anthropomorphic depictions common in sketch representations. For instance, while a butterfly biologically possesses a \textit{proboscis}, it is frequently depicted in sketches with a simple, smiling \textit{mouth}. To accommodate both anatomical accuracy and sketch-style abstraction, we separated these features into distinct annotation categories. This distinction was similarly applied to other classes such as ant, bee, and octopus. This separation ensures that the model is evaluated on what is actually drawn rather than what is biologically expected.

\paragraph{Strict Criteria for Visual Presence.}
To establish ground truth labels, we applied clear guidelines to determine the presence of an element, operating under the principle that visual existence takes precedence over biological facts.

Certain textural elements, such as \textit{fur} for animals or \textit{feathers} for birds, are intrinsically present in the real-world subjects. However, in our annotation, these elements were marked as present only if the sketch explicitly contained additional strokes or shading to depict texture. If an animal was drawn with a simple, smooth outline without specific textural details, attributes like fur or feathers were labeled as absent, even if they naturally belong to the subject. This strict criterion enables us to distinguish whether the model relies on actual visual evidence or merely depends on background knowledge.

\section{SEA: Sketch Evaluation metric for Abstraction efficiency}
\label{sec:appx-metric}
\subsection{Basic properties of SEA}
\paragraph{Notation.}
We denote by $E$ the number of class-defining commonsense elements that are available for a given class,
by $V$ the number of visual elements that are actually rendered in a sketch,
and by $P \in [0,1]$ the predicted probability of the ground-truth class from the recognition model.
We consider the domain
\[
E \in \mathbb{N}_{\ge 1}, 
\qquad
0 < V \le E,
\qquad
0 < P < 1,
\]
and write $v = V / E$ for the visual ratio.
Intuitively, $E$ captures semantic coverage, $v$ measures how much of the available visual information is expressed, and $P$ measures recognizability.

In implementation, we apply a small numerical clipping
$V \leftarrow \min(\max(V,0),E)$ and $P \leftarrow \min(\max(P,\varepsilon),1-\varepsilon)$
with $\varepsilon \approx 10^{-6}$, but this does not affect the analytical properties discussed below.

\paragraph{SEA formulation.}
Given $(E,V,P)$, the SEA score $S(E,V,P)$ is defined as
\[
S(E,V,P) \;=\; \tanh\bigl(\alpha \, Z(E,V,P)\bigr),
\]
where $\alpha>0$ is a fixed scale parameter and
\[
Z(E,V,P)=\text{reward}(E,V,P)-\text{penalty}(E,V,P).
\]
The reward term encourages efficient abstraction,
while the penalty term suppresses both excessive visual expression and low recognizability.
In other words, the reward favors sketches that remain recognizable with minimal visual expression,
and the penalty discourages sketches that either use too much visual detail or fail to be recognized.

We first define the visual ratio
\[
v \;=\; \frac{V}{E}.
\]
The \emph{economy of expression} is given by
\[
u(v) \;=\; \log\!\left(\frac{1+\delta}{v+\delta}\right),
\]
where $\delta > 0$ is a small numerical constant.
This term is positive and monotonically decreasing in $v$:
it is large when the sketch uses few visual elements ($v$ small) and approaches zero as $v \to 1$.
Thus, for fixed recognizability, sketches with more visual representation efficiency (smaller $v$) receive higher $u$.

The \emph{centered gate} compares visual expression and recognizability:
\[
g(P,v) \;=\; \tanh\!\left(\tfrac{1}{2}\beta \,\log\!\frac{P+\delta}{v+\delta}\right),
\]
where $\beta>0$ controls the sharpness of the transition.
We have $g(P,v)=0$ when $v = P$;
if $v<P$ (the sketch is more recognizable than its level of visual expression would suggest), then $g(P,v)>0$ and the reward is amplified;
if $v>P$ (the sketch is more detailed than necessary for its recognizability), then $g(P,v)<0$ and the reward is attenuated.
When $v$ and $P$ are well aligned, the gate remains near zero and does not dramatically amplify or suppress the reward.

The reward term combines economy of expression, the centered gate, and recognizability:
\[
\text{reward}(P,v)
\;=\;
P^{\gamma} \, u(v) \, g(P,v),
\]
where $\gamma>0$ controls how strongly the reward is guided by recognizability.
As a result, sketches with high recognizability $P$, strong economy of expression $u$, and a positive centered gate $g$ receive large positive reward.

\begin{figure}[t]
\centering
\begin{subfigure}[t]{0.45\columnwidth}
  \centering
  \includegraphics[width=\linewidth]{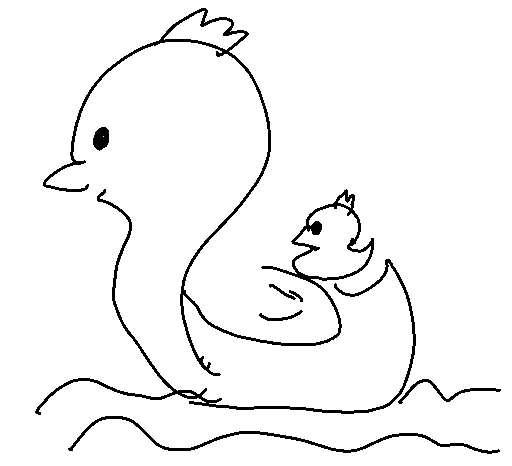}
  \caption{A detailed sketch with relatively high visual ratio and moderate recognizability.}
\end{subfigure}
\hfill
\begin{subfigure}[t]{0.45\columnwidth}
  \centering
  \includegraphics[width=\linewidth]{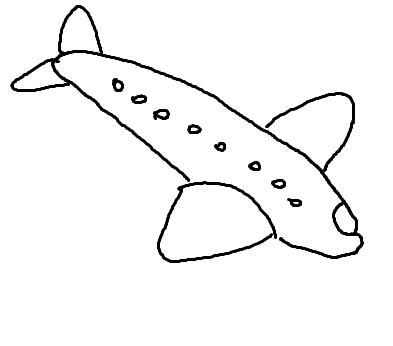}
  \caption{A visually sparse sketch with low recognizability despite a moderate visual ratio.}
\end{subfigure}

\vspace{4pt}

\footnotesize
\setlength{\tabcolsep}{3pt}
\renewcommand{\arraystretch}{1.0}
\begin{tabular}{p{2.5cm}cccc}
\toprule
\textbf{Example} & \shortstack[c]{\textbf{Visual}\\\textbf{ratio ($v$)}} & \shortstack[c]{\textbf{Probability}\\\textbf{($P$)}} & \shortstack[c]{\textbf{SEA score}\\\textbf{(w/ $g$)}} & \shortstack[c]{\textbf{SEA score}\\\textbf{(w/o $g$)}} \\
\midrule
(a) Detailed sketch & 0.69 & 0.63 & -0.43 & +0.17 \\
(b) Low-$P$ sketch   & 0.60 & 0.18 & -0.93 & +0.03 \\
\bottomrule
\end{tabular}

\vspace{2pt}
\caption{\textbf{Effect of the gate function in SEA scoring.}
Two representative cases illustrate how the gate term changes the final SEA score. The gated formulation assigns substantially lower scores in both cases, showing that it penalizes overly detailed or weakly recognizable sketches more strongly.}
\label{fig:gate_examples}
\end{figure}

The reward term is designed to (i) reward lower $v$ with diminishing returns via a log-ratio, (ii) penalize misalignment between $v$ and $P$ using the centered gate $g(\cdot)$, and (iii) use a bounded, differentiable $\tanh$ gate to prevent extreme values from dominating. Fig.~\ref{fig:gate_examples} shows that removing $g(\cdot)$ inflates scores in both (a) detailed and (b) low-$P$ cases, confirming the necessity of $g(\cdot)$. Also, SEA quantifies excessive detail at the element level as over-expression rather than identifying specific strokes as responsible.

The penalty term consists of two parts:
\[
\text{penalty}(P,v)
\;=\;
\lambda \, v^{\eta} (1-P)^{k}
\;+\;
\tau \,(1-P)^{r},
\]
where $\lambda>0$ scales the penalty on excessive visual expression,
$\eta \in (0,1)$ controls the curvature with respect to $v$ (reducing sensitivity at very high usage),
$k>0$ and $r>0$ control the sensitivity to low recognizability,
and $\tau>0$ sets the base penalty for low $P$ regardless of $v$.
The first term penalizes sketches that use many visual elements ($v$ large) when $P$ is low,
while the second term ensures that sketches with very low recognizability are penalized even if they are visually sparse.
Thus, the penalty discourages both excessive visual expression and failure to be recognized.
\cref{hyper_abl} provides an ablation over these hyperparameters.

\paragraph{Boundedness.}
By construction, SEA is a bounded score.
For any admissible $(P,v)$, the inner quantity $Z(P,v)$ is real-valued, and $\alpha>0$ is a constant.
Since the hyperbolic tangent satisfies $-1 < \tanh(x) < 1$ for all real $x$, it follows that
\[
-1 < S(P,v) = \tanh\bigl(\alpha Z(P,v)\bigr) < 1
\]
for all admissible $(P,v)$.
This boundedness makes SEA directly comparable across classes and datasets
and avoids scale issues when aggregating scores.

\paragraph{Continuity and differentiability.}
We now formalize the smoothness of SEA.
On the domain $E \ge 1$, $0<V\le E$, and $0<P<1$,
we have $v \in (0,1]$ and $P \in (0,1)$.
Since $\delta>0$, the arguments $v+\delta$ and $P+\delta$ are strictly positive.
The logarithm, powers $x\mapsto x^{\gamma}$, $x\mapsto x^{k}$, $x\mapsto x^{r}$,
and the hyperbolic tangent are all smooth on $(0,\infty)$ (and on $\mathbb{R}$ for $\tanh$).
Therefore $u(v)$ and $g(P,v)$ are smooth in $(V,P)$,
and so are the reward and penalty.
The function $Z(E,V,P)$ is a sum of these smooth terms, hence smooth in $(V,P)$,
and $S(E,V,P) = \tanh(\alpha Z(E,V,P))$ is a smooth composition of smooth functions.
In particular, $u(E,V)$, $g(P,v)$, $\text{reward}(E,V,P)$,
and $\text{penalty}(E,V,P)$ are continuous in $(E,V,P)$ and differentiable in $(V,P)$,
and the same holds for $Z(E,V,P)$ and $S(E,V,P)$.

Smoothness is beneficial when SEA is used not only as an evaluation metric but also
as a reward or critic in optimization, e.g., for training sketch generative models.

\paragraph{Extreme cases.}
We summarize the qualitative behavior of SEA in several extreme regimes, which follow directly from the definitions of $u$, $g$, the reward, and the penalty.

First, consider \emph{unrecognizable sketches}, where $P \to 0$.
As $P$ approaches zero, the factor $P^{\gamma}$ in the reward term tends to zero, so the reward vanishes.
At the same time, $(1-P)^{k}$ and $(1-P)^{r}$ both approach $1$,
so the penalty converges to $\lambda v^{\eta} + \tau > 0$.
Thus, $Z(E,V,P)$ tends to $-(\lambda v^{\eta} + \tau)$,
and $S(E,V,P)$ moves toward the lower end of its range.
This reflects the design choice that sketches not recognized by the classifier
are treated as abstraction failures regardless of visual detail.

Second, consider \emph{efficient abstraction}, where $P \to 1$ and $v \ll 1$.
As $P$ approaches one, both $(1-P)^{k}$ and $(1-P)^{r}$ tend to zero, so the penalty vanishes.
For very small $v$, the economy of expression $u(E,V) = \log\bigl((1+\delta)/(v+\delta)\bigr)$ is large and positive.
Since $P$ is close to one, we typically have $P>v$, implying
$\log\bigl((P+\delta)/(v+\delta)\bigr)>0$ and hence $g(P,v)>0$.
Consequently, the reward becomes large and positive, $Z(E,V,P) > 0$,
and $S(E,V,P)$ moves toward the upper end of its range.
This corresponds to sketches that remain highly recognizable while using very few strokes.

Finally, consider \emph{over-detailed sketches}, where $v \to 1$.
When the sketch renders nearly all available elements, the visual ratio $v$ approaches one,
and the economy of expression $u(E,V)$ approaches
\[
\log\!\left(\frac{1+\delta}{1+\delta}\right) = 0,
\]
so the reward becomes small.
In contrast, the first term of the penalty approaches $\lambda (1-P)^{k}$,
which is strictly positive for any $P<1$, and the second term $\tau(1-P)^{r}$ is also non-negative.
Therefore, $Z(E,V,P)$ decreases and $S(E,V,P)$ is reduced even when $P$ is high.
This reflects our design choice that overly detailed sketches exhibit lower abstraction
and should therefore receive lower SEA scores.

Together, these properties show that SEA is a bounded, smooth scoring function that rewards semantically efficient sketches with high recognizability and minimal visual expression, while penalizing both unrecognizable and unnecessarily detailed drawings.
In the next section, we analyze how these qualitative behaviors are enforced through
monotonicity and consistency constraints on the partial derivatives of $S(E,V,P)$.

\subsection{Analysis of SEA under constraint conditions}\label{2_2}

\paragraph{Constraint conditions.}
In this section, we formalize the global constraints that SEA is designed to satisfy in the interior of the domain.
The formulation reflects four main conditions: recognizability monotonicity, visual representation efficiency, the failure region, and the efficient-abstraction region.

First, SEA is designed to be monotone with respect to recognizability.
When the amount of visual content $V$ and the semantic capacity $E$ are fixed within a typical range, increasing the recognizability score $P$ should not reduce the SEA score.
This reflects the principle that sketches with higher recognizability, and thus stronger semantic alignment, should not be penalized.

Second, the metric promotes visual representation efficiency.
When a sketch exhibits low recognizability, increasing the visual ratio $v$ should not increase the score; additional strokes cannot compensate for semantic failure.
Conversely, when recognizability is sufficiently high, SEA encourages moderate visual usage.
The score increases with visual representation only up to an optimal usage level $v^\ast(E,P)$, after which excessive detail is penalized.
This ensures that SEA rewards sketches that maintain recognizability using only the essential visual elements.

The third condition defines the failure region:
all sketches whose recognizability falls below a threshold $P_{\mathrm{fail}}$ must receive non-positive SEA scores, regardless of visual representation.
This ensures that a sketch that fails to convey its semantic identity is treated as an abstraction failure, even if it uses very few strokes.

The final condition defines the efficient-abstraction region.
Sketches that achieve high recognizability with economical visual information should receive strictly positive SEA scores.
Thus, for recognizability values above a threshold $P_{\mathrm{good}}$ and visual ratio below an efficiency boundary $v_{\mathrm{eff}}$, the score must be positive.
This region captures the central goal of SEA: rewarding sketches that preserve semantic content with minimal visual expression.

Together, these four conditions govern the global behavior of SEA, determining how recognizability, visual usage, and semantic efficiency interact to produce the final score.

\paragraph{Derivative analysis.}
To understand how these constraints arise from the functional form of SEA, we analyze the partial derivatives of the score, where \(v = V/E\). Recall that
\[
\begin{aligned}
S(E,V,P) &= \tanh\bigl(\alpha Z(E,V,P)\bigr), \\
Z(E,V,P) &= \mathrm{reward}(E,V,P) - \mathrm{penalty}(E,V,P),
\end{aligned}
\]
with
\[
\begin{aligned}
u(E,V) &= \log\!\left(\frac{1+\delta}{v+\delta}\right), \\
g(P,V,E) &= \tanh\!\left(\tfrac{1}{2}\beta \log\!\frac{P+\delta}{v+\delta}\right), \\
\mathrm{reward}(E,V,P) &= P^{\gamma} \, u(E,V) \, g(P,V,E), \\
\mathrm{penalty}(E,V,P) &= \lambda v^{\eta}(1-P)^k + \tau(1-P)^r.
\end{aligned}
\]

Because the hyperbolic tangent is strictly increasing, the sign of the derivatives of \(S\) matches that of \(Z\). The derivative with respect to recognizability is
\[
\frac{\partial S}{\partial P}
=
\alpha \bigl(1-\tanh^2(\alpha Z)\bigr)\frac{\partial Z}{\partial P}.
\]
An explicit expansion of \(\partial Z / \partial P\) shows that it consists of two positive penalty-related terms and two reward-related terms. Under the hyperparameter setting used in our experiments, the reward-related terms are non-negative over the classifier's operating range, while the penalty-related terms remain strictly positive for all \(0<P<1\). Numerical evaluation over \(P \in [0.1,0.99]\), \(v \in [0.05,1.0]\), and \(E \in \{4,8,16,32\}\) confirms that \(\partial Z/\partial P \ge 0\) throughout this region. Consequently, \(\partial S/\partial P \ge 0\), establishing recognizability monotonicity in practice.

For the derivative with respect to the visual ratio \(v\), we obtain
\[
\frac{\partial S}{\partial v}
=
\alpha \bigl(1-\tanh^2(\alpha Z)\bigr)\frac{\partial Z}{\partial v},
\]
where \(\partial Z/\partial v\) decomposes into reward-related terms involving \(\partial u/\partial v\) and \(\partial g/\partial v\), and a penalty-derived term \(\lambda \eta v^{\eta-1}(1-P)^k\). When recognizability is low, the factor \(P^{\gamma}\) suppresses the reward derivatives, leaving the penalty derivative dominant and strictly negative, which yields \(\partial S/\partial v \le 0\). This enforces the principle that additional strokes do not help a sketch that is already semantically unrecognizable.

When recognizability is high, $P^{\gamma}$ is large enough for the reward derivatives to compete with the penalty derivative. For small $v$, the efficiency term $u(E,V)$ is large and positive, and $g(P,v)$ is typically positive because recognizability exceeds usage. Hence $\partial Z/\partial v > 0$ and the score increases with visual usage. As $v$ grows, $u$ decreases and the penalty derivative increases, eventually causing $\partial Z/\partial v$ to become negative. This produces an interior optimum $v^\ast(E,P)$ where the score is maximized, and ensures that excessively detailed sketches are penalized.

Taken together, these derivative properties show that the SEA formulation, combined with its chosen hyperparameters, enforces the global constraint conditions described above.  
The behavior observed in the extreme cases of Section~S.1 extends across the full interior of the domain, ensuring coherent and consistent scoring of sketches according to abstraction efficiency.

\subsection{Ablation Studies on Hyperparameters}
\paragraph{Default hyperparameter setting.}\label{hyper_abl}
The SEA score is determined by a compact set of hyperparameters that control the scale, sharpness, and relative strength of the reward and penalty components. Throughout all main-paper experiments, we use the following default values:
\[
\begin{aligned}
\alpha = 2.2, 
\beta = 8.0,
\lambda = 1.0,
\eta = 0.8,
k = 2.3, \\
\tau = 0.4,
r = 1.7,
\gamma = 1.7,
\delta = 10^{-6}.
\end{aligned}
\]
This configuration produces a stable and interpretable scoring surface. Efficiently abstracted sketches, which show high recognizability with minimal visual representation, tend to obtain positive scores. In contrast, sketches that lack recognizability or contain unnecessary visual details tend to obtain negative scores. These hyperparameters therefore define the overall operating regime of SEA and serve as the baseline for subsequent ablation studies.

\paragraph{One-dimensional sweeps.}
To illustrate how SEA responds to variations in visual representation, we perform one-dimensional sweeps over the normalized visual representation $v$. Since $v$ is normalized by the number of available elements, the choice of $E$ does not affect the qualitative shape of the SEA curve. We therefore fix $E=10$ for all sweeps.

\begin{figure}[t]
\centering
\includegraphics[width=0.95\linewidth]{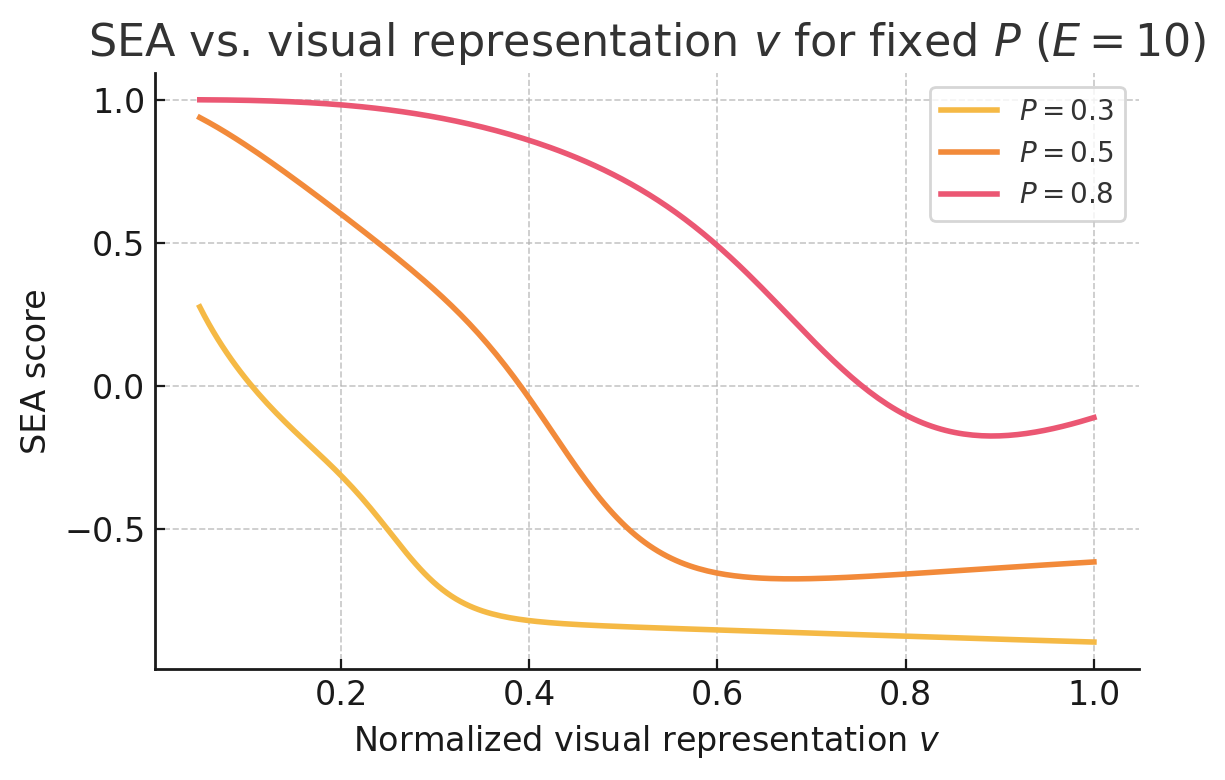}
\caption{\textbf{SEA vs.\ normalized visual representation.}
For fixed $E=10$, the SEA score decreases with $v$ at low $P$, shows a mild plateau then decline at moderate $P$, and peaks at moderate $v$ when $P$ is high, illustrating SEA's preference for efficient abstraction.}
\label{fig:sweep}
\end{figure}
\begin{figure*}[!t]
\centering
\includegraphics[width=0.89\linewidth]{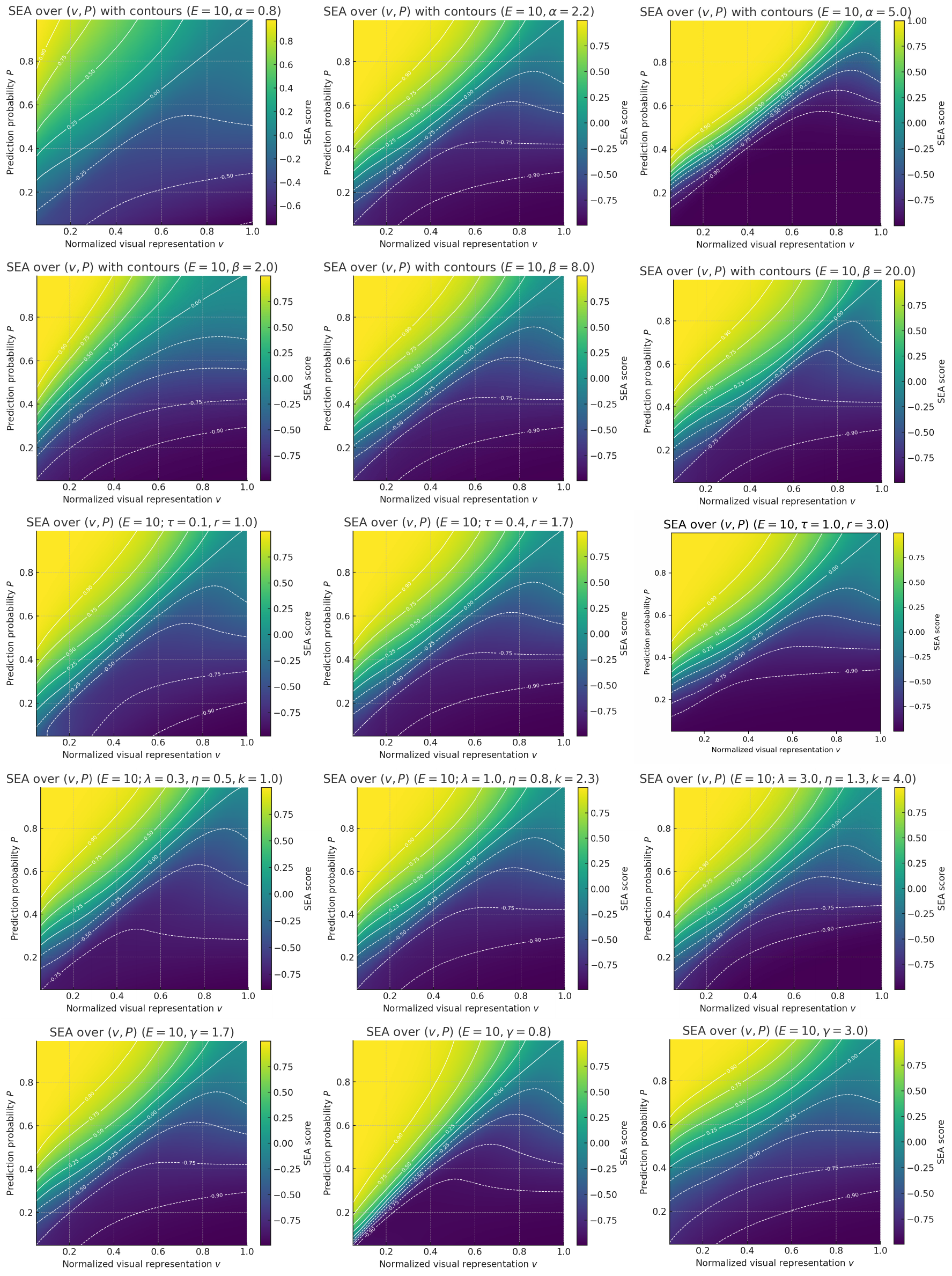}
\caption{\textbf{Effect of SEA hyperparameters on the score surface:} each row varies a single parameter (left: decreased, center: default, right: increased), illustrating how $\alpha$, $\beta$, visual representation efficiency penalties $(\lambda,\eta,k)$, base penalties $(\tau,r)$, and recognizability guidance $\gamma$ reshape the $(v,P)$ score landscape.}
\label{fig:hyper_heatmap}
\end{figure*}
\cref{fig:sweep} shows SEA scores for $v \in [0.05,1.0]$ at recognizability levels $P = 0.3, 0.5, 0.8$. The curves exhibit three characteristic regimes. When recognizability is low ($P=0.3$), the score decreases monotonically as $v$ increases, indicating that additional visual detail does not compensate for low recognizability. At moderate recognizability ($P=0.5$), SEA remains nearly flat for small $v$ and then gradually declines as the sketch becomes more detailed. When recognizability is high ($P=0.8$), the score first increases, reaches a maximum at a moderate level of visual representation, and then decreases as excessive detail is added. These trends highlight SEA's preference for efficient abstraction: highly recognizable sketches with minimal visual representation achieve higher scores, whereas unrecognizable or overly detailed sketches are penalized.

\paragraph{Hyperparameter-wise 2D heatmap analysis.}
To examine SEA over the joint space of visual ratio $v$ and recognizability $P$, we generate two-dimensional heatmaps under different hyperparameter settings. Since $v = V/E$, the qualitative structure of the SEA surface does not depend on the absolute value of $E$. For consistency, all visualizations in \cref{fig:hyper_heatmap} use $E = 10$. \cref{fig:hyper_heatmap} presents a $5 \times 3$ grid, where each row varies one hyperparameter group. The center column shows the default setting, the left column decreases the parameter, and the right column increases it. This layout provides a comparison of how each component reshapes the score surface over $(v,P)$.

The first row shows the effect of the scale parameter $\alpha$. When $\alpha$ is reduced, the heatmap becomes smoother, with broader contour bands and more gradual transitions between positive and negative scores. Increasing $\alpha$ has the opposite effect: the surface becomes dominated by the saturated extremes of $\pm 1$, and intermediate contours collapse toward the decision boundaries. Thus, $\alpha$ controls the contrast and saturation of the outer $\tanh$ activation.

The second row examines the gate sharpness parameter $\beta$, which determines how sharply SEA separates under-drawn and over-drawn sketches around $P \approx v$. With smaller $\beta$, the transition becomes broad and diffuse, producing a wide intermediate band. At the default setting, the boundary is clear but not overly sharp, whereas larger $\beta$ makes it razor-thin and causes the score to change more abruptly across the boundary. This confirms that $\beta$ mainly controls the sharpness of the consistency gate.

The third row shows how the visual representation efficiency penalty, governed by $(\lambda,\eta,k)$, changes the over-detailed region. Weakening the penalty makes the upper-right region less negative and shifts positive contours rightward, allowing more high-$v$ sketches to remain in the efficient region. In this case, $\eta<1$ softens the tail penalty as $v \rightarrow 1$, and smaller $k$ makes the low-$P$ failure region thinner. Strengthening the penalty shifts the contours leftward, sharply reducing the permissible visual representation; here, $\eta>1$ steepens the decline near $v=1$, while larger $k$ expands the failure region upward. Overall, $\lambda$ controls penalty strength, $\eta$ its curvature, and $k$ the severity of low-recognizability penalization.

The fourth row examines the base penalty parameters $\tau$ and $r$, which control how strongly SEA penalizes low recognizability regardless of visual representation. Smaller values yield a milder failure region, allowing moderately low $P$ to remain near zero. Larger values expand the negative region across the bottom of the heatmap, enforcing a clearer failure regime for insufficient recognizability.

The final row shows the effect of the recognizability guidance parameter $\gamma$. Lower values distribute the reward across recognizability levels, enlarging the efficient-abstraction region for moderate $P$. Higher values concentrate the reward near $P \approx 1$, reducing the positive region and increasing sensitivity to recognizability differences.

Overall, \cref{fig:hyper_heatmap} shows how each hyperparameter shapes the SEA landscape. Adjusting $\alpha$, $\beta$, $(\lambda,\eta,k)$, $(\tau,r)$, and $\gamma$ systematically expands or contracts the failure, efficient, and over-detailed regions. These comparisons provide a direct visual interpretation of SEA's core principle: sketches with high recognizability and efficient visual representation should be rewarded, whereas unrecognizable or overly detailed sketches should not.

\subsection{SEA as a Training Critic}
In this work we primarily use SEA as an evaluation metric. Given a set of sketches produced by a generative model or drawn from a reference dataset, we compute \(S(E,V,P)\) for each sketch and aggregate the scores to compare different generation methods, training regimes, or datasets. In this setting, SEA plays a similar role to existing scalar metrics such as FID or CLIP-based similarity~\cite{heusel2017gans,radford2021learning}: it is applied post-hoc to fixed samples and does not directly affect the training dynamics of the generator.

SEA can also be used as a reward or critic for learning. For a sampled sketch \(s\), we can define a scalar reward
\[
R(s) = S(E(s), V(s), P(s)),
\]
and consider an objective of the form
\[
\max_{\phi}\ \mathbb{E}_{s \sim \pi_{\phi}}[R(s)],
\]
where \(\pi_{\phi}\) is a sketch generator parameterized by \(\phi\). When the computation of \(E(s), V(s), P(s)\) involves non-differentiable components such as a VQA model or a discrete classifier, one may combine SEA with standard techniques for learning from non-differentiable rewards, e.g., policy-gradient methods such as REINFORCE~\cite{williams1992simple}, stop-gradient tricks, or continuous relaxations such as the Gumbel--Softmax estimator~\cite{jang2017categorical,maddison2017concrete} for discrete stroke decisions. In cases where parts of the pipeline are differentiable (e.g., CLIP-based recognizability), gradients can be back-propagated through those components while treating the remaining terms as a black-box reward.

In summary, the experiments and analyses in this paper focus on SEA from an evaluation perspective and use it only to assess models and datasets. Using SEA as a generation reward or training critic is therefore left as a extension and a promising direction for future work.

\section{Detailed Analysis on SEA Components}
\label{sec:qualitative-sea}

We disentangle SEA’s three components (commonsense elements, visual representation, and zero-shot prediction) through staged qualitative analyses. \cref{sec:appx-classifier} varies the zero-shot backbone (CLIP vs.\ OpenCLIP) and presents SEA-scored examples on SEVA and CommonSketch using each model’s prediction probabilities. \cref{sec:appx-4o-oss} varies the commonsense database (GPT-4o vs.\ GPT-OSS) and shows the same qualitative SEA diagnostics on both datasets. In all settings, visual representation is measured with two annotators, GPT-4o and Qwen2.5-VL. Across these controlled swaps, SEA scores track abstraction efficiency in a stable way, indicating robustness to changes in both the classifier and the commonsense source.

\subsection{CLIP vs. OpenCLIP on Classification}
We fixed the classifier to CLIP when computing and analyzing SEA, since CLIP exhibited the strongest alignment with human judgments on SEVA~\cite{mukherjee2024seva}. 
We additionally evaluated two classifiers, OpenCLIP ~\cite{ilharco2021openclip}and CoCa~\cite{yu2022coca}. Supplementary qualitative examples on SEVA and CommonSketch using OpenCLIP are provided in \cref{fig:seva_6classes,fig:commonsketch_clip1,fig:commonsketch_clip2}.
For this comparison, we re-estimated model–human alignment by leveraging human responses from the sketch classification questions in our user study and matching them against each model’s predictions. As shown in \cref{tbl:pred_human}, CoCa achieves the highest top-1 accuracy, whereas OpenCLIP demonstrates the strongest correlations with human assessments. Accordingly, in the subsequent analysis we adopt OpenCLIP as an alternative zero-shot classifier and conduct a qualitative comparison between SEA computed with OpenCLIP and with CLIP.
\label{sec:appx-classifier}
\begin{table}[t]
\centering
\scriptsize
\caption{Comparison of performance on Top-1 accuracy and correlation between Human assessment.}
\label{tbl:pred_human}
\begin{tabular}{lcccc}
\toprule
\textbf{Model} & \textbf{Top-1 Acc} & \textbf{Spearman’s $\rho$} & \textbf{Kendall’s $\tau$} & \textbf{Pearson’s $r$} \\
\midrule
Human     & 0.952 & --      & --      & --      \\
\midrule
CLIP      & 0.794 & 0.369  & 0.284  & 0.496  \\
OpenCLIP  & 0.912 & \textbf{0.534}  & \textbf{0.435}  & \textbf{0.785}  \\
CoCa      & \textbf{0.941} & 0.518  & 0.426  & 0.650  \\
\bottomrule
\end{tabular}

\end{table}

\subsection{GPT-4o vs. GPT-OSS on Commonsense}
\label{sec:appx-4o-oss}
\begin{figure}[t]
\centering
\includegraphics[width=\linewidth]{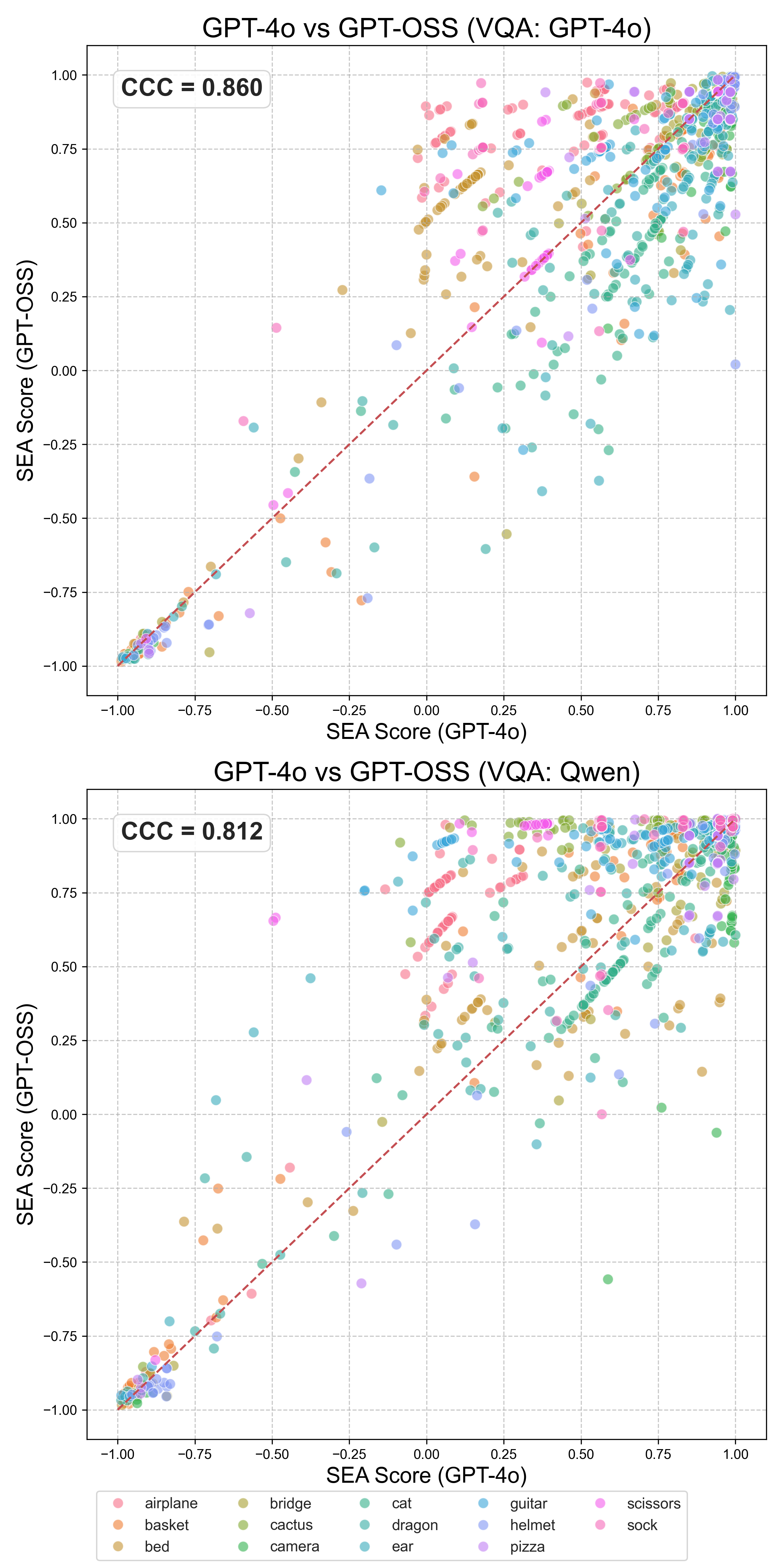}
\caption{\textbf{Comparison of SEA scores obtained using GPT-4o and GPT-OSS for commonsense extraction.} 
The top and bottom panels show results with GPT-4o and Qwen as VQA models, respectively. The red dashed line indicates the identity line ($y=x$).}
\label{fig:appendix_scatter_oss_extraction}
\vspace{-6pt}
\end{figure}

We examine whether the proprietary model GPT-4o can be replaced by an open-weights alternative for commonsense extraction, while keeping the remainder of the SEA pipeline unchanged. We therefore extract commonsense elements using multiple open-source LLMs, including GPT-OSS, Qwen-2.5, Llama 3, and Mistral, and compare their extraction behavior on the 14 classes shown in \cref{fig:all_kde_plots}. Among these candidates, GPT-OSS shows the closest agreement with GPT-4o, both in the distribution of extracted element counts and in the overall class-wise extraction tendency. We therefore adopt GPT-OSS as the open-weights extractor for feasibility verification. Using this setting, we compute SEA scores with commonsense elements derived from GPT-OSS and compare them against those obtained with GPT-4o.

In the experiment, only the commonsense extraction stage is switched from GPT-4o to GPT-OSS; all other components remain fixed, including CLIP as the classifier and the same VLM annotators for visual representations. This controlled setup allows us to isolate the effect of replacing the commonsense extractor without conflating it with changes in recognizability estimation or visual-element annotation. \cref{fig:appendix_scatter_oss_extraction} summarizes the quantitative agreement. When GPT-4o is used as the VQA annotator, SEA scores computed with GPT-OSS closely match the baseline, achieving a concordance correlation coefficient (CCC) of 0.86. This agreement remains after switching to Qwen, with a CCC of 0.812. In both cases, scatter points concentrate near the identity line ($y=x$), indicating that GPT-OSS preserves the same semantic ranking patterns as GPT-4o regardless of annotator choice.

\begin{figure}[t]
\centering
\includegraphics[width=0.99\linewidth]
{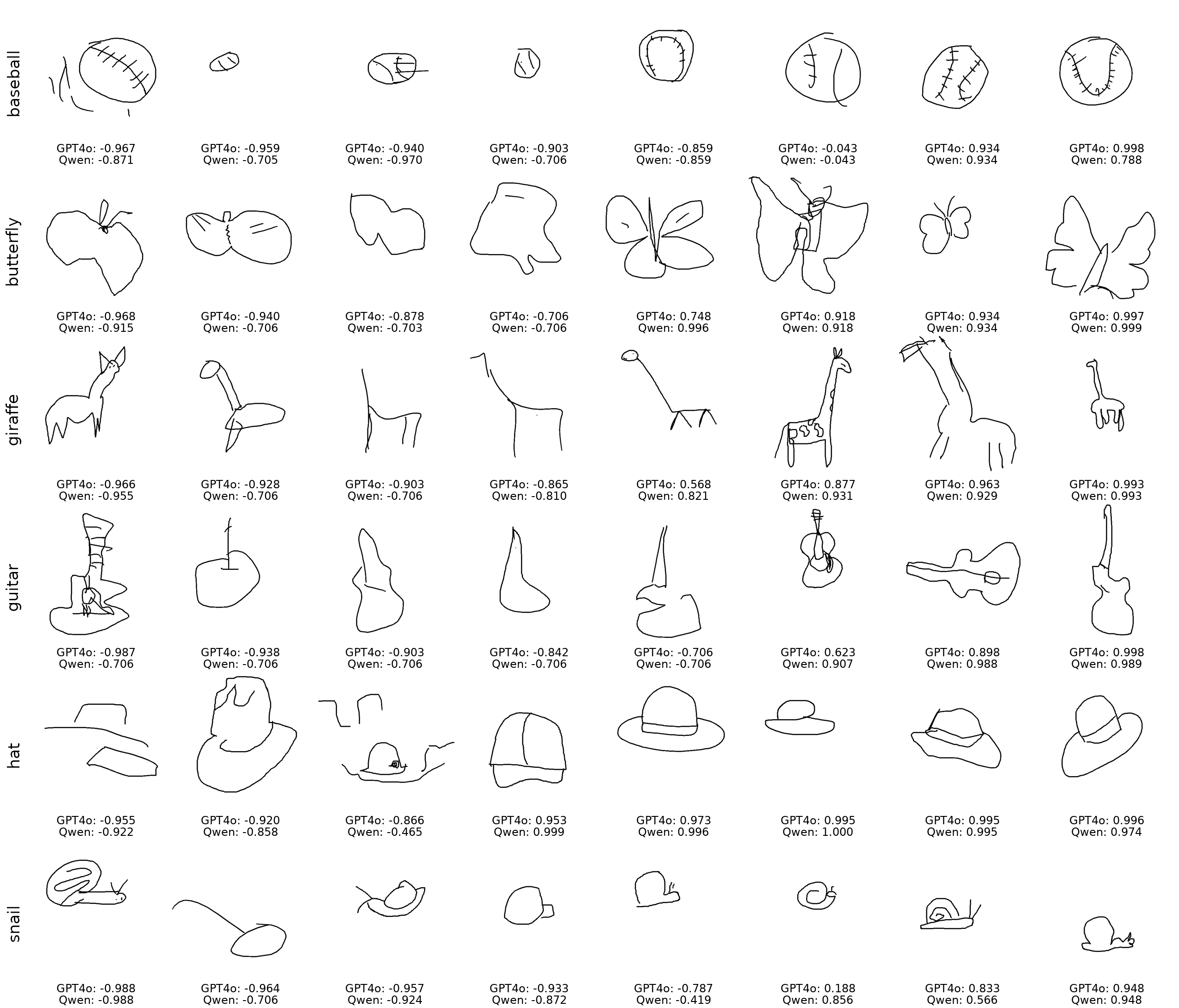}
\caption{\textbf{Qualitative examples by SEA score on SEVA.}
Six SEVA classes are shown (\texttt{baseball}, \texttt{butterfly}, \texttt{giraffe}, \texttt{guitar}, \texttt{hat}, \texttt{snail}),
with eight sketches per class selected to span a low-to-high SEA range for visual inspection.
All sketches display four SEA variants computed under OpenCLIP as a classifier:
\emph{GPT4o} (GPT-4o database + GPT-4o annotations),
\emph{Qwen} (GPT-4o database + Qwen annotations).}
\label{fig:seva_6classes}
\end{figure}

\begin{figure}[t]
\centering
\includegraphics[width=0.99\linewidth]
{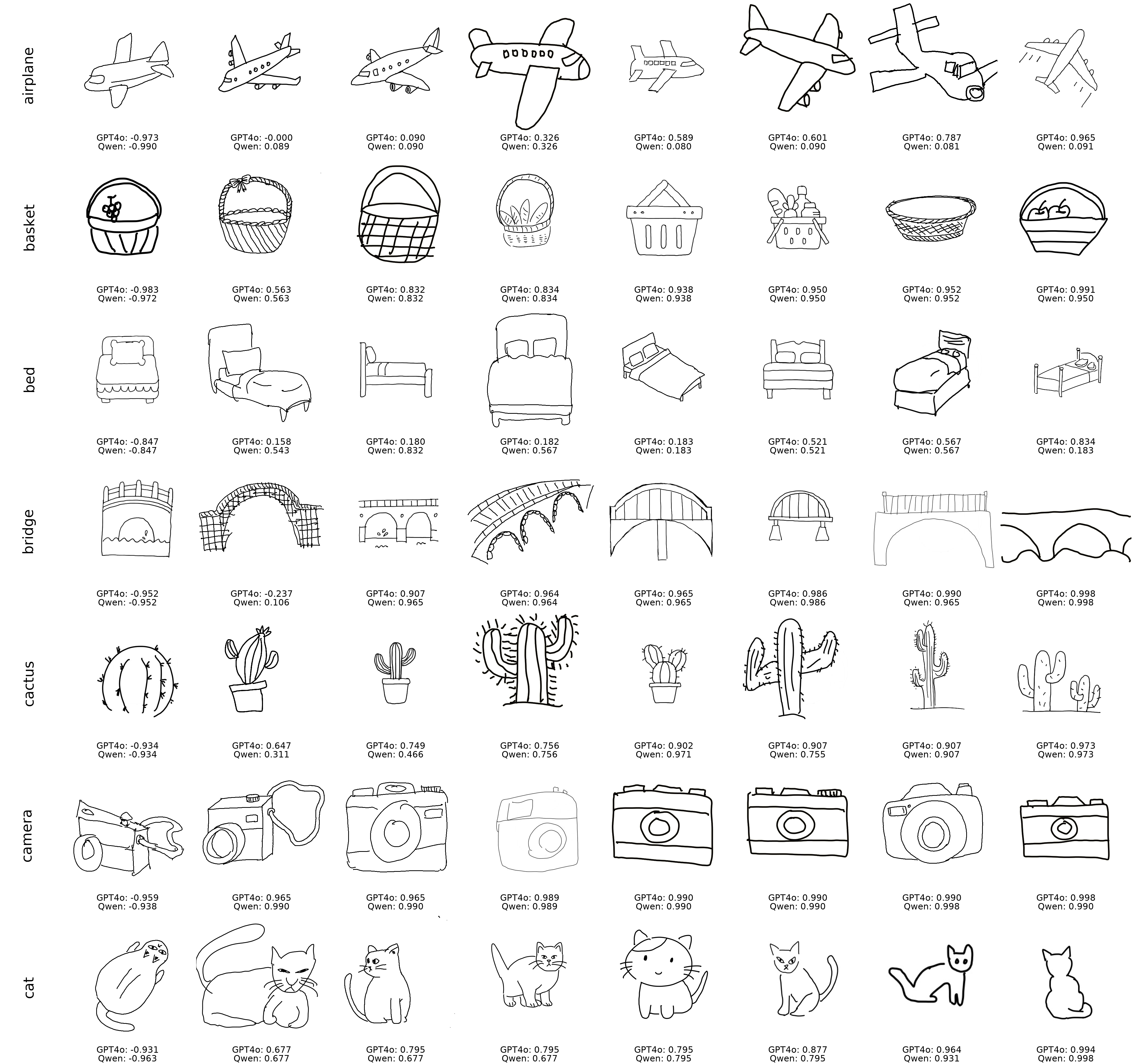}
\caption{\textbf{Qualitative examples by SEA score on CommonSketch (set 1).}
Seven CommonSketch classes are shown (\texttt{airplane}, \texttt{basket}, \texttt{bed}, \texttt{bridge},
\texttt{cactus}, \texttt{camera}, \texttt{cat}).
All sketches display four SEA variants computed under OpenCLIP as a classifier:
\emph{GPT4o} (GPT-4o database + GPT-4o annotations),
\emph{Qwen} (GPT-4o database + Qwen annotations).}
\label{fig:commonsketch_clip1}
\end{figure}

\begin{figure}[t]
\centering
\includegraphics[width=0.99\linewidth]
{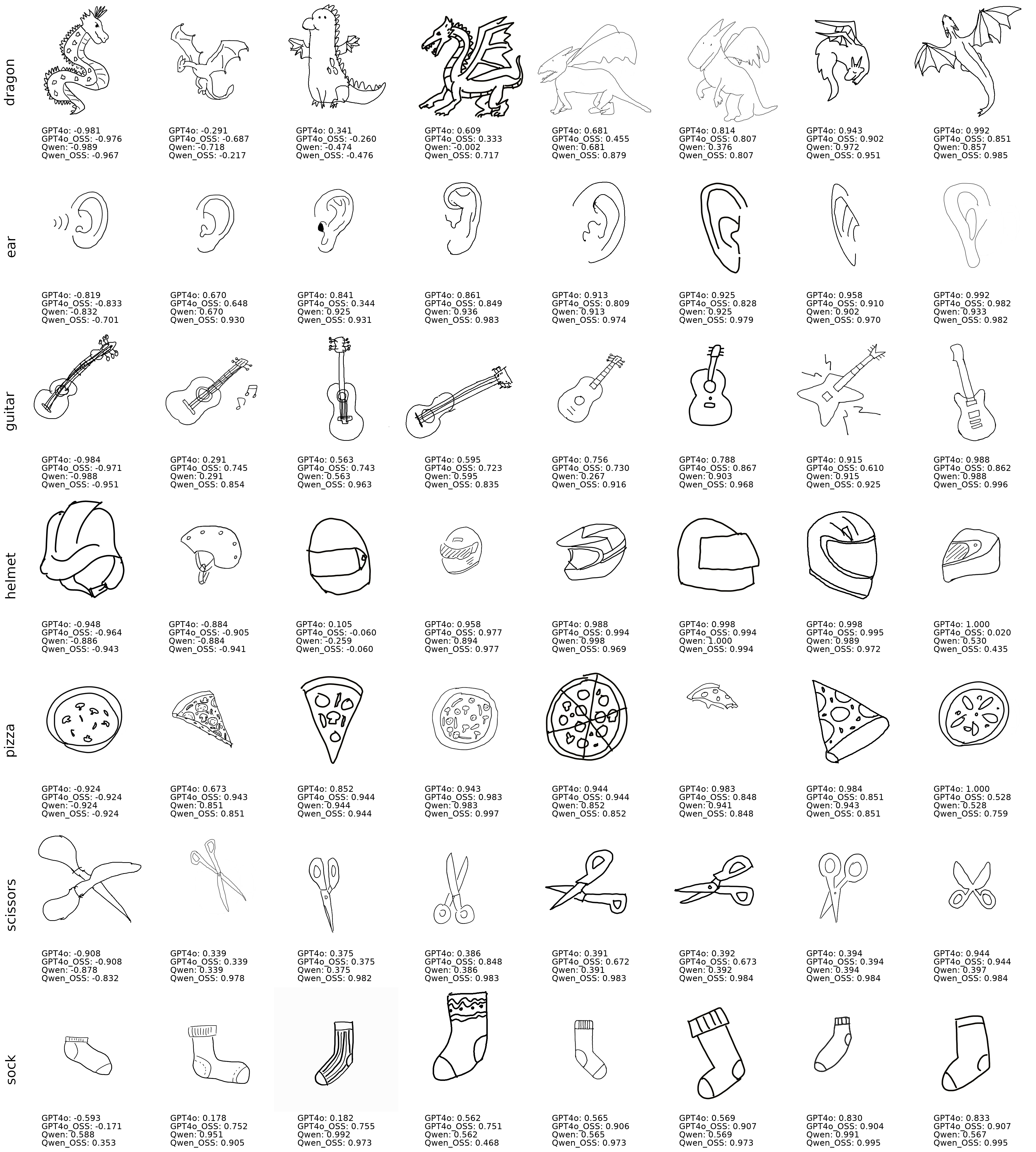}
\caption{\textbf{Qualitative examples by SEA score on CommonSketch (set 2).}
Seven additional CommonSketch classes are shown (\texttt{dragon}, \texttt{ear}, \texttt{guitar},
\texttt{helmet}, \texttt{pizza}, \texttt{scissors}, \texttt{sock}).
All sketches display four SEA variants computed under OpenCLIP as a classifier:
\emph{GPT4o} (GPT-4o database + GPT-4o annotations),
\emph{Qwen} (GPT-4o database + Qwen annotations).}
\label{fig:commonsketch_clip2}
\end{figure}

Qualitative results in \cref{fig:seva_6classes,fig:commonsketch_clip1,fig:commonsketch_clip2} corroborate this trend. Across both SEVA and CommonSketch, sketches are ordered within each class, with the score increasing from left to right. Low-scoring examples are typically hard to recognize and therefore receive negative SEA values together with low prediction probabilities. Mid-range examples contain sufficient detail to support recognition, yielding high visual representations and prediction probabilities that correspond to a reasonable abstraction regime. High-scoring examples maintain strong recognizability despite comparatively lower visual representations, reflecting superior abstraction efficiency in which minimal depiction suffices for reliable classification. This same progression is preserved when GPT-OSS is used for commonsense extraction, indicating that the resulting score differences do not alter the qualitative ordering of sketches within each class. These results show that SEA tracks visual abstraction efficiency in a stable and interpretable manner, and that its qualitative ordering and quantitative scores remain consistent even when the LLM used for commonsense element extraction is replaced.

\onecolumn
\def\figwidth{0.246\linewidth}

\newlength{\figrowspace}
\setlength{\figrowspace}{11.5pt}

\newcommand{\addsketch}[1]{%
    \includegraphics[width=\linewidth, trim=20pt 20pt 20pt 20pt, clip]{#1}%
}

\begin{figure}[h!]
    \centering
    \setlength{\tabcolsep}{1pt} 
    \vspace{10pt} 

    \begin{subfigure}[b]{\figwidth}
        \centering
        \addsketch{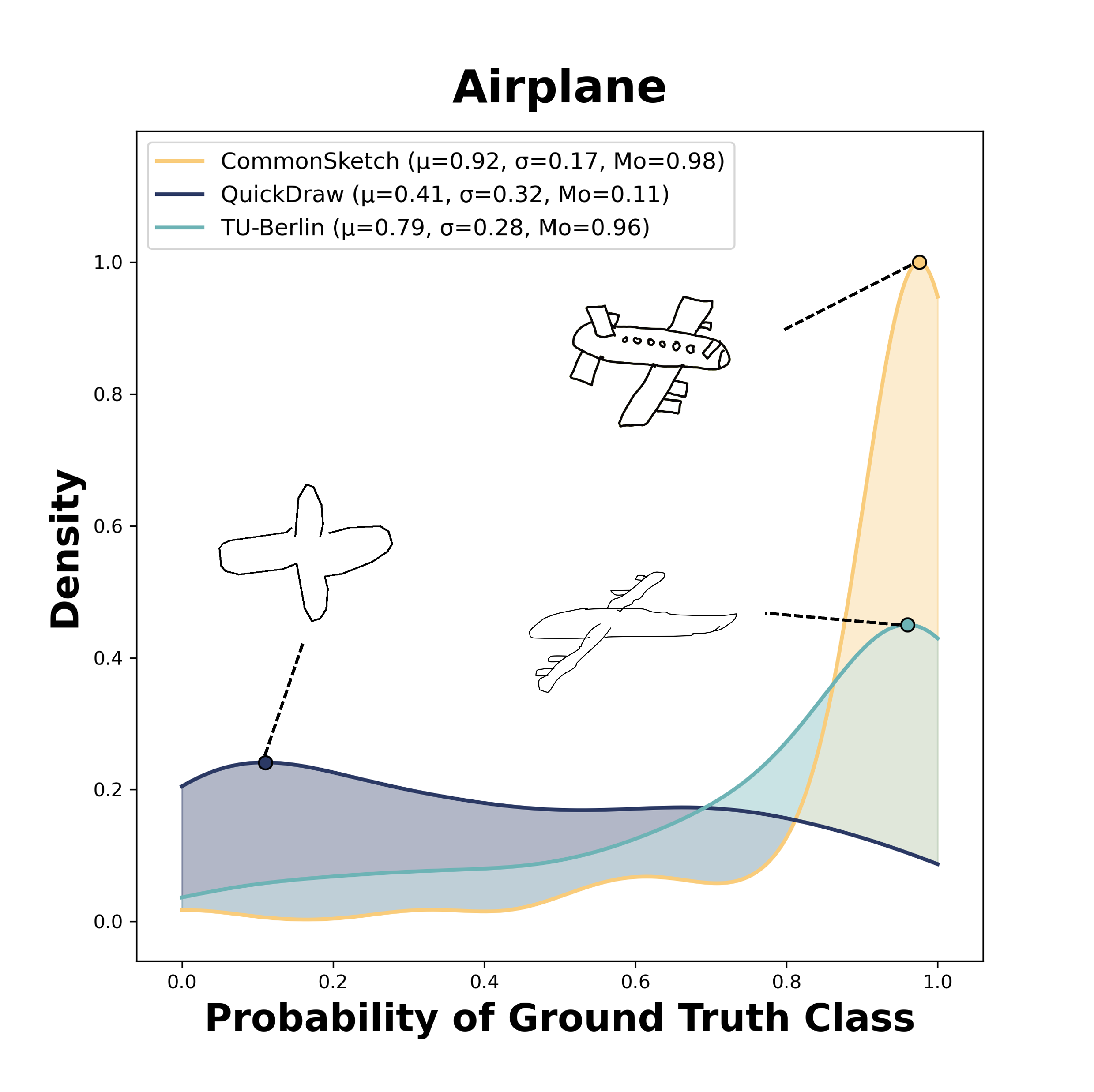}
    \end{subfigure}
    \hfill
    \begin{subfigure}[b]{\figwidth}
        \centering
        \addsketch{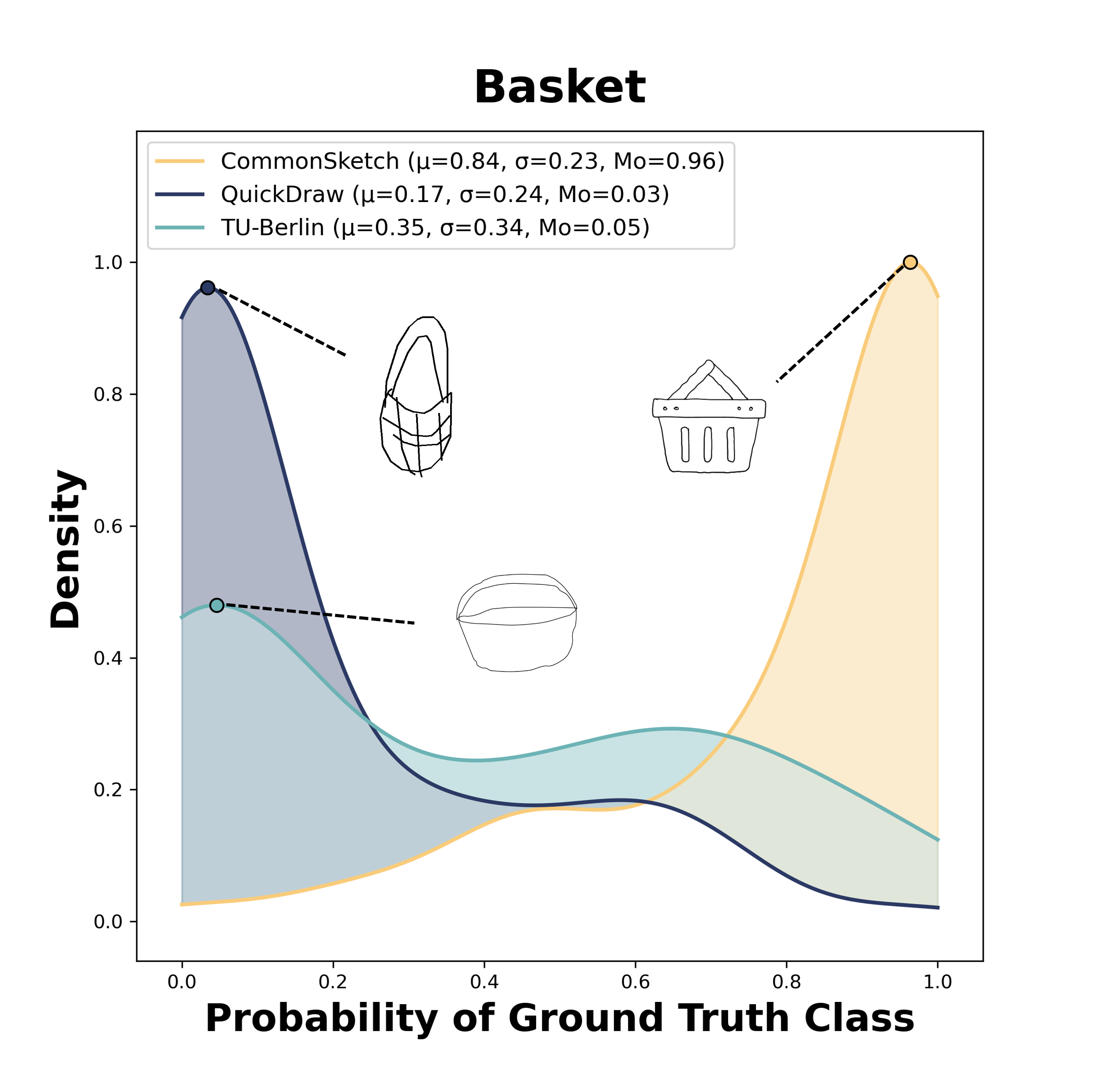}
    \end{subfigure}
    \hfill
    \begin{subfigure}[b]{\figwidth}
        \centering
        \addsketch{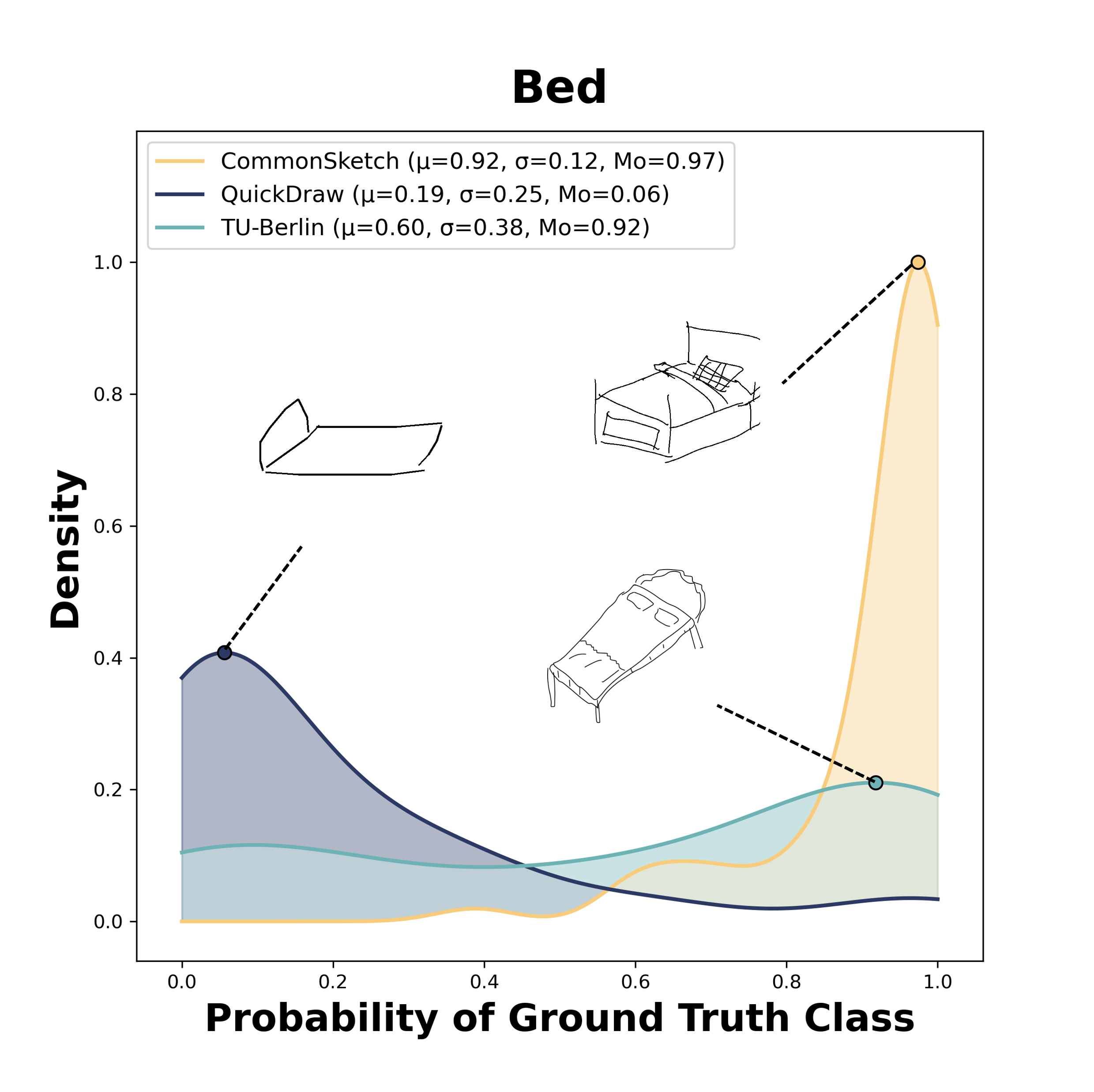}
    \end{subfigure}
    \hfill
    \begin{subfigure}[b]{\figwidth}
        \centering
        \addsketch{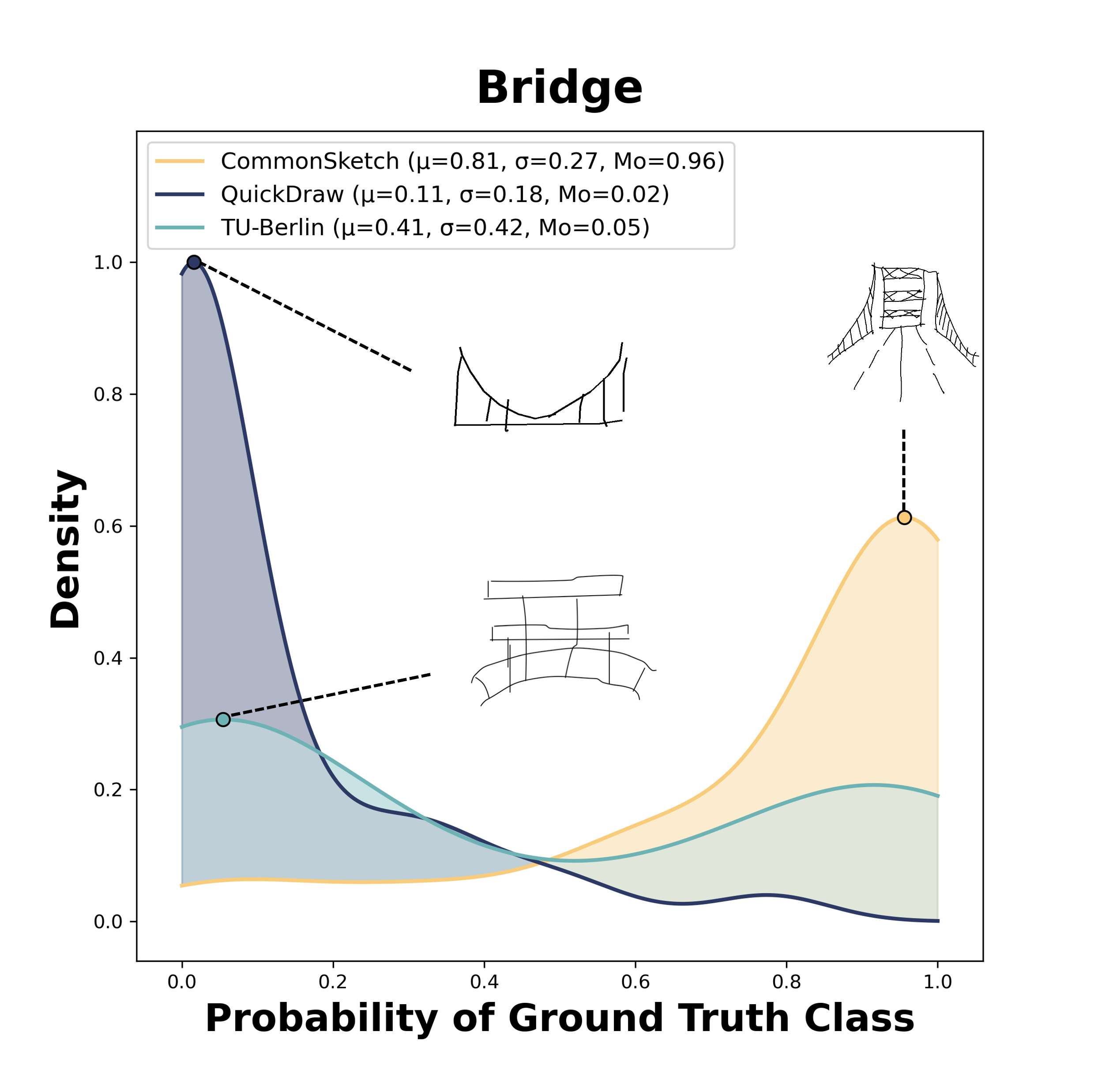}
    \end{subfigure}
    
    \vspace{\figrowspace}

    \begin{subfigure}[b]{\figwidth}
        \centering
        \addsketch{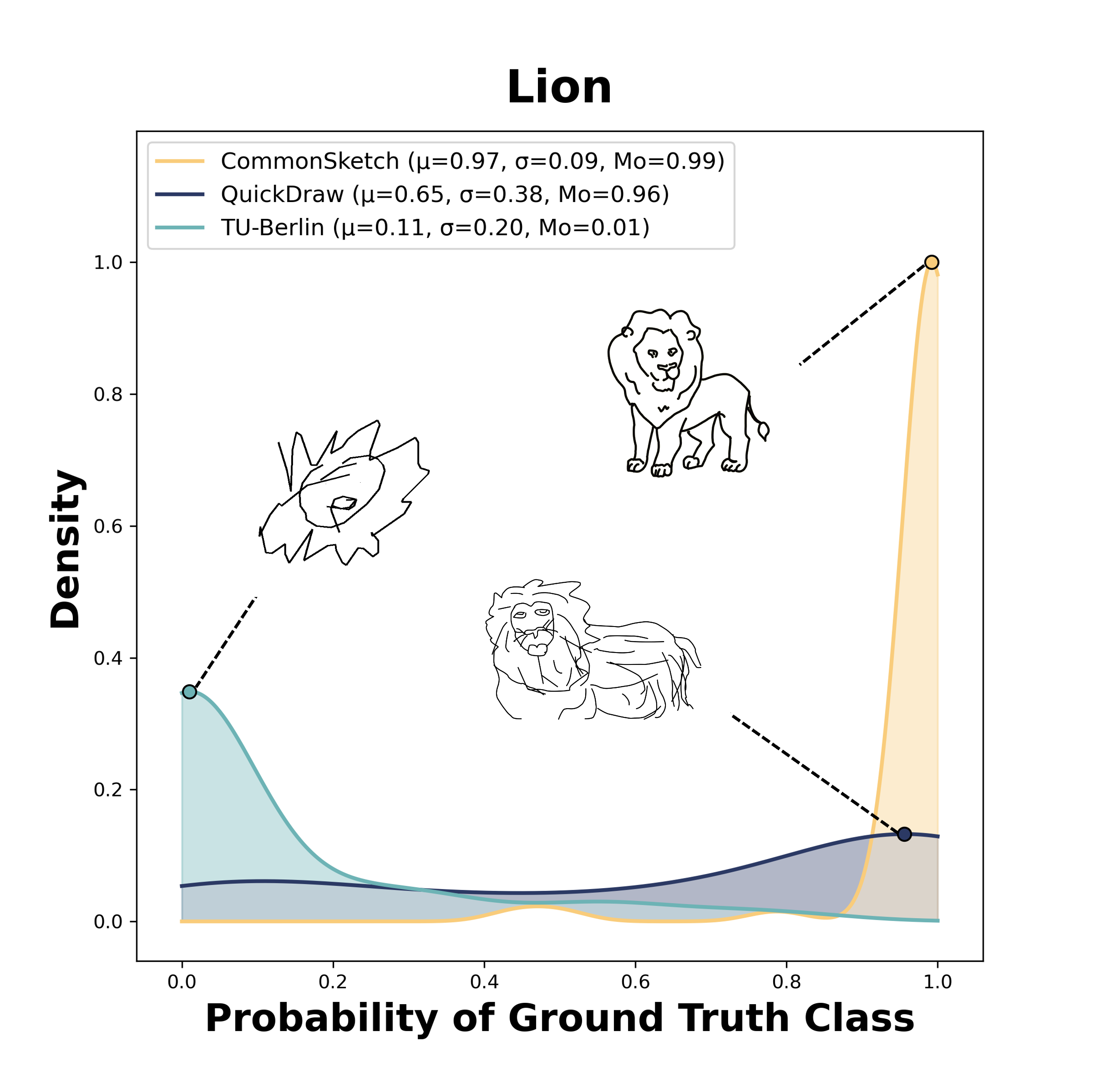}
    \end{subfigure}
    \hfill
    \begin{subfigure}[b]{\figwidth}
        \centering
        \addsketch{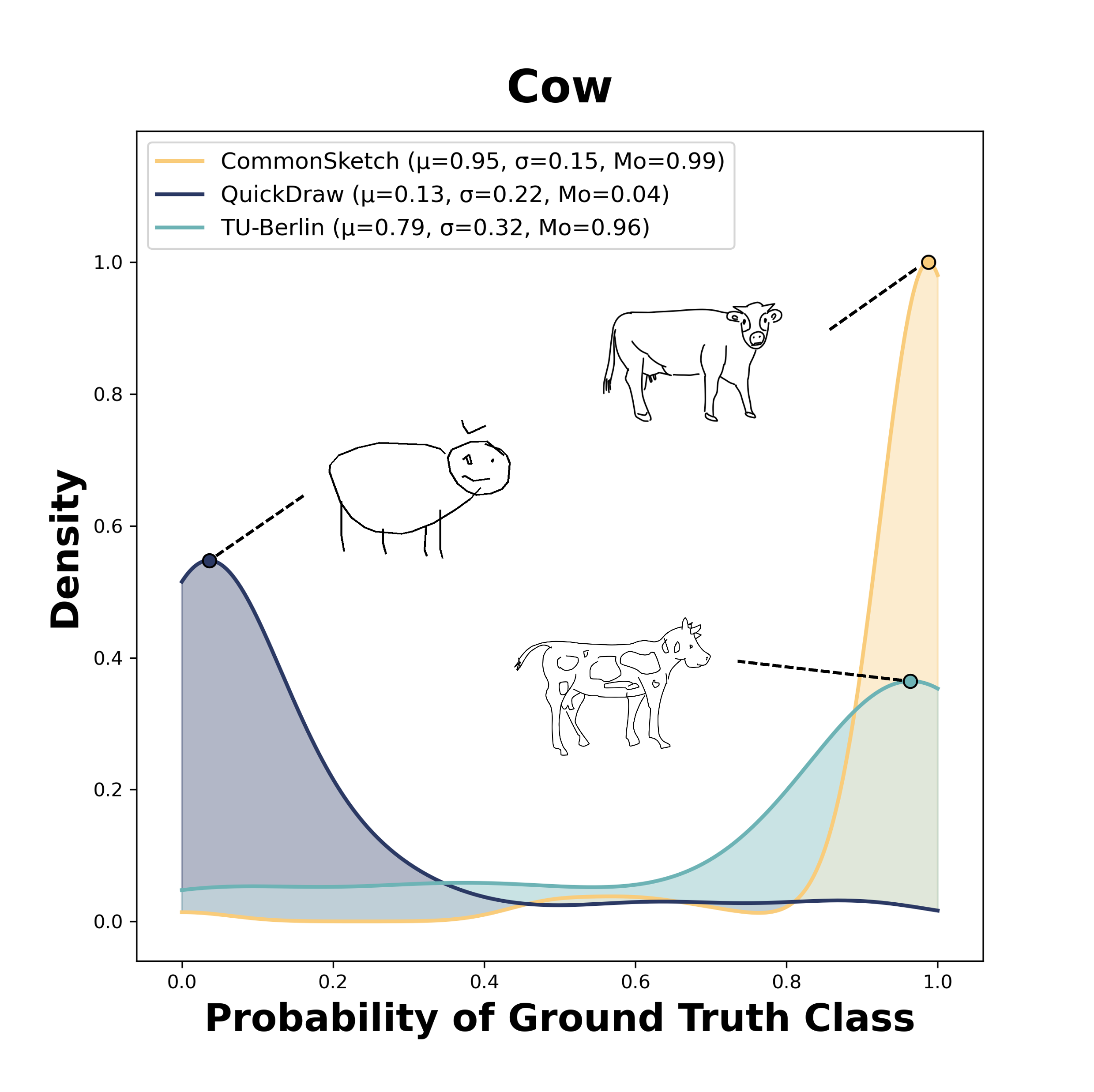}
    \end{subfigure}
    \hfill
    \begin{subfigure}[b]{\figwidth}
        \centering
        \addsketch{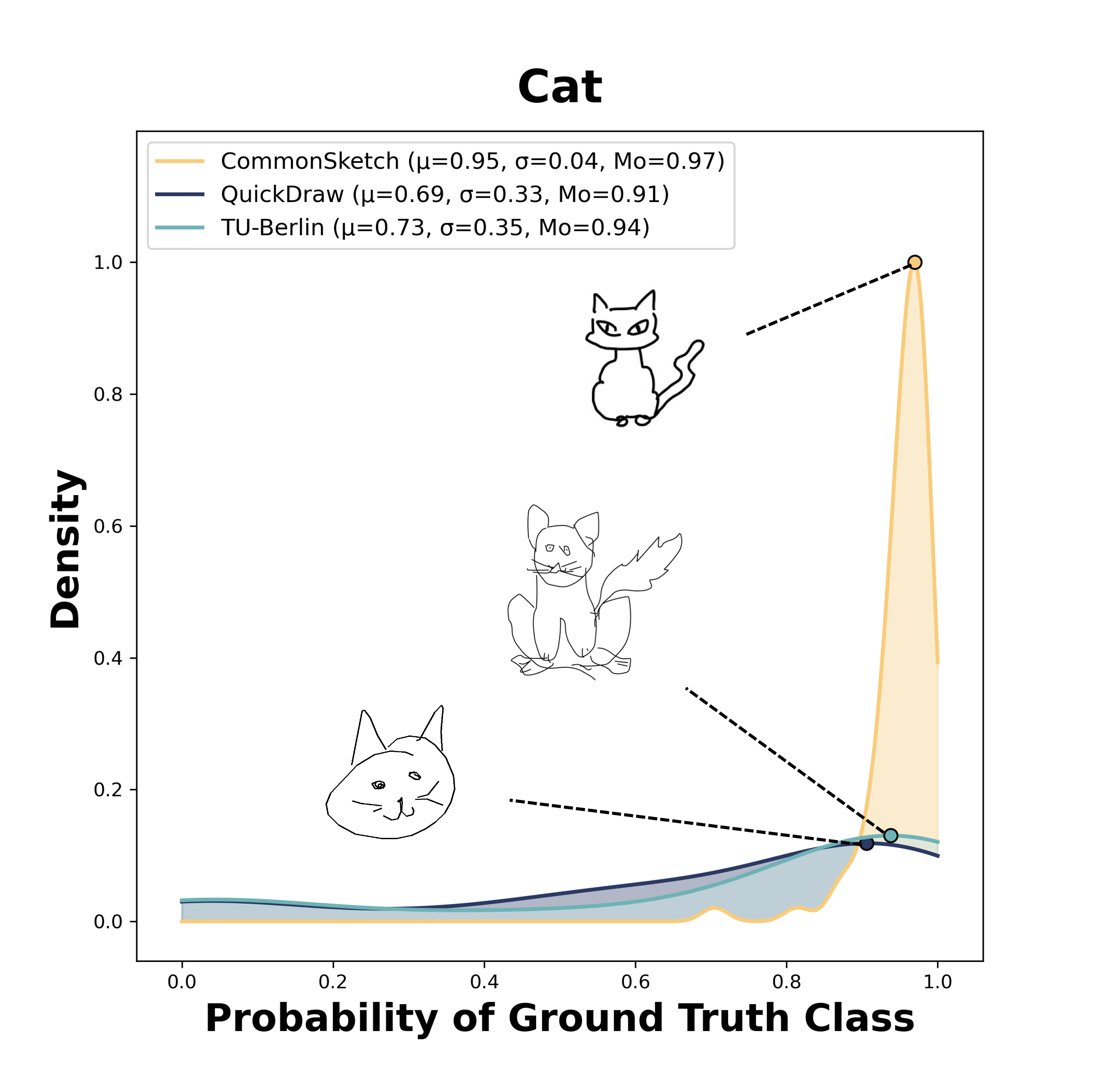}
    \end{subfigure}
    \hfill
    \begin{subfigure}[b]{\figwidth}
        \centering
        \addsketch{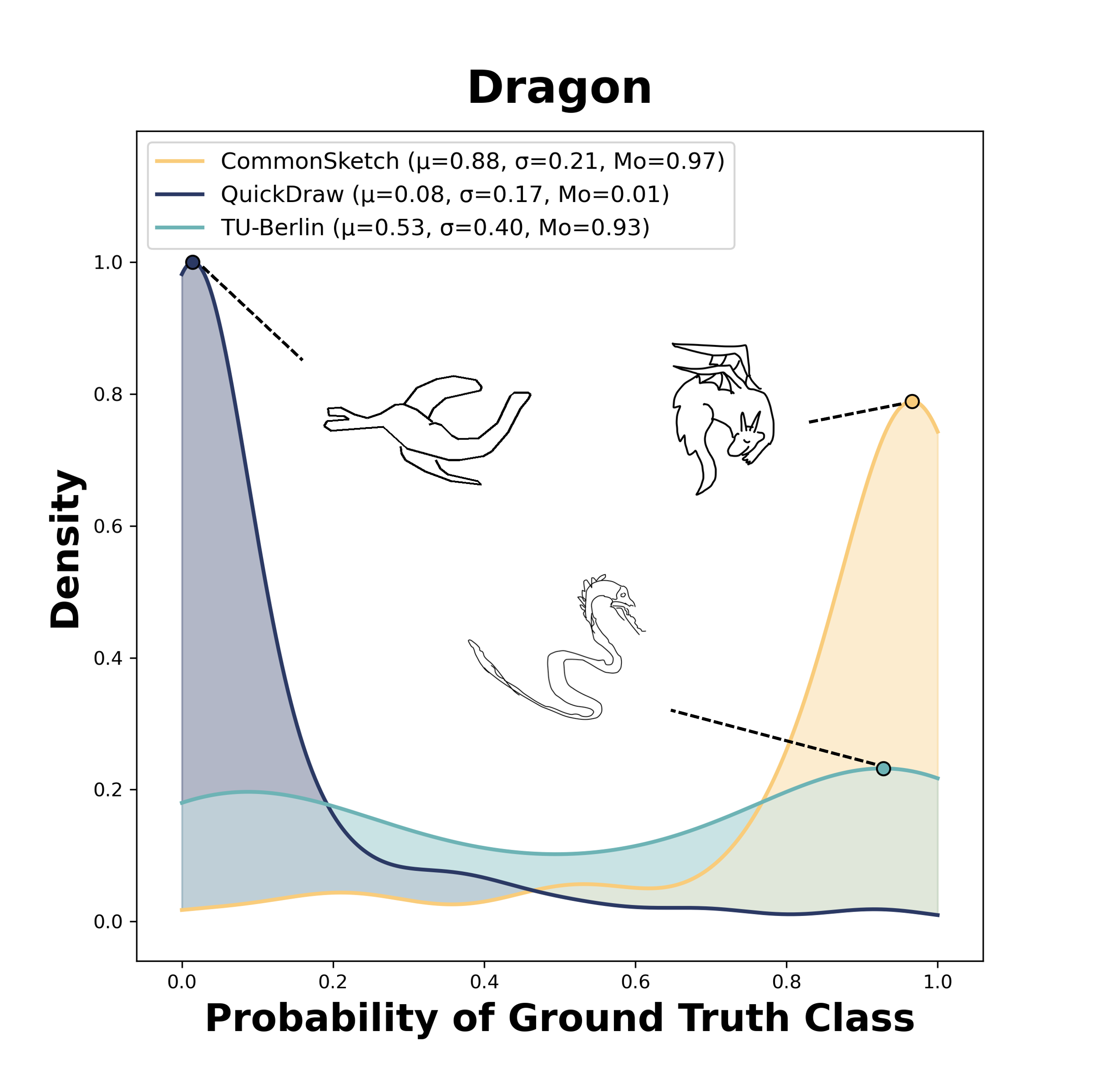}
    \end{subfigure}

    \vspace{\figrowspace}

    \begin{subfigure}[b]{\figwidth}
        \centering
        \addsketch{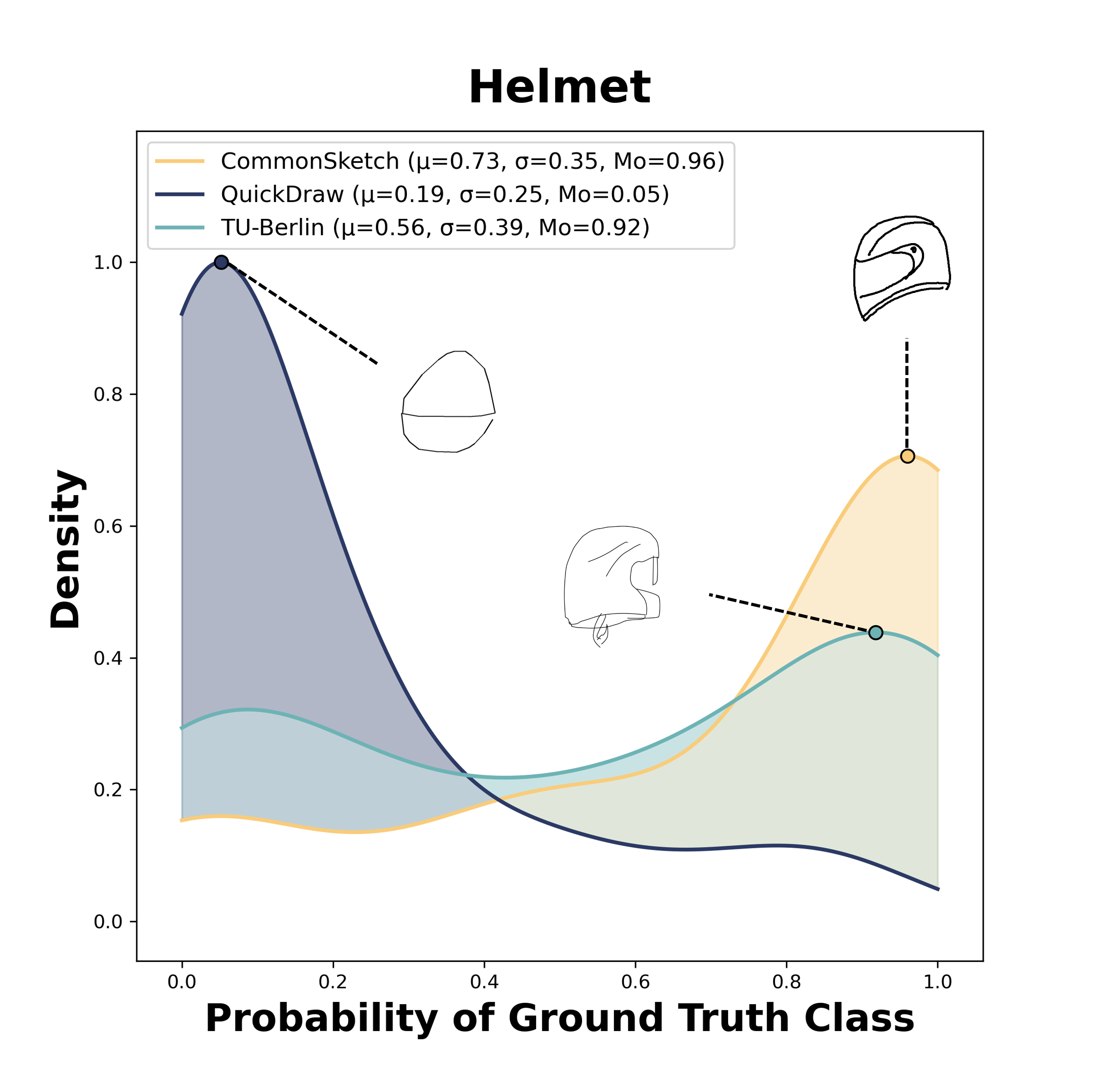}
    \end{subfigure}
    \hfill
    \begin{subfigure}[b]{\figwidth}
        \centering
        \addsketch{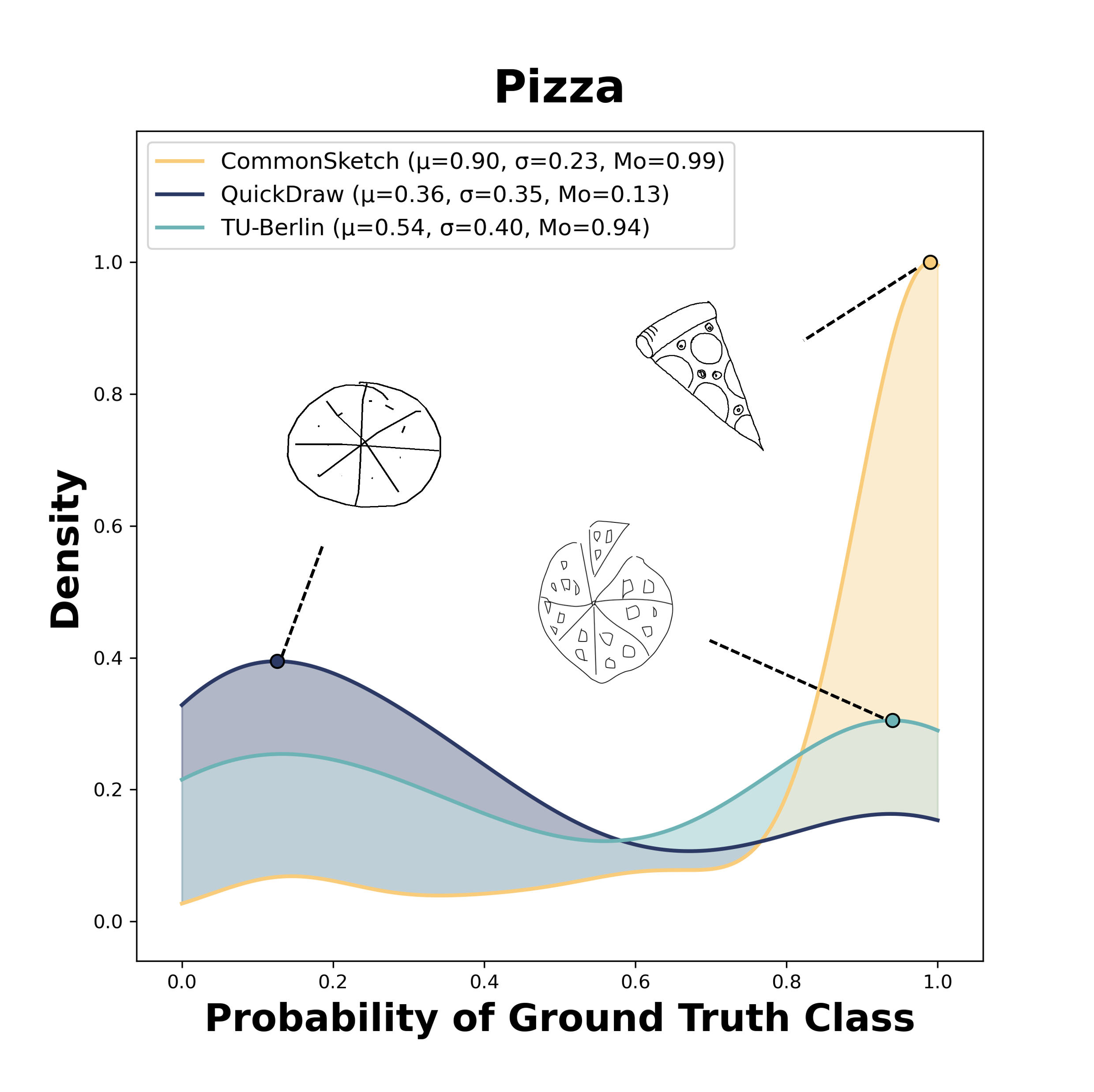}
    \end{subfigure}
    \hfill
    \begin{subfigure}[b]{\figwidth}
        \centering
        \addsketch{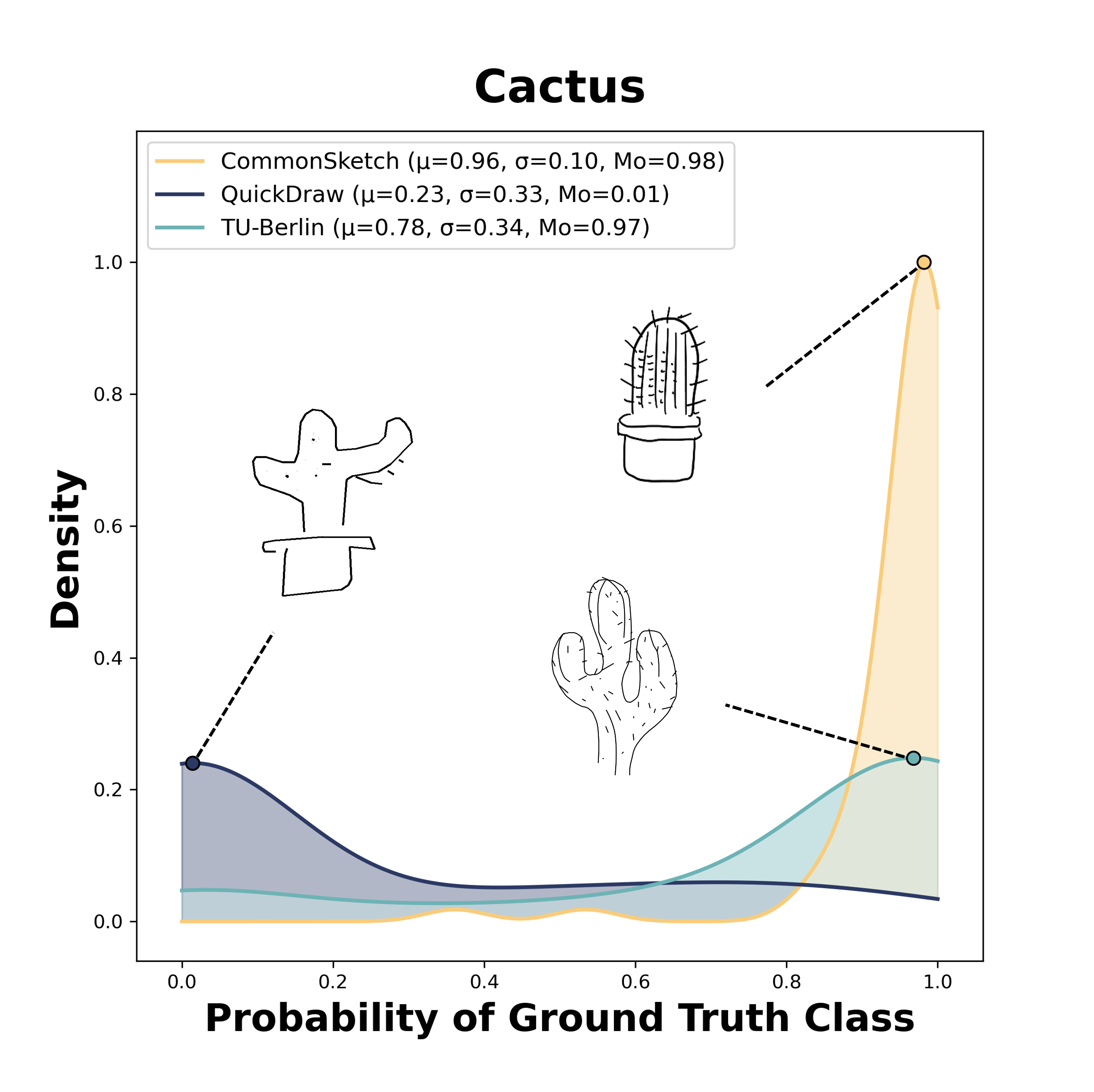} 
    \end{subfigure}
    \hfill
    \begin{subfigure}[b]{\figwidth}
        \centering
        \addsketch{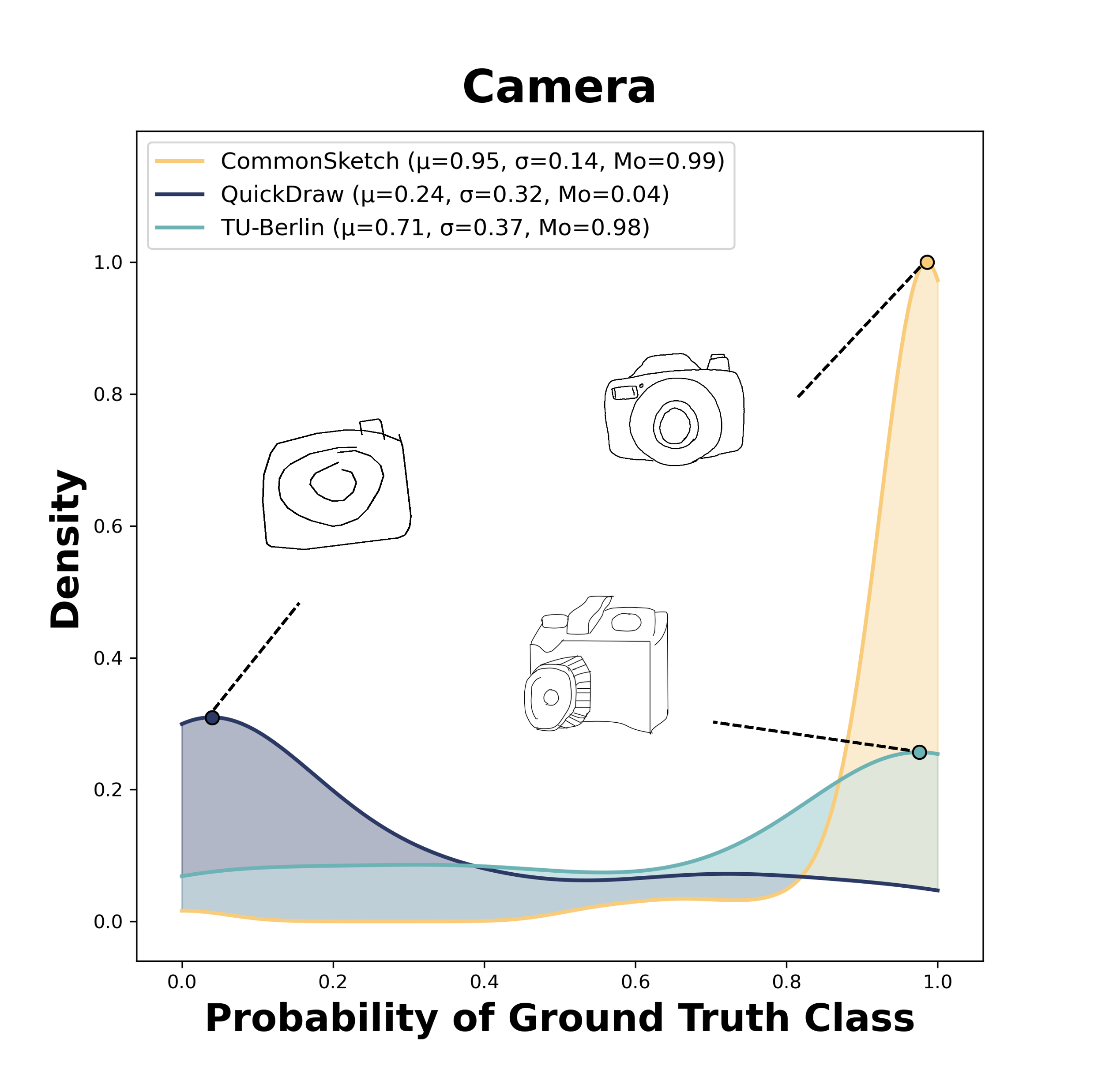} 
    \end{subfigure}

    \vspace{\figrowspace}

    \begin{subfigure}[b]{\figwidth}
        \centering
        \addsketch{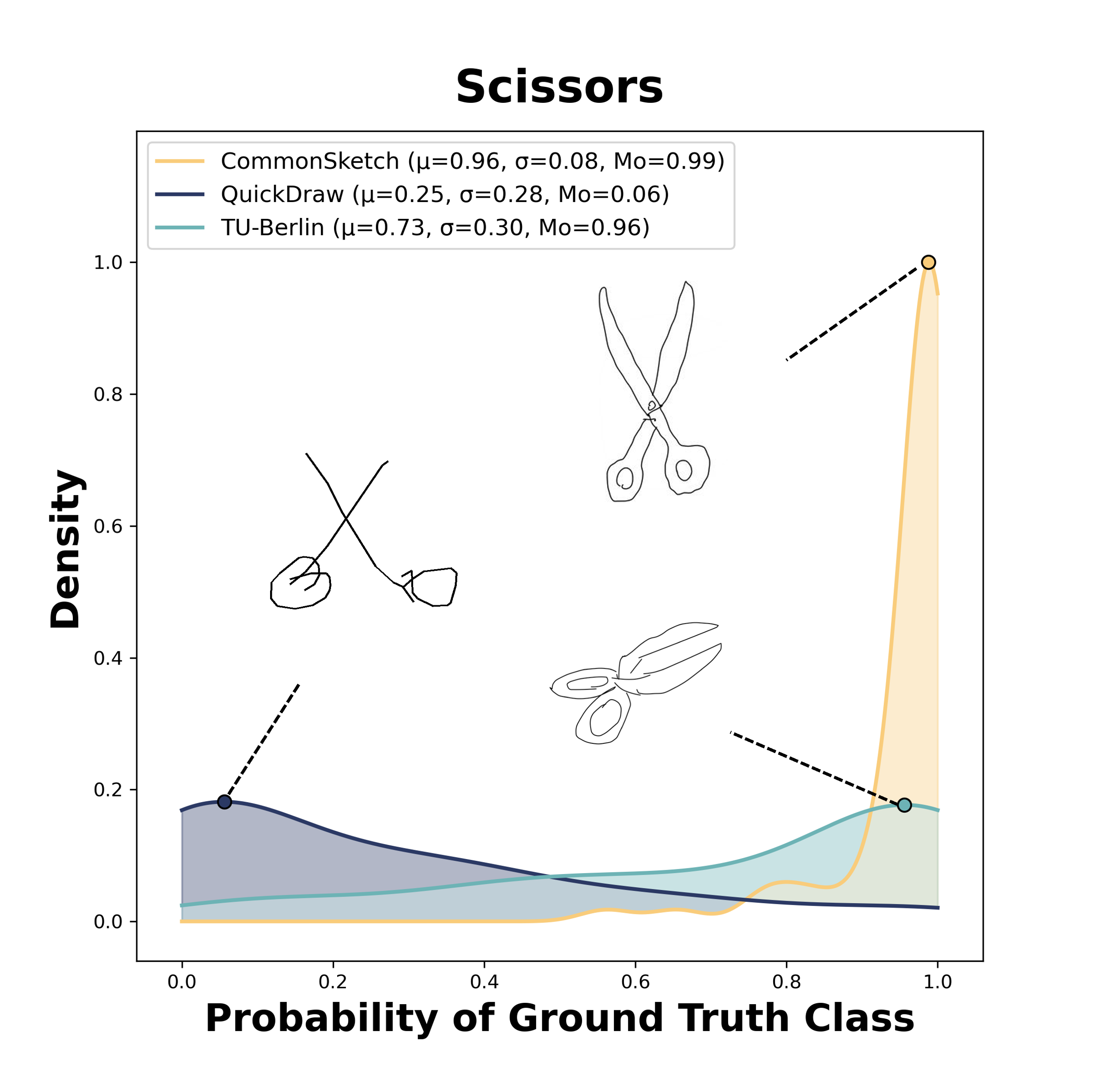}
    \end{subfigure}
    \hfill
    \begin{subfigure}[b]{\figwidth}
        \centering
        \addsketch{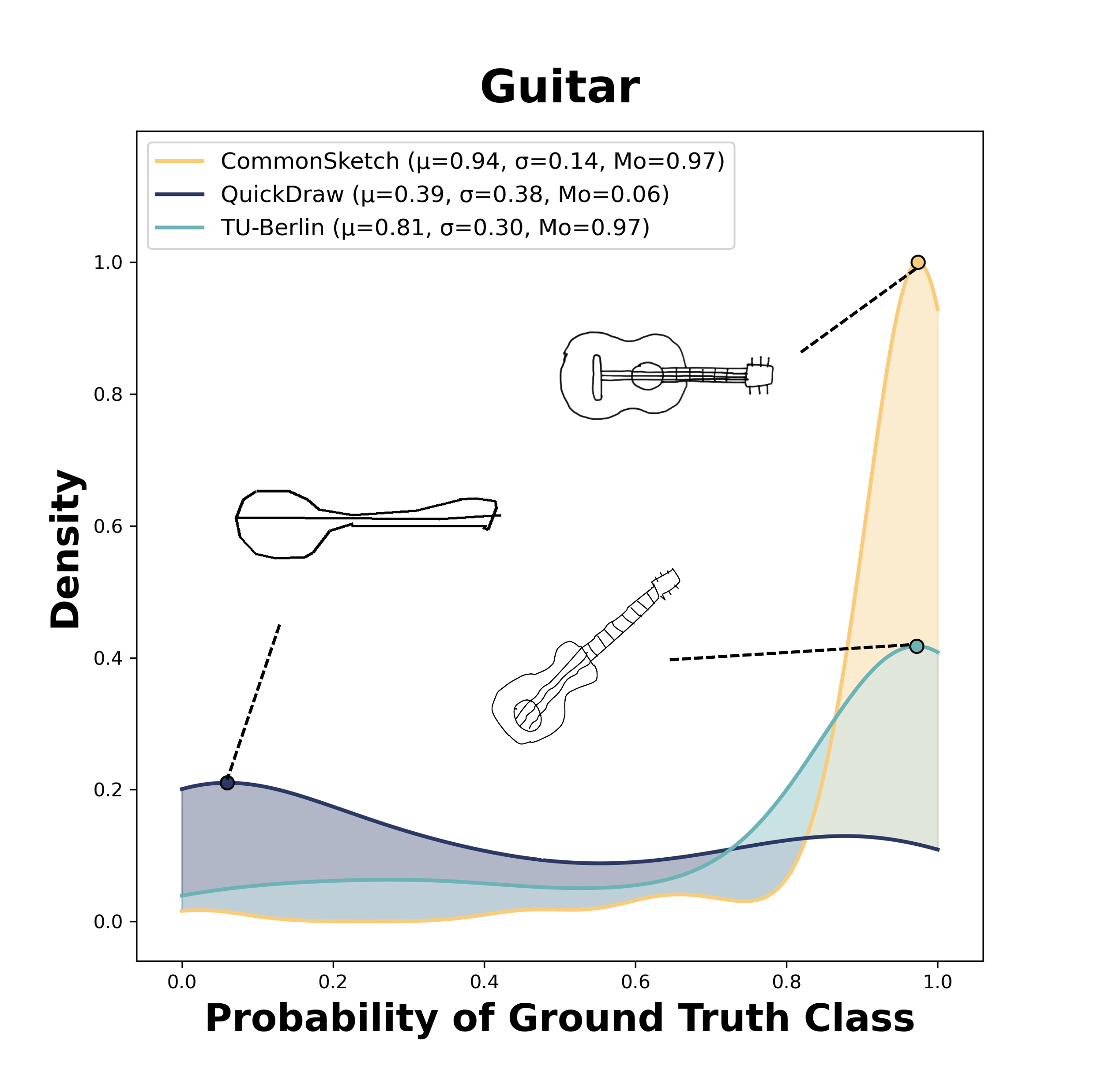}
    \end{subfigure}
    \hfill
    \begin{subfigure}[b]{\figwidth}
        \centering
        \addsketch{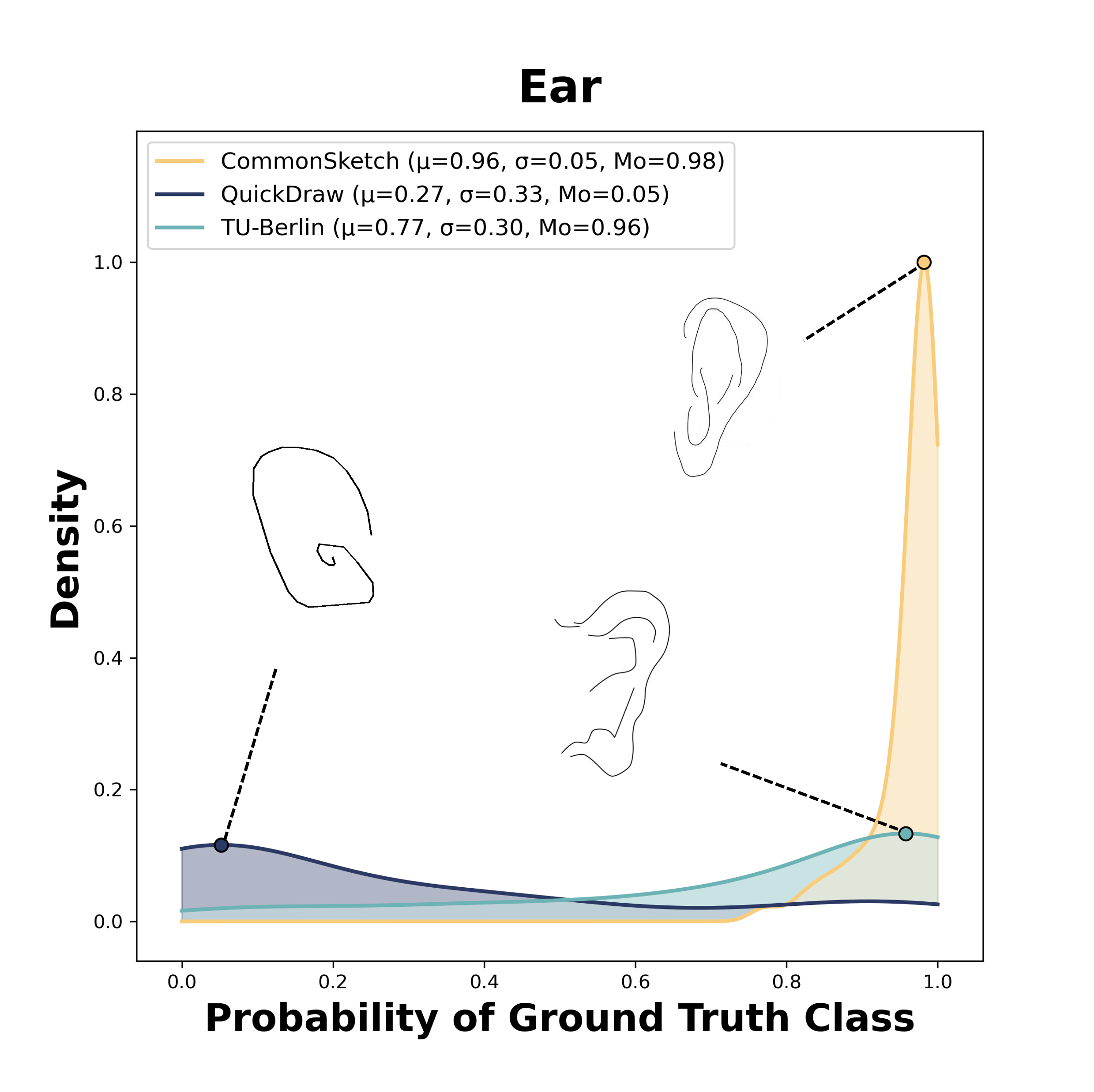}
    \end{subfigure}
    \hfill
    \begin{subfigure}[b]{\figwidth}
        \centering
        \addsketch{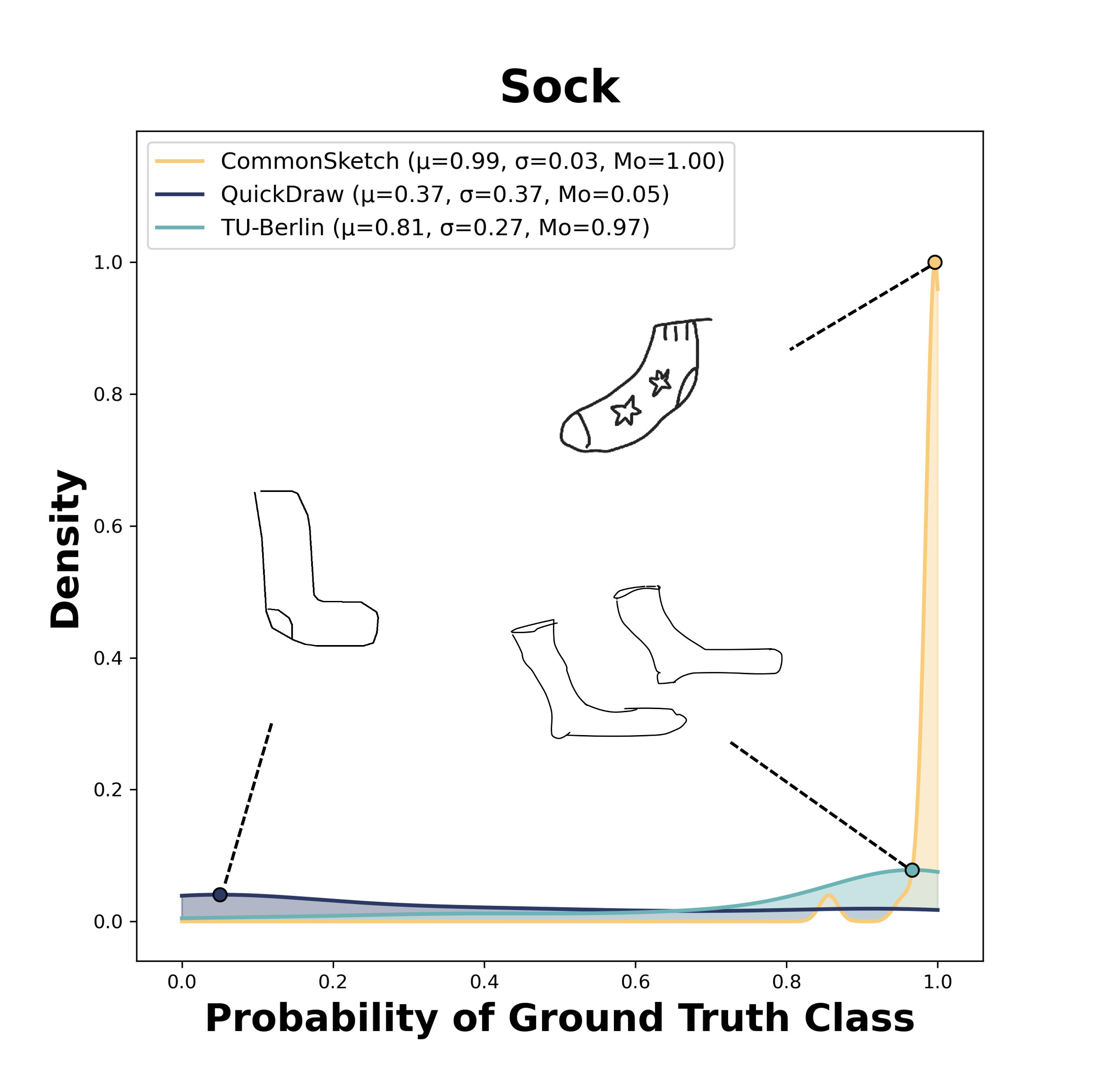}
    \end{subfigure}

    \vspace{10pt}    
    \caption{\textbf{Cross-dataset sketch quality comparison.}
    KDE plots show the \emph{Probability of the Ground Truth Class} for 14 classes shared across CommonSketch, QuickDraw, and TU-Berlin, with one class chosen from each of the 14 categories.
    For each sketch, class probability is computed separately with CLIP, OpenCLIP, and CoCa, and the three values are then averaged to obtain a model-agnostic recognizability score.
    The distributions highlight differences in typical recognizability and variance across datasets, providing a fine-grained view beyond single-number quality metrics.
    Representative sketches at the \emph{mode} of each KDE are displayed to link the quantitative trends to visually typical samples.}
    \label{fig:all_kde_plots}
\end{figure}

\clearpage
\twocolumn

\onecolumn
\twocolumn

\begin{figure}[t]
\centering
\includegraphics[width=0.99\linewidth]
{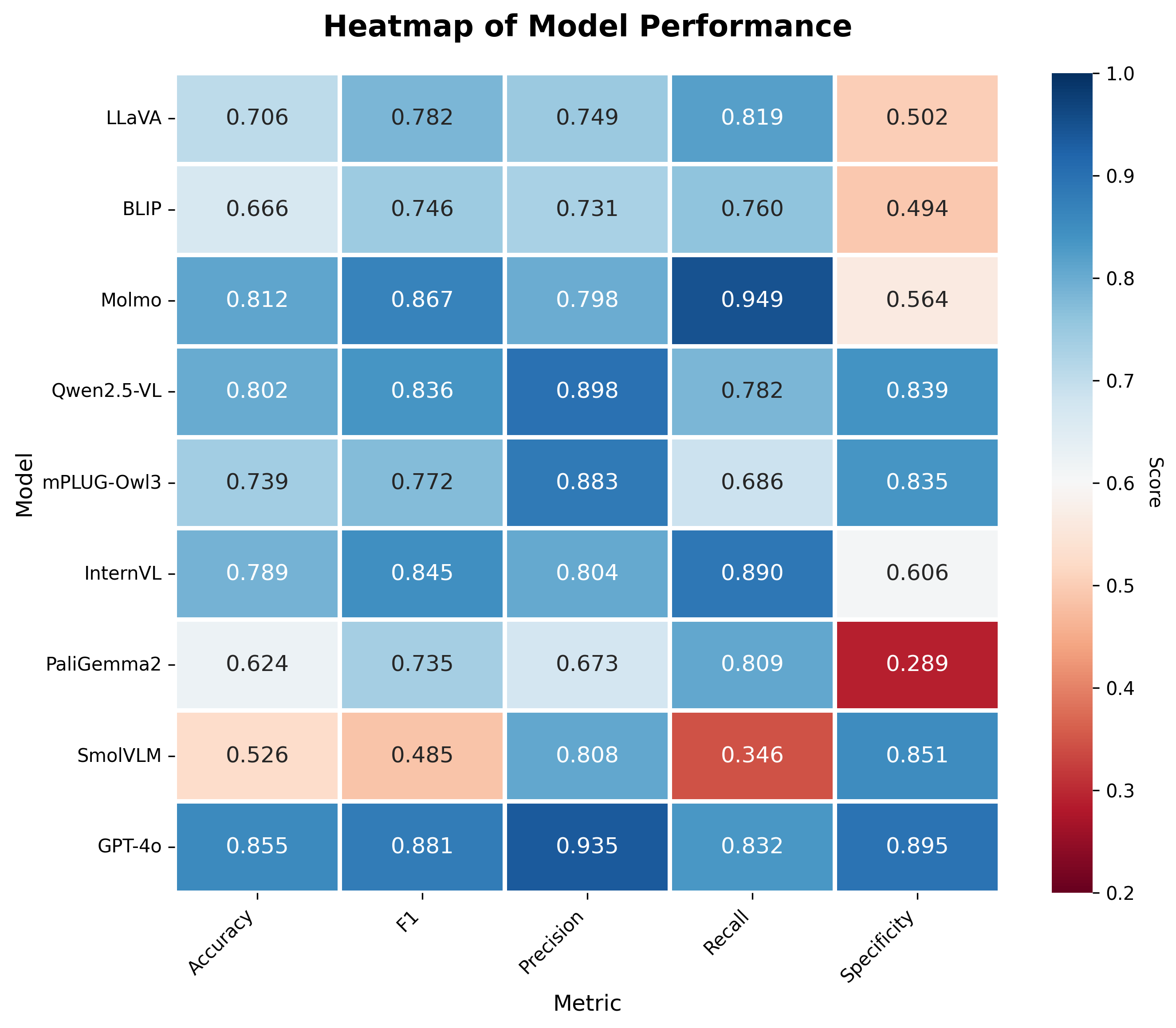}
\caption{\textbf{Comprehensive performance evaluation of nine VLMs.} 
The heatmap visualizes model performance across five key metrics. Higher scores are in blue, and lower scores in red.
}
\label{fig:appendix_heatmap}
\end{figure}
\begin{figure*}[t]
\centering
\includegraphics[width=0.99\linewidth]{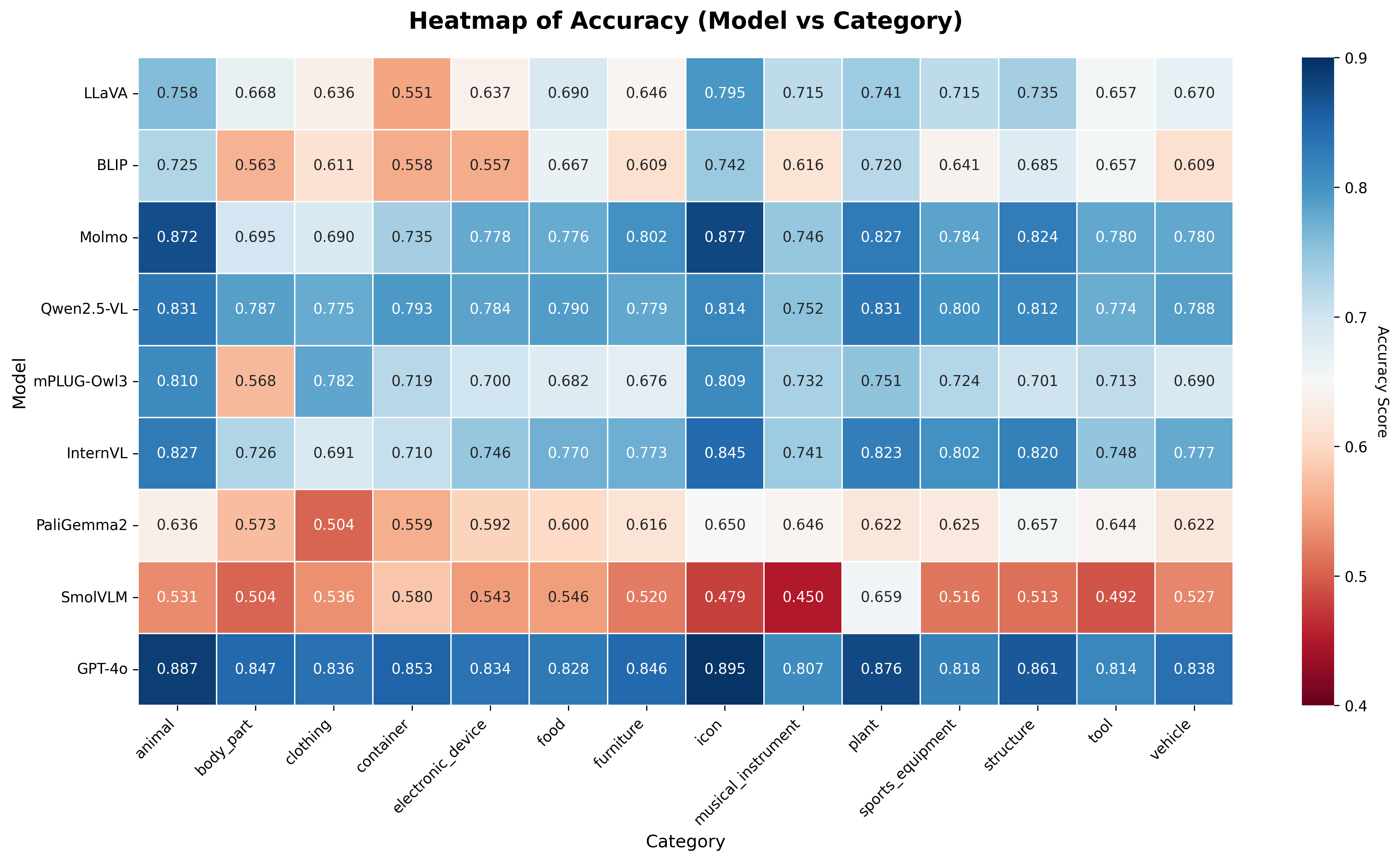}
\caption{\textbf{Per-category performance comparison of nine VLMs across the 14 categories of CommonSketch.} The heatmap visualizes the accuracy scores, where blue indicates higher performance and red indicates lower performance.}
\label{fig:appendix_heatmap_cat}
\end{figure*}

\subsection{VLM Comparison on Annotation}
We benchmark nine VLMs as element-presence annotators for SEA. \cref{fig:appendix_heatmap} summarizes overall Accuracy, F1, Precision, Recall, and Specificity, and \cref{fig:appendix_heatmap_cat} reports category-wise accuracy across the 14 CommonSketch groups. GPT-4o is the strongest and most consistent annotator across categories. Among open-source models, Molmo~\cite{deitke2024molmo} and Qwen2.5-VL~\cite{bai2025qwen2} perform best: Molmo achieves the highest open-source accuracy but is recall-heavy, whereas Qwen2.5-VL exhibits a more balanced precision--recall profile with higher specificity, and follows GPT-4o’s per-category trends most closely. InternVL3~\cite{zhu2025internvl3} and mPLUG-Owl3~\cite{ye2024mplug} remain competitive but show larger category-to-category fluctuations, while LLaVA~\cite{liu2023visual} and BLIP~\cite{li2022blip} perform substantially lower overall. PaliGemma2~\cite{steiner2024paligemma} often over-predicts elements which mean lower specificity, whereas SmolVLM~\cite{marafioti2025smolvlm} tends to under-predict, lower recall.

Performance also depends on category. Annotation is relatively easier for categories with clear and repeatedly visible part structure (e.g., \texttt{animal}, \texttt{structure}, \texttt{sports\_equipment}), and more difficult for categories with high intra-class shape variation or subtle defining cues (e.g., \texttt{clothing}, \texttt{container}). \cref{fig:appendix_heatmap} further indicates that maintaining both precision and specificity is important for SEA, since false positives can inflate the visual-representation term. Overall, these results support GPT-4o as the primary annotator and motivate Qwen2.5-VL as the strongest open-source substitute, while highlighting categories where annotator choice can most affect SEA scores.

\section{User Study Details}
\label{sec:appx-user}
We report the human evaluation protocol, including the survey interface, question format, abstraction rating guidelines, and sampling strategy. The study was approved by an Institutional Review Board (IRB).
\begin{figure*}
\centering
\includegraphics[width=0.99\linewidth]{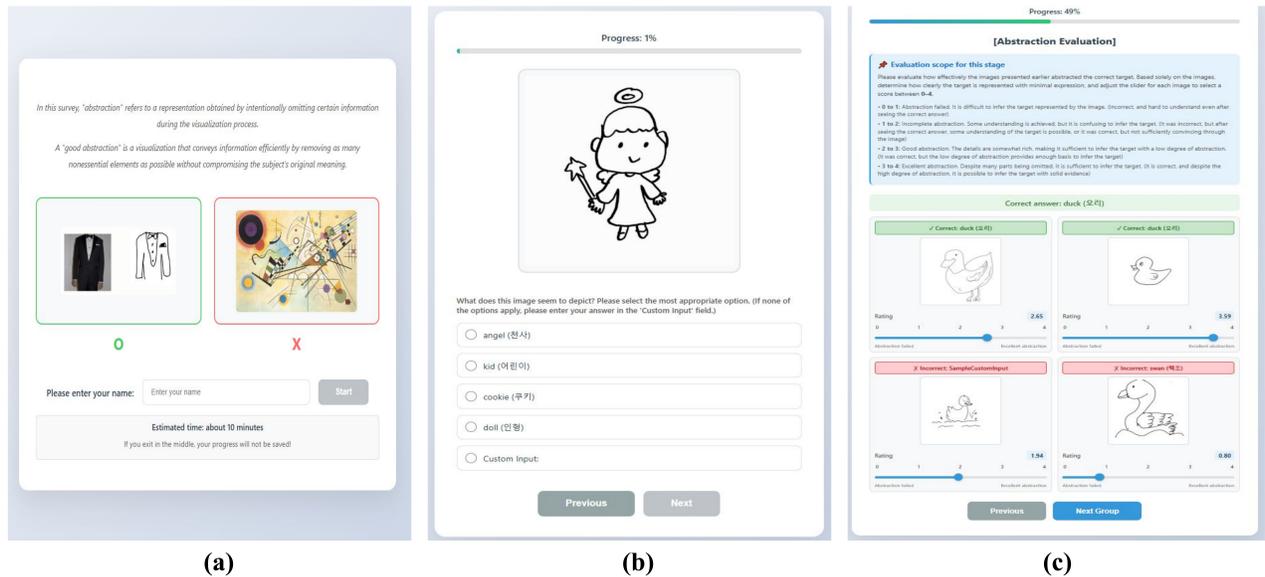}
\caption{\textbf{Human survey UI snapshots.}
(a) Landing page with study overview and consent.
(b) Classification interface where participants choose among four candidates, with an optional free-response field for alternative labels.
(c) Abstraction rating interface, where participants score each sketch on a continuous 0--4 slider with level-specific guidelines.}

\label{jaeyoonUI}
\end{figure*}
\subsection{User Study Interface and Instructions}
We conducted the study on a custom web-based survey platform. \cref{jaeyoonUI} presents the interface: participants first saw a short introduction and consent page (\cref{jaeyoonUI}a), then answered a classification question for a single sketch (\cref{jaeyoonUI}b), and finally rated abstraction for a set of four sketches from the same class (\cref{jaeyoonUI}c). During classification, a progress indicator was shown, and after submission participants were informed only whether the response was correct, without revealing the true label, to limit learning effects across questions.

For classification, we used a four-option multiple-choice format to reduce burden given the long session length. Because our model benchmarks span roughly 430 possible classes across CommonSketch, QuickDraw, and TU-Berlin, we additionally provided an optional free-response field so that participants could enter an alternative label when none of the four candidates matched their judgment.

For abstraction scoring, participants used a continuous 0--4 slider with interval-specific guidelines aligned to SEA:
\begin{itemize}
    \item 0--1: Abstraction failed; the target is hard to infer.
    \item 1--2: Incomplete abstraction; some cues exist but the target remains unclear.
    \item 2--3: Good abstraction; the target is clear but with noticeable detail.
    \item 3--4: Excellent abstraction; the target is clear despite strong simplification.
\end{itemize}

\begin{figure*}
\centering
\includegraphics[width=0.95\linewidth]{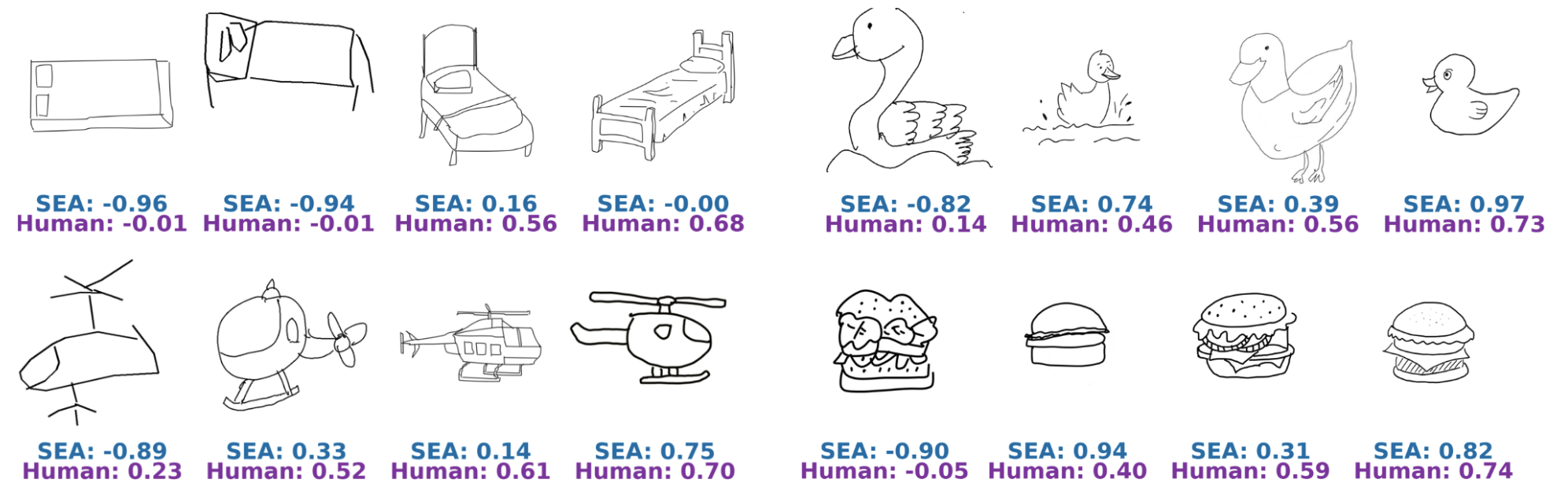}
\caption{Comparison of human abstraction score distributions and SEA metric scores across four images per class. Density curves represent individual images, with point markers showing their medians and vertical dashed lines indicating SEA scores. Four images in each class are annotated with their SEA scores and mean human abstraction scores.}
\label{jaeyoonreanalysis}
\end{figure*}

\paragraph{Sampling and study composition.}
To keep the session manageable, we evaluated 88 sketches in total. The core comparison set was drawn from CommonSketch, QuickDraw, and TU-Berlin: we selected one shared class from each of the 14 CommonSketch categories and sampled four sketches per class using quartiles of the model score distribution to cover a range of abstraction levels. 

We additionally tested generalization in two settings. First, for class-level out-of-distribution evaluation, we used SEVA sketches and selected four classes with four sketches each, following SEVA’s predefined abstraction levels. Second, for domain-shift evaluation, we used pictogram-style sketches from Art Pictogram~\cite{Pictogram2020} and Flaticon~\cite{Flaticoncrane1,Flaticoncrane2,Flaticoncrane3,Flaticonhorse1,Flaticonhorse2,Flaticonflower1,Flaticonflower2,Flaticonflower3,Flaticonflower4}, again selecting four classes with four sketches per class.

The survey was presented in three blocks: (i) the core cross-dataset comparison set, (ii) SEVA OOD samples, and (iii) pictogram-domain samples. Within each block, sketch order was randomized, and the four multiple-choice options were shuffled per question. Extremely unclear or unfinished sketches were excluded and replaced.

Multiple-choice candidates were initialized from the top CLIP predictions and manually filtered to remove obviously unrelated labels. When CLIP did not yield plausible distractors, alternatives were selected from a larger GPT-5 candidate list, while ensuring the correct label was always included among the four options.

\subsection{Human--SEA Alignment Analysis}
\label{sec:human-sea-alignment}

We assess alignment between SEA and human abstraction judgments by comparing their score distributions on the same sketches. \cref{jaeyoonreanalysis} shows four sketches per class with paired SEA and human scores. The two measures agree at the extremes: unrecognizable sketches receive low scores, while recognizable and well-abstracted sketches receive high scores, and class-level ranking trends are largely consistent. A clear gap emerges in the mid-range around a human score of about 0.5, where participants tend to favor drawings that are immediately recognizable even with extra detail, whereas SEA rewards sketches that retain core elements with fewer marks. Overall, SEA matches human intuition for failed versus successful abstraction, but applies a stricter preference for minimal sufficient evidence in borderline cases.

\subsection{Generalization on Pictograms}

\begin{figure}[h]
\centering
\includegraphics[width=0.95\linewidth]{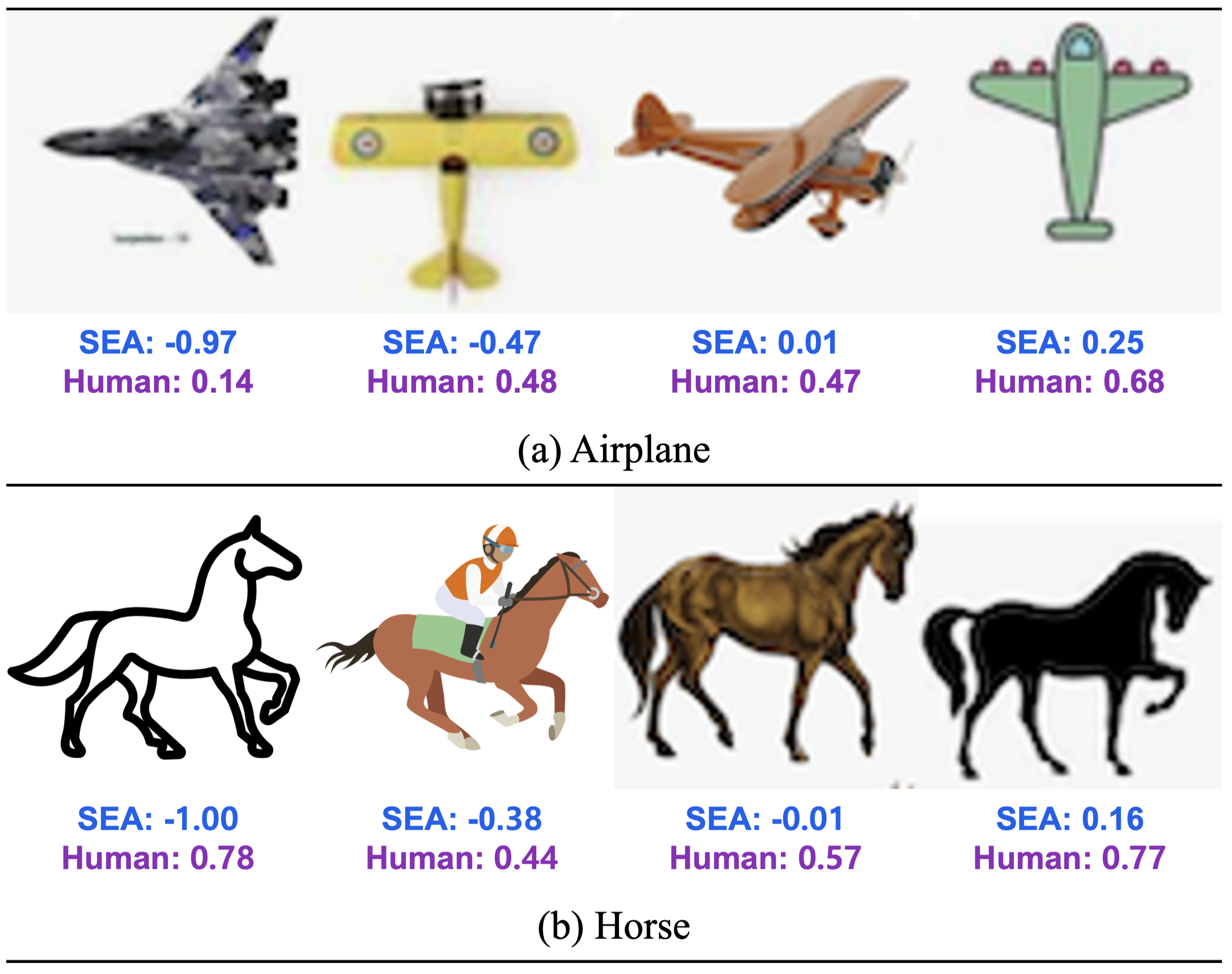}
\caption{Comparison of human scores and SEA values across pictograms to assess generalization ability.}
\label{jaeyoonoodexample}
\end{figure}
\begin{figure}[h]
\centering
\includegraphics[width=0.93\linewidth]{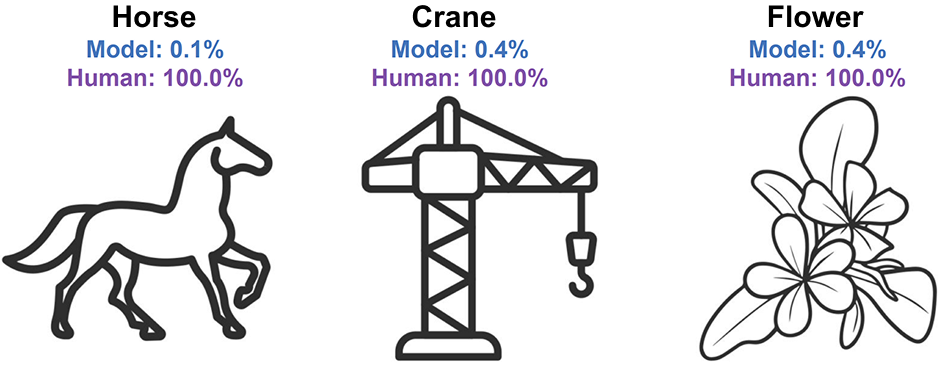}
\caption{Comparison between classification model performance and human accuracy.}
\label{jaeyoonpictoclip}
\vspace{-6pt}
\end{figure}
To examine whether the SEA generalizes beyond the sketch domain, we compared SEA with human evaluations. We observed that the model’s generalization ability increased in a manner similar to human scores, which can be seen in \cref{jaeyoonoodexample}. However, some clean line drawings occasionally caused an unexpected drop in classification performance, which can be seen in \cref{jaeyoonpictoclip}, leading to outlier metric values. These results suggest that SEA can be meaningfully applied to pictographic symbols that share the same underlying class semantics as our sketches, and that the metric exhibits a degree of generalization beyond the original sketch domain, provided that the backbone model has sufficient coverage of the visual style.

\clearpage
\onecolumn
\section{Prompts for Dataset Construction}
\label{sec:appx-prompt}

\subsection{Sketch Validation Cycle}
\label{sec:prompt-sketch-validation}
We use \textbf{GPT-4o} to generate multiple captions per sketch during the validation cycle.
The exact prompt is shown below.
\begin{promptbox}
\small
\textbf{Model}: GPT-4o

\medskip
\textbf{System message}\\
You are a helpful assistant for generating image captions.

\medskip
\textbf{User message}\\
Please describe this image 5 times based on the following format.
The input image is provided as a base64-encoded JPEG string.

\medskip
\textbf{Output template}
\begin{quote}
"A black line drawing of \{\{text1\}\} on a white background."\\
\textbf{OR}\\
"A simple drawing of \{\{text1\}\} on a white background."
\end{quote}

\textbf{Instructions}
\begin{itemize}
    \item Replace \{\{text1\}\} with a detailed description of the image.
    \item Avoid vague descriptions; focus on clear details such as objects, shapes, and actions.
    \item The fourth and fifth descriptions must focus on unexplained details in the other descriptions, except for the main object.
    \item Do not include ``\{\{\}\}'' in the final output.
    \item Choose the appropriate template based on the complexity of the image.
    \item Separate each description with \texttt{\textbackslash n\textbackslash n}.
    \item Do not put any numbers or symbols in front of the descriptions.
    \item Do not use commas (``,'').
\end{itemize}
\end{promptbox}

\subsection{Commonsense Extraction}
\label{sec:prompt-commonsense}

We use four language models for commonsense element extraction:
\textbf{GPT-4o}, \textbf{GPT-OSS-20B}, \textbf{Qwen-2.5 32B}, \textbf{Llama 3 8B}, and \textbf{Mistral 7B}.
The exact prompts are shown below.

\begin{promptbox}
\small
\textbf{Model}: GPT-4o

\medskip
\textbf{Instruction prompt}\\
You are a sketch analysis expert. Your task is to extract a structured list of \textbf{common visual elements} that are typically included — or \textbf{semantically expected} — when humans sketch a given object class.

Use the object class name, along with general visual common sense and knowledge of object structure, to infer \textbf{as many relevant visual components as possible}.

Your goal is to produce a \textbf{comprehensive and fine-grained breakdown} of visual parts, including:
\begin{itemize}
    \item core parts,
    \item minor or optional parts,
    \item functional attachments,
    \item repeated units,
    \item motion-related components (e.g., rotating blades, walking legs),
    \item relevant environmental or contextual elements.
\end{itemize}

Even if a part is rarely drawn, include it if it is \textbf{semantically meaningful or distinctive} for understanding or sketching the object, and assign lower \texttt{importance\_score} accordingly.

This output will be used to build a \textbf{commonsense database} for sketch abstraction, so prioritize \textbf{coverage and interpretability}.

\medskip
For each visual element, return the following fields:
\begin{itemize}
    \item \texttt{id}: in the form \texttt{\textless class\textgreater.\textless element\_name\textgreater}
    \item \texttt{name}: the name of the part
    \item \texttt{shape}: geometric form (e.g., circle, triangle, curve)
    \item \texttt{position}: typical relative location in the object
    \item \texttt{count}: usual number (e.g., 1, 2, or ``varies'')
    \item \texttt{importance\_score}: integer from 1 to 5 (5 = essential; 1 = optional or rare)
    \item \texttt{optional}: \texttt{true} or \texttt{false}
    \item \texttt{description}: what it looks like and why it is relevant
\end{itemize}

Return the result strictly in the following structure:
\begin{quote}
\ttfamily
\{\\
\quad "class": "\{class\_name\}",\\
\quad "total\_elements": \textless number\_of\_elements\textgreater,\\
\quad "elements": [\\
\quad\quad ...\\
\quad ]\\
\}\\
\end{quote}

Important general rules:
\begin{itemize}
    \item Do not include color information.
    \item Do not include fictional or humorous features.
    \item Do not include decorative elements unless they are functionally or culturally tied to the class.
    \item Use consistent, interpretable IDs in the format \texttt{\textless class\textgreater.\textless element\_name\textgreater}.
    \item Include both (1) frequently drawn elements and (2) structurally important elements even if rarely drawn.
    \item Include context or environment features only if they are logically essential to how the object is typically depicted.
    \item Think about what makes this class visually different from nearby classes and reflect that in part selection.
    \item Favor over-inclusion: include more elements with appropriately scaled \texttt{importance\_score}.
\end{itemize}

Class: \texttt{\{class\_name\}}
\end{promptbox}

\begin{promptbox}
\small
\textbf{Models}: GPT-OSS 20B, Qwen2.5 32B, Llama 3 8B, Mistral 7B

\medskip
\textbf{Instruction prompt}\\
You are a \textbf{Structured Visual Object Analyzer for sketches}.  
Your job is to output a \textbf{single JSON object} describing the visible parts of a sketched object class.

\medskip
\textbf{Hard rules} (follow strictly):
\begin{enumerate}
    \item \textbf{No environment-only items.}\\
    Do not include background or scene items that are not intrinsic parts of the object (e.g., no clouds for \emph{sun}, no road or buildings for \emph{car}).

    \item \textbf{Visible-only.}\\
    Include only parts plausibly visible in a typical sketch; exclude hidden internals (e.g., car engine, phone mainboard).

    \item \textbf{Variants allowed, naming rules apply.}\\
    Common sketch variants replacing or decorating real parts may be included with \texttt{"optional": true} (e.g., \texttt{human\_mouth} on an insect).  
    Do not use the word ``stylized''; do not use parentheses or brackets.  
    All names must be in \textbf{snake\_case} (lowercase, digits allowed, words separated by a single underscore).  
    Examples: \texttt{steam\_lines}, \texttt{wing\_vein\_lines}, \texttt{tail\_fan}, \texttt{human\_mouth}.

    \item \textbf{Expressive lines/effects.}\\
    Expressive effects (e.g., \texttt{airflow\_lines}, \texttt{motion\_lines}, \texttt{steam\_lines}, \texttt{sparkle}) are excluded by default.  
    They may be included only when they represent an essential and commonly used feature of the object’s sketch and must then be marked \texttt{"optional": true}.  
    Background-only elements (ground, sky, clouds, water, etc.) must still be excluded.

    \item \textbf{Merge symmetric or duplicated parts.}\\
    Merge symmetric repeats (e.g., left/right wheels, pairs of legs) into a single element (e.g., \texttt{wheels}, \texttt{legs}).

    \item \textbf{Coverage and granularity.}\\
    Produce a rich but concise set of features (recommended 9--16 elements).  
    Prefer coarse-to-mid granularity: split obvious appendages or facial parts (head, arms, legs) instead of using a single \texttt{body}.  
    Consider including elements from:  
    (a) anatomy or core shape,  
    (b) facial features,  
    (c) iconic clothing or accessories,  
    (d) explicit surface, texture, or pattern marks (e.g., \texttt{seed\_dots}, \texttt{peel\_lines}, \texttt{feather\_lines}, \texttt{shell\_pattern}, \texttt{fur\_lines}),  
    (e) expressive lines only if visibly drawn.

    \item \textbf{Ground truth first, then variants.}\\
    List physically correct parts first, then common variants or expressive features with \texttt{"optional": true}.

    \item \textbf{Labelability and non-ambiguous features.}\\
    Every feature must be binary labelable (0/1) from the sketch without subjective judgment.  
    Disallow vague descriptors (e.g., \texttt{smooth\_surface}, \texttt{shiny\_surface}).  
    Do not describe the absence of texture (e.g., \texttt{no\_texture}).  
    Prefer positive, observable evidence (lines, dots, edges, explicit patterns), such as \texttt{glaze\_lines}, \texttt{seed\_dots}, \texttt{crack\_lines}, \texttt{slice\_lines}.
\end{enumerate}

\medskip
\textbf{Output format (exact structure):}
\begin{quote}
\ttfamily
\{\\
\quad "class": "\textless object\_name\textgreater",\\
\quad "total\_elements": \textless int\textgreater,\\
\quad "elements": [\\
\quad\quad \{\\
\quad\quad\quad "id": "\textless class\_name\textgreater.\textless part\_name\textgreater",\\
\quad\quad\quad "name": "\textless part\_name\textgreater",\\
\quad\quad\quad "optional": \textless true or false\textgreater\\
\quad\quad\},\\
\quad\quad \dots\\
\quad ]\\
\}\\
\end{quote}

Additional guidance:
\begin{itemize}
    \item \texttt{"total\_elements"} must equal the number of objects in \texttt{"elements"}.
    \item Output JSON only (no commentary outside the JSON).
    \item \texttt{\textless part\_name\textgreater} must be snake\_case; IDs must be of the form \texttt{\textless class\_name\textgreater.\textless part\_name\textgreater}.
    \item Ensure at least 8 elements (prefer 9--16) and reasonable coverage of anatomy, facial features, accessories, surface or pattern, and expressive lines.
    \item Ensure all features are 0/1 labelable and not environment-only.
\end{itemize}

Final instruction: Now, provide the structured JSON for the following object: \texttt{\{word\}}.
\end{promptbox}

\subsection{Element Annotation}
\label{sec:prompt-element-annotation}

For element-level commonsense VQA, we use the following vision--language models:
\textbf{GPT-4o}, \textbf{Qwen2.5-VL 7B}, \textbf{mPLUG-Owl3 7B}, \textbf{InternVL3 8B},
\textbf{Molmo 7B}, \textbf{PaliGemma2 3B}, \textbf{SmolVLM 500M}, \textbf{LLaVA 1.5 7B},
and \textbf{BLIP}.  
We list the exact prompts below, grouping models that share the same template.

\begin{promptbox}
\small
\textbf{Models}: GPT-4o, Qwen2.5-VL 7B, mPLUG-Owl3 7B

\medskip
\textbf{Instruction prompt}\\
\texttt{<|image|>}\\
You are a strict vision auditor for sketched objects. \\
Target class: "\texttt{\{class\_name\}}".\\
Valid elements for this class (use \textbf{only} these ids; do not add new keys):\\
\texttt{\{element\_block\}}\\[0.3em]
Task: For \textbf{each} element id above, return only whether the element is depicted (true/false).\\
Do not return counts. If ambiguous, use false.\\
Return only a compact JSON object with element ids as keys and boolean values (true/false).\\
No prose, no code block, no extra keys.\\[0.3em]
Example schema (structure only):
\begin{quote}
\ttfamily
\{\\
\quad "element\_id\_1": true,\\
\quad "element\_id\_2": false,\\
\quad \dots\\
\}\\
\end{quote}
\end{promptbox}

\begin{promptbox}
\small
\textbf{Models}: InternVL3 8B, PaliGemma2 3B, SmolVLM 500M

\medskip
\textbf{Question template}\\
In this \texttt{\{class\_name\}} image, does this sketch contain a \texttt{\{e\}}?\\
Answer exactly ``yes'' or ``no''.

\medskip
For SmolVLM we wrap the question in the model's chat template:
\begin{quote}
\ttfamily
\textless|user|\textgreater\\
\textless image\textgreater\\
In this \{class\_name\} image, does this sketch contain a \{e\}? Answer exactly 'yes' or 'no'.\\
\textless|end|\textgreater\\
\textless|assistant|\textgreater
\end{quote}
\end{promptbox}

\begin{promptbox}
\small
\textbf{Model}: Molmo-7B-D-0924

\medskip
\textbf{Instruction prompt}\\
You are an assistant that analyzes an image of a \texttt{\{category\}} and answers in JSON format only.

Task: For the given \texttt{\{category\}} image, decide if each of the following elements is present (1) or not present (0):\\
\texttt{[\{element\_list\}]}.

Return the result strictly as a JSON object in the following format:
\begin{quote}
\ttfamily
\{\\
\quad "\{file\_name\}": \{\\
\quad\quad \{element\_lines\}\\
\quad\}\\
\}\\
\end{quote}
Do not include explanations or extra text. Output only valid JSON.
\end{promptbox}

\begin{promptbox}
\small
\textbf{Model}: LLaVA 1.5 7B

\medskip
\textbf{Question template}\\
In this \texttt{\{category\}} image, is there a \texttt{\{element\}}?\\
Answer Yes or No.
\end{promptbox}

\begin{promptbox}
\small
\textbf{Model}: BLIP (blip-vqa-capfilt-large)

\medskip
\textbf{Question template}\\
In this \texttt{\{category\}} image, is there a \texttt{\{element\}}?
\end{promptbox}

\begin{center}
\scriptsize
\setlength{\tabcolsep}{3pt}
\begin{longtable}{@{}L{0.10\textwidth}L{0.12\textwidth}C{0.07\textwidth}L{0.68\textwidth}@{}}
\caption{\textbf{Per-class element list.}
For each class, we list the commonsense elements extracted by GPT-4o and GPT-OSS.
Black indicates elements shared by both models;
\textcolor{red}{red} denotes elements unique to GPT-4o;
\textcolor{blue}{blue} denotes elements unique to GPT-OSS.
The 4o/OSS column reports the number of elements extracted by GPT-4o and GPT-OSS 20B, respectively.}
\label{tab:per-class-elements-merged}\\
\toprule
Category & Class & 4o/OSS & Elements \\
\midrule
\endfirsthead

\toprule
Category & Class & 4o/OSS & Elements \\
\midrule
\endhead

\midrule
\multicolumn{4}{r}{\emph{Continued on next page}}\\
\bottomrule
\endfoot

\bottomrule
\endlastfoot

\multirow[t]{61}{*}{animal} & alpaca & 13/11 & body, ears, eyes, head, legs, tail, \textcolor{red}{feet}, \textcolor{red}{fleece}, \textcolor{red}{hooves}, \textcolor{red}{muzzle}, \textcolor{red}{neck}, \textcolor{red}{nostrils}, \textcolor{red}{smile}, \textcolor{blue}{fur\_lines}, \textcolor{blue}{motion\_lines}, \textcolor{blue}{mouth}, \textcolor{blue}{nose}, \textcolor{blue}{whisker\_lines} \\
 & ant & 11/9 & abdomen, antennae, head, legs, mandibles, thorax, \textcolor{red}{compound eyes}, \textcolor{red}{jointed legs}, \textcolor{red}{mouth}, \textcolor{red}{petiole}, \textcolor{red}{stinger}, \textcolor{blue}{body}, \textcolor{blue}{eyes}, \textcolor{blue}{segment\_lines} \\
 & bat & 14/12 & body, ears, eyes, head, mouth, nose, tail, wings, \textcolor{red}{feet}, \textcolor{red}{fingers}, \textcolor{red}{fur}, \textcolor{red}{legs}, \textcolor{red}{teeth}, \textcolor{red}{wing membrane}, \textcolor{blue}{claws}, \textcolor{blue}{fur\_lines}, \textcolor{blue}{motion\_lines}, \textcolor{blue}{wing\_vein\_lines} \\
 & bee & 14/10 & abdomen, antennae, head, legs, mouth, stinger, thorax, wings, \textcolor{red}{compound eyes}, \textcolor{red}{flower context}, \textcolor{red}{flying motion}, \textcolor{red}{hairs}, \textcolor{red}{mouthparts}, \textcolor{red}{stripes}, \textcolor{blue}{body\_stripe\_lines}, \textcolor{blue}{eyes} \\
 & bird & 15/10 & beak, body, feet, head, legs, tail, wings, \textcolor{red}{branch}, \textcolor{red}{crest}, \textcolor{red}{eyes}, \textcolor{red}{feathers}, \textcolor{red}{flying pose}, \textcolor{red}{standing pose}, \textcolor{red}{tail feathers}, \textcolor{red}{wing feathers}, \textcolor{blue}{eye}, \textcolor{blue}{feather\_lines}, \textcolor{blue}{motion\_lines} \\
 & boar & 13/9 & body, ears, head, legs, snout, tail, tusks, \textcolor{red}{brush}, \textcolor{red}{eyes}, \textcolor{red}{hair}, \textcolor{red}{hooves}, \textcolor{red}{hump}, \textcolor{red}{mane}, \textcolor{blue}{eye}, \textcolor{blue}{fur\_lines} \\
\end{longtable}
\end{center}

\clearpage
\twocolumn


\end{document}